%% file: main.tex
\documentclass{article}

\usepackage{microtype}
\usepackage{graphicx}
\usepackage{subfigure}
\usepackage{booktabs} % for professional tables

\usepackage{hyperref}

\usepackage[accepted]{icml2023}

\usepackage{amsmath}
\usepackage{amssymb}
\usepackage{mathtools}
\usepackage{amsthm}

\usepackage[capitalize,noabbrev]{cleveref}

\theoremstyle{plain}

\theoremstyle{definition}

\theoremstyle{remark}

\usepackage[textsize=tiny]{todonotes}

\input{math_commands.tex}

\usepackage{hyperref}
\usepackage{url}
\usepackage{float}

\def\x{{\bm{x}}}

\def\w{{\bm{w}}}

\def\dt{\eta}
\def\I{{\bm{I}}}
\def\Ba{{\mathbb{B}}}

\newcommand{\beq}{\begin{equation}}
\newcommand{\eeq}{\end{equation}}
\newcommand{\lpa}{\left(}
\newcommand{\rpa}{\right)}

\newcommand{\loss}{l}

\newcommand{\cov}{\bm{\Sigma}}

\newcommand{\ed}{\delta}
\newcommand{\ec}{\gamma}
\newcommand{\ei}{\zeta}

\definecolor{NavyBlue}{rgb}{0.0, 0.0, 0.5}
\definecolor{Blue}{rgb}{0.0, 0.0, 1.0}
\usepackage{hyperref}
\hypersetup{
    colorlinks=true,
    linkcolor=blue, % BlueViolet
    filecolor=magenta,      
    urlcolor=NavyBlue,
    citecolor=Blue
}

\definecolor{green}{RGB}{31,198,0}
\definecolor{airforceblue}{rgb}{0.36, 0.54, 0.66}
\definecolor{amaranth}{rgb}{0.9, 0.17, 0.31}
\definecolor{matthieu}{RGB}{0,0,255}

\definecolor{airforceblue}{rgb}{0.36, 0.54, 0.66}

\icmltitlerunning{Dissecting the Effects of SGD Noise in Distinct Regimes of Deep Learning}

\begin{document}

\twocolumn[
\icmltitle{Dissecting the Effects of SGD Noise in Distinct Regimes of Deep Learning}



\icmlsetsymbol{equal}{*}

\begin{icmlauthorlist}
\icmlauthor{Antonio Sclocchi}{epfl}
\icmlauthor{Mario Geiger}{mit}
\icmlauthor{Matthieu Wyart}{epfl}
\end{icmlauthorlist}

\icmlaffiliation{epfl}{Institute of Physics, École Polytechnique Fédérale de Lausanne, Lausanne, 1015, Switzerland}
\icmlaffiliation{mit}{Department of Electrical Engineering and Computer Science, Massachusetts Institute of Technology, Cambridge, MA, USA}

\icmlcorrespondingauthor{Antonio Sclocchi}{antonio.sclocchi@epfl.ch}

\icmlkeywords{SGD, Machine Learning, ICML}

\vskip 0.3in
]



\printAffiliationsAndNotice{}  % leave blank if no need to mention equal contribution
% for the arXiv
% \printAffiliations

\begin{abstract}
Understanding when the noise in stochastic gradient descent (SGD) affects generalization of deep neural networks remains a challenge, complicated by the fact that networks can operate in distinct training regimes. Here we study how the magnitude of this noise $T$ affects performance as the size of the training set $P$ and the scale of initialization $\alpha$ are varied. For gradient descent, $\alpha$ is a key parameter that controls if the network is `lazy'  ($\alpha\gg1$) or instead  learns features ($\alpha\ll1$). 
For classification of MNIST and CIFAR10 images,  our central results are:
\textit{(i)} obtaining phase diagrams for performance in the $(\alpha,T)$ plane. They show that SGD noise can be detrimental or instead useful depending on the training regime. Moreover, although increasing $T$ or decreasing $\alpha$ both allow the net to escape the lazy regime, these changes can have opposite effects on performance. 
\textit{(ii)} Most importantly, we find that the characteristic temperature $T_c$ where the noise of SGD starts affecting the trained model (and eventually performance) is a power law of $P$. We relate this finding with the observation that  key dynamical quantities, such as the total variation of weights during training, depend on both $T$ and $P$ as power laws.
These results indicate that a key effect of SGD noise occurs late in training, by affecting the stopping process whereby all data are fitted. Indeed, we argue that due to SGD noise, nets must develop a stronger `signal', i.e. larger informative weights, to fit the data, leading to a longer training time. A stronger signal and a longer training time are also required when the size of the training set $P$ increases. We confirm these views in the perceptron model, where signal and noise can be precisely measured. Interestingly,  exponents characterizing the effect of SGD  depend on the density of data near the decision boundary, as we explain. 
\end{abstract}

\section{Introduction}
Optimizing the generalization performances of overparametrized neural networks is one of the main challenges in machine learning. A crucial role is played by gradient-based training algorithms, which converge to solutions which generalize well also when no explicit regularization of the model is used \citep{zhang2021understanding}. Mini-batch stochastic gradient descent (SGD) %, together with its variations, 
is the workhorse algorithm to train modern neural networks. Yet, key aspects of these algorithms are debated. 

{\it Effect on performance:}
A popular idea has been that mini-batch SGD can generalize better than full batch gradient descent (GD) \citep{heskes1993, lecun2012, keskar2016, hochreiter1997flat, jastrzkebski2017, entropysgd2019},
yet this view is debated \citep{hoffer2017, dinh2017, shallue2018, zhang2019}. 
In fact, comparing SGD and GD at fixed number of training epochs leads to a generalization gap \citep{keskar2016} that can be closed by training longer with a fixed number of training steps \citep{hoffer2017, smith2020}. More generally, the choice of the computational budget can 
affect which algorithm performs better \citep{shallue2018,smith2020}.

{\it Theories for the role of SGD:} Several works have argued that larger SGD stochasticity leads the dynamics toward flatter minima of the loss landscape, and it has been argued that this effect leads to improved performances \citep{hochreiter1997flat, keskar2016, zhang2018energy, smith2018bayesian, wu2018sgd}. By contrast, other studies  suggest that the  SGD noise biases the model in a manner similar to initializing the network with small weights, and helps recovering sparse predictors \citep{blanc2020implicit, haochen2021, flammarion2021}.

\subsection{This work}

In this work, we clarify these two debates by performing  systematic empirical studies of how  performance is affected by the noise magnitude of SGD or temperature $T$ (the ratio between the learning rate $\dt$ and the batch size $B$ \citep{jastrzkebski2017, zhang2019, smith2020}), by the initialization scale $\alpha$, and by the size of the training set $P$. The initialization scale $\alpha$ was rarely considered in empirical studies so far, yet it governs the training regimes in which nets operate. For large $\alpha$, tiny changes of weights are sufficient to fit the data: the predictor is approximately linear in its parameters, corresponding to the \textit{kernel} or \textit{lazy} regime \citep{jacot2018, chizat2019lazy}. By contrast for small initialization, 
networks can learn the relevant features of the task and the dynamics is non-linear, corresponding to the so-called feature-learning regime \citep{rotskoff2018, mei2018, sirignano2020}.

We also deal with the computational budget issue by considering the hinge loss $\loss(y,\hat{y})=(1-y\hat{y})^+$, allowing us to train networks until the time $t^*$ where the loss is strictly zero, and the dynamics stops.
Importantly, this training methodology is not restrictive, as it yields similar outcomes compared to training with the cross-entropy loss and performing early stopping.
\footnote{In Appendix \ref{app:hinge_cross} we verify that the two training methodologies give identical power-law dependencies for all the quantities we analyse in this work.}

Our central empirical results are: 
\begin{itemize}
    \item[(i)] obtaining phase diagrams for performance in the $(\alpha,T)$ plane. They show that SGD noise can be detrimental or instead useful depending on the training regime, even in the absence of budget constraints. This observation clarifies why different conclusions on the benefits of SGD were previously made.
    \item[(ii)] Although we find that increasing $T$ or decreasing $\alpha$ both allow the net to escape the lazy regime, these changes can have opposite effects on performance, in disagreement with simple models  \citep{flammarion2021}.
    \item[(iii)] We reveal that several observables characterizing the dynamics follow scaling laws in $T$ and $P$. Denote by $\Delta w$ the relative weight variation accumulated after training and  $t^*$ the training time defined as the learning rate times the number of training steps required to bring a hinge loss to zero. We find that
    \begin{equation}
     \Delta w\sim   T^\ed P^\ec,\ \ t^*\sim T P^b,
     \label{eq:intro_scalings}
    \end{equation}
    where $\ed,\ec,b$ are exponents depending on the model and the training regime.    
    \item[(iv)] Most importantly,
    we find that SGD noise starts affecting the trained model at a characteristic temperature scale $T_c$ which depends on the size of the training set $P$ as
    \beq
        T_{c}\sim P^{-a},
        \label{eq:Tc}
    \eeq
    where $a$ is a model-dependent exponent. This result can be understood as follows.
    For the lazy regime $\alpha\gg 1$, $T_c$ is the temperature at which the network exits the lazy regime, i.e. $\Delta w={\cal O}(1)$. Together with \ref{eq:intro_scalings}, it  gives $a=\gamma/\delta$ in agreement with our observations.
    For the feature regime, $T_c$ corresponds to the transition between a low-$T$ regime, where $\Delta w$ is unaffected by SGD noise and is found to scale as $\Delta w\sim P^\ei$, and a high-$T$ regime where \ref{eq:intro_scalings} applies.
    These two empirical relationships imply that $T_c \sim P^{-a}$, with the exponent $a$ satisfying $a=(\ec-\ei)/\ed$, consistent with our experimental observations.
    For fully-connected architectures, we observe that $T_c$ also characterizes the temperature where SGD affects performance. By contrast, for CNNs such a characteristic temperature is hard to extract from the performance curves, while it is clearly identified from the weight variation. 

\item[(v)] We rationalize these findings using a teacher-student perceptron model, for which $ \Delta w$ and $t^*$ also display power-law dependence on $T$ and $P$. We show that SGD noise increases weights in directions irrelevant to the task, implying that the correct weights must grow much larger to fit data, thus increasing both $t^*$ and $\Delta w$. We compute the dependence of these effects on the size of the training set, and show that this dependence varies qualitatively with the distribution of data near the decision boundary.
\end{itemize}

Overall, instead of a static view where SGD noise would bias networks toward broader minima of the population loss, these results support a dynamical viewpoint where SGD noise delays the end of training. This effect allows the weights to grow more, affecting performance the most when the network escapes the lazy regime.

\begin{figure*}
    \centering
    \includegraphics[width=0.49\textwidth]{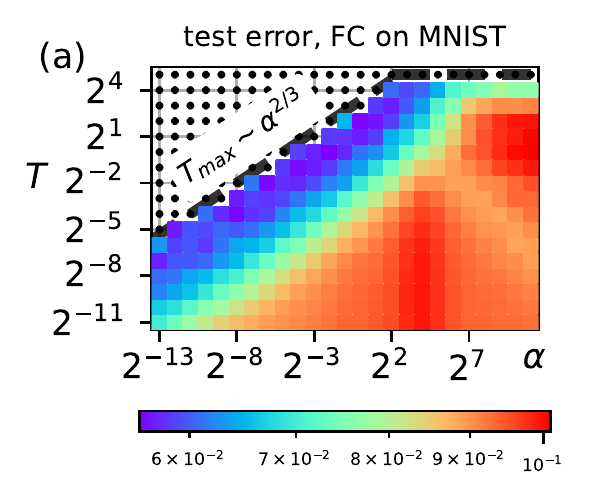}
    \includegraphics[width=0.49\textwidth]{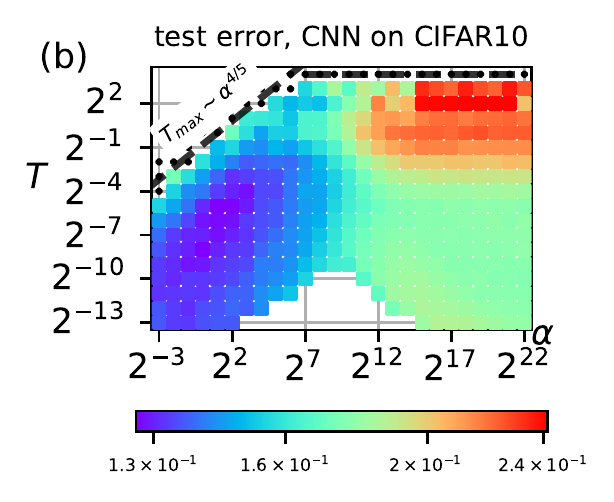}
    \vspace{-.5 cm}
    \caption{\textbf{Test error of deep networks on image data-sets, for varying $T$ and $\alpha$ and fixed $P=1024$.} Batch size $B$ is kept fixed and learning rate $\eta$ is varied ($T=\eta/B$). \textbf{(a)} 5-hidden layers fully-connected network (FC) on parity MNIST with $B=16$. \textbf{(b)} 9-hidden layers CNN (MNAS) on CIFAR (animals vs the rest) with $B=64$.
    Black dots correspond to  training runs that do not converge. The black dashed lines indicates the maximal temperatures $T_{max}$.
    The lowest test error is achieved in the feature regime ($\alpha\ll 1$), for a temperature $T_{opt}\propto T_{max}$. In the lazy regime ($\alpha\gg 1$), performance is best for the highest $T$ for FC on MNIST. Although it is not apparent here for CNN on CIFAR, it is also the case as the training set increases (see below). In \textit{(a)}, the number of hidden layers is $D=5$ and $T_{max}\sim \alpha^{\frac{D-1}{D+1}} = \alpha^{\frac{2}{3}}$ (black dashed line) when $\alpha\ll1$ as argued in \ref{eq:T_feature}. Similarly in \textit{(b)}, $D=9$ and $T_{max}\sim \alpha^{\frac{D-1}{D+1}} = \alpha^{\frac{4}{5}}$ (black dashed line).
    }
    \label{fig:test_phase}
\end{figure*}

\subsection{Related works}
More related works are indicated in Appendix \ref{app:other_works}.

\section{Empirical analysis}
\label{sec:empiric}

\subsection{General setting and notation}
\label{sec:definition}
We consider binary classification on the data $\{\x_{\mu}\}_{\mu=1,...,P}$ with labels $\{y_\mu\}_{\mu=1,...,P} \in \{-1, +1\}$. $P$ is the size of the training set. Given a predictor $\hat{y}_{\mu}$, the hinge loss on the sample $\mu$ is defined as $\loss(y_{\mu},\hat{y}_{\mu}) = (1-y_{\mu}\hat{y}_{\mu})^+$, where $(x)^+ = \max(0,x)$. To control between feature and lazy training, we 
multiply the model output by $\alpha$ \citep{chizat2019lazy}. For the hinge loss, this is equivalent to changing the loss margin to $1/\alpha$. Therefore we study the training loss
\beq
L(\w) = \frac{1}{P} \sum_{\mu=1}^P (\alpha^{-1}-y_{\mu} F(\w,\x_{\mu}))^+,
\label{eq:hingeLoss}
\eeq
where $F(\w,\x_{\mu})$ is the model predictor with weights $\w$ on the datum $\x_{\mu}$. The model predictor at time $t$ corresponds to $F(\w,\x_{\mu}) = f(\w^t,\x_\mu) - f(\w^0,\x_\mu)$, where $f(\w^t,\x_\mu)$ is the  output of a neural net with weights $\w^t$ at time $t$ and $\w^0$ are the weights at initialization. 
For a network of width $h$, the weights are initialized as Gaussian random numbers with standard deviation $1/\sqrt{h}$ for the hidden layers and $1/h$ for the output layer.
Such an initialization ensures that the feature learning limit corresponds to $\alpha\ll1$ while the lazy training limit corresponds to $\alpha\gg 1$, and that every layer has a similar change of weights \citep{geiger2020disentangling,yang2021tensor}.\\

The stochastic gradient descent updating equation is:
\beq\label{eq:SGD}
    \w^{t+\dt} =\w^t + \frac{\dt}{B}\sum\limits_{\mu \in \Ba_t} \theta\lpa\alpha^{-1}-y_{\mu} F(\w,\x_{\mu})\rpa y_{\mu} \nabla_{\w} f(\w^t,\x_\mu)
\eeq
where $\theta(x)$ is the Heaviside step function, $\Ba_t \subset \{1,...,P\}$ is the batch at time $t$ and $B$ is its size. The time $t$ corresponds to the number of training steps times the learning rate $\dt$. The batch $\Ba_t$ is randomly selected at each time step among all the $P$ data. %, and this introduces stochasticity in the process. 
The learning rate $\dt$ is kept constant during training. The end of training is reached when $L(\w^{t^*})=0$.\\
The batch size $B$ is taken small enough to be in the ``noise dominated'' regime \citep{smith2020, zhang2019}, where the dynamics depends on the SGD temperature $T=\dt/B$. Empirical verification of this fact is provided in Appendix \ref{app:eta_B}.\\

Below we use a 5-hidden-layers fully-connected (FC) network and a 9-hidden-layers convolutional neural network (CNN) (MNAS architecture \citep{mnasnet}). In Appendix \ref{app:plots_lazy} we report data also for a 3-hidden layers CNN (simple-CNN). We consider the binary datasets MNIST (even vs odd numbers) and CIFAR10 (animals vs the rest). All the networks use ReLU as activation functions.
The code with all the details of the experiments is provided at \href{https://tinyurl.com/mrys4uyp}{https://tinyurl.com/mrys4uyp}.

\begin{figure*}
    \centering
    \includegraphics[width=1.0\textwidth]{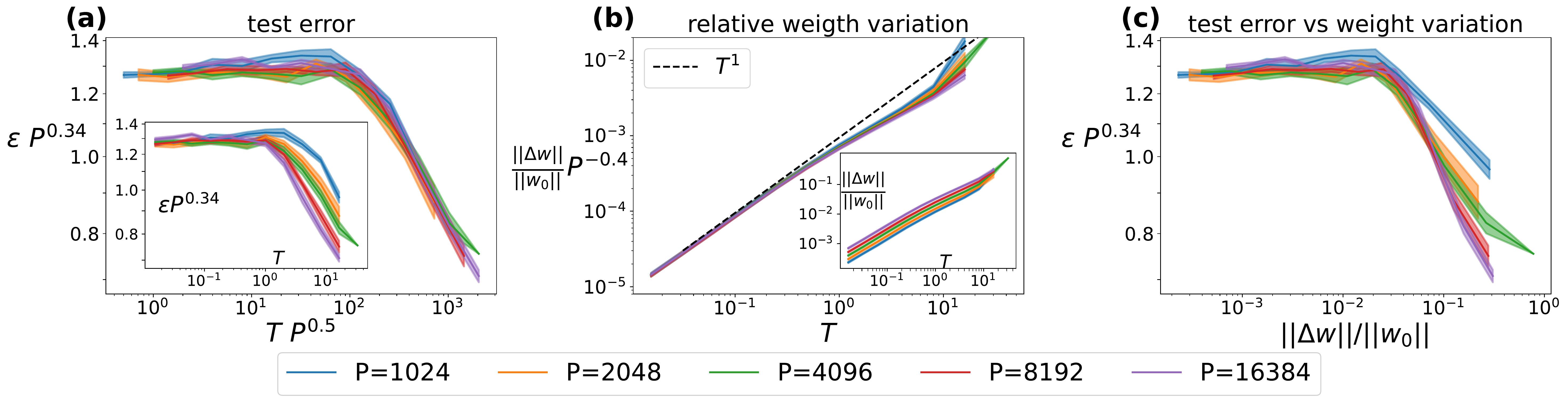}
    \vspace{-.5 cm}
    \caption{\textbf{FC on MNIST, lazy regime, $\alpha=32768$, $B=16$, $T=\eta/B$.} 
    \textbf{(a): test error ($\epsilon$) vs temperature ($T$).} \textit{Inset:} $\epsilon$ starts improving at a cross-over temperature $T_{c}$ depending on $P$. The y-axis is rescaled by $P^\beta$, with $\beta$ some fitting exponent, to align $\epsilon$ at $T_c$. \textit{Main:} Rescaling the x-axis by $P^{0.5}$ aligns horizontally the points where $\epsilon$ starts improving, suggesting a dependence $T_c \sim P^{-0.5}$.
    \textbf{(b): total weight variation at the end of training normalized with respect to their initialization ($||\Delta \w||/||\w_0||$) vs $T$.} \textit{Inset:} $||\Delta \w||/||\w_0||$ increases with both $T$ and $P$. \textit{Main:} Plotting $\Delta w P^{-\ec}$ yields a curve increasing approximately as $T^\ed$, suggesting $\Delta w\sim T^\ed P^{\ec}$, with $\ec\approx 0.4$ and $\ed\approx 1$.
    \textbf{(c): test error vs weight variation.}
    Plotting $\epsilon$ vs $||\Delta \w||/||\w_0||$ for different $P$ aligns the point where $\epsilon$ starts improving.
    }
    \label{fig:FC_lazy}
\end{figure*}

\subsection{Performance in the $(\alpha, T)$ phase diagram}
Fig. \ref{fig:test_phase}-(a)  shows the test error for a FC network trained on MNIST and Fig. \ref{fig:test_phase}-(b) shows the same quantity obtained after training a CNN on CIFAR10. The black dots correspond to training loss exploding to infinity due to too large learning rate. Therefore, the dashed back lines indicate the maximal temperature $T_{max}$ for which SGD converges. \\

From Fig. \ref{fig:test_phase} we make the following observations:

(i) In the feature regime, both $T_{max}$ and the temperature of optimal performance $T_{opt}$ follow $T_{max}\sim T_{opt}\sim \alpha^k$. In Appendix \ref{app:scaling_feature}, we relate the exponent $k$ to the number $D$ of hidden layers of the network as $k=(D-1)/(D+1)$.
In the lazy regime, $T_{max}$ and $T_{opt}$ are independent of $\alpha$.

(ii) In Fig. \ref{fig:test_phase}-(a), in the lazy regime (largest $\alpha$), increasing $T$ leads to an initial slight degradation of the test error followed by an improvement just before reaching the instability $T_{max}$.

(iii) In Fig. \ref{fig:test_phase}-(b), in the lazy regime, increasing $T$ leads to a degradation of the test error before reaching the instability $T_{max}$ (for larger $P$, a region of good performance appears near $T_{max}$, see below). In this regime increasing $T$ or decreasing $\alpha$ have opposite effects, showing that in general an increase of SGD noise is not equivalent to making the initialization smaller.

\subsection{Role of size of the training set $P$}
\label{sec:role_P}

This section focuses on the impact of the size of the training set which, surprisingly, determines the SGD noise scale that affects performances.

\subsubsection{Lazy regime}
\begin{figure*}
    \centering
    \includegraphics[width=1.0\textwidth]{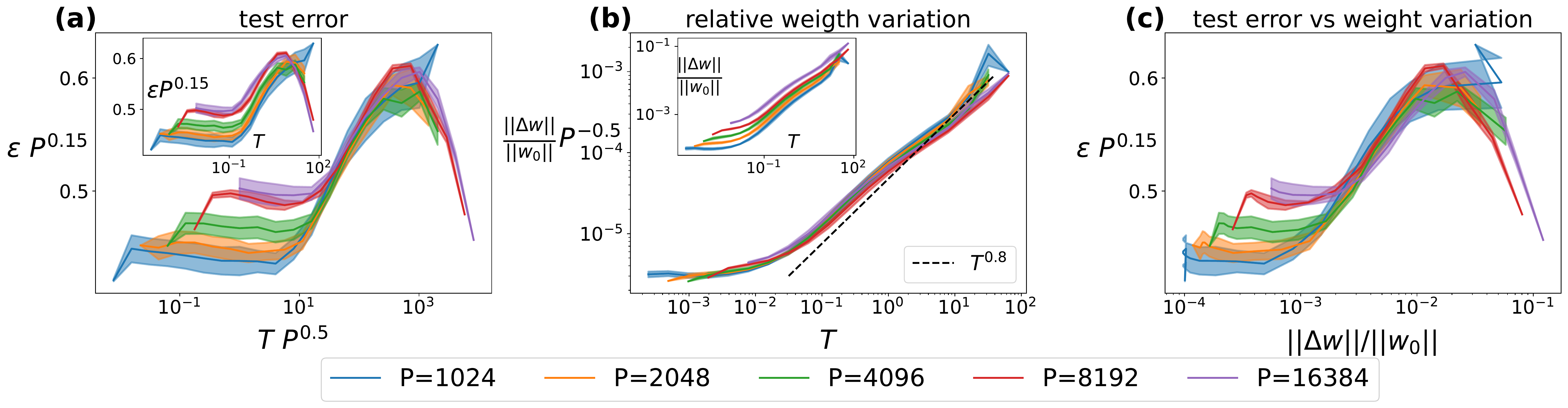}
    \vspace{-.5 cm}
    \caption{\textbf{CNN on CIFAR, lazy regime, $\alpha=32768$, $B=16$, $T=\eta/B$.} 
    \textbf{(a): test error ($\epsilon$) vs temperature ($T$).} \textit{Inset:} $\epsilon$ starts improving at a cross-over temperature $T_{c}$ depending on $P$. The y-axis is rescaled by $P^\beta$, with $\beta$ some fitting exponent, to align $\epsilon$ at $T_c$. \textit{Main:} Rescaling the x-axis by $P^{0.5}$ aligns horizontally the points where $\epsilon$ starts improving, suggesting a dependence $T_c \sim P^{-0.5}$.
    \textbf{(b): total weight variation at the end of training normalized with respect to their initialization ($||\Delta \w||/||\w_0||$) vs $T$.} \textit{Inset:} $||\Delta \w||/||\w_0||$ increases with both $T$ and $P$. \textit{Main:} Plotting $||\Delta \w||/||\w_0|| P^{-\ec}$ yields a curve increasing approximately as $T^\ed$, suggesting $||\Delta \w||/||\w_0||\sim T^\ed P^{\ec}$, with $\ec\approx 0.5$ and $\ed\approx 0.8$.
    \textbf{(c): test error vs weight variation.}
    Plotting $\epsilon$ vs $||\Delta \w||/||\w_0||$ for different $P$ approximately aligns the point where $\epsilon$ starts improving.
    }
    \label{fig:CNN_lazy}
\end{figure*}

{\it Generalization error:} Fig. \ref{fig:test_phase} suggests that increasing $T$ leads to a larger test error in the lazy regime. This is evident for the CNN in Fig. \ref{fig:test_phase}-(b).
However, a detailed analysis for larger $P$ reveals that the test error for the CNN has a non-monotonic behaviour in $T$. Fig. \ref{fig:CNN_lazy}-(a) shows that increasing the number of training points, the performances of the CNN in the lazy regime, after degrading, start improving for increasing $T$. Also for the FC performances improve for increasing $T$ (Fig. \ref{fig:FC_lazy}-(a)).
In both cases, the improvement in performances corresponds to a cross-over temperature $T_c$ that changes with $P$. In fact, plotting the test error with respect to $T P^a$, with some fitting exponent $a$, aligns the point where the test error starts improving (Figs. \ref{fig:CNN_lazy}-(a), \ref{fig:FC_lazy}-(a)). This establishes the existence of a characteristic temperature $T_c$ where SGD affects performances, having an asymptotic dependence on $P$ as
\beq
\label{crit}
    T_{c} \sim P^{-a},
\eeq
with exponent values $ a \simeq 0.5$ as reported in Table \ref{tab:exponents}.

{\it Changes of weights:} To rationalize this finding, it is useful to consider how the total weight variation relative to their initialization, $\Delta w = \frac{||\w^{t^*}-\w^{0}||}{||\w^{0}||}$, increases with $T$.
In Figs. \ref{fig:FC_lazy}-(b),\ref{fig:CNN_lazy}-(b) we observe an empirical scaling 
\beq
    \Delta w \sim T^\ed P^\ec
    \label{eq:lazy-weights}
\eeq
with exponents' values $\ed\simeq 1$ (slightly lower for CNNs where $\ed\simeq 0.8, 0.9$) and $\ec\simeq 0.4$. The values are reported in Table \ref{tab:exponents}.

The dependence of the weight variations on $T$ apparent in Eq. \ref{eq:lazy-weights} suggests the following hypothesis: the characteristic temperature $T_{c}$ governing the test error corresponds to the exit from the kernel regime, which occurs when $\Delta w={\cal O}(1)$. We test this hypothesis in two ways. Firstly, if it is true then the test error plotted as a function of $\Delta w$ should be maximum at the same value of this argument, independently of the size of the training set $P$.  We confirm this result in Figs. \ref{fig:FC_lazy}-(c), \ref{fig:CNN_lazy}-(c).
Secondly, imposing that $\Delta w={\cal O}(1)$ and using Eq. \ref{eq:lazy-weights} leads to a characteristic temperature $T_c\sim  P^{-{\ec}/{\ed}}$, yielding Eq. \ref{crit} with $a=\frac{\ec}{\ed} $. This prediction is approximately verified, as shown in Table \ref{tab:exponents}.

{\it Convergence time:} We expect that a larger change of weights requires a longer training time $t^*$. We confirm that indeed the increase of $T$ in the lazy regime is accompanied by an increase of the training time $t^*$ (Fig. \ref{fig:time_lazy} in Appendix \ref{app:plots_lazy}) and we empirically find the asymptotic behaviour
\beq
    t^*\sim T P^b
    \label{eq:lazy-time}
\eeq
with values of $b$ around $1.3$ (see Table \ref{tab:exponents}).

\subsubsection{Feature regime}
The power-law behaviours of Eqs. \ref{crit}, \ref{eq:lazy-weights}, \ref{eq:lazy-time} are observed also in the feature-learning regime, with slightly different values of the exponents (see Table \ref{tab:exponents}).\\

{\it Characteristic temperature:} Unlike in the lazy limit, where $T_c$ corresponds to the transition from the linear to the non-linear regime, in the feature regime, we empirically observe that $T_c$ distinguishes between a low $T$ regime where dynamical observables such as $\Delta w$ remain unaffected by SGD noise and a high $T$ regime where the power-law behaviors of Eqs. \ref{eq:lazy-weights} and \ref{eq:lazy-time} hold. Appendix \ref{app:feature_regime} contains the data and their detailed discussion.\\
In particular, the empirical scaling relationships $\Delta w \sim P^\ei$ for $T\ll T_c$ (e.g. for FC on MNIST $\ei\approx 0.1$) and $\Delta w \sim T^\ed P^\ec$ for $T\gg T_c$ imply that $T_c\sim P^{-a}$ with an exponent satisfying $a=(\ec-\ei)/\ed$, as we observe (see Table \ref{tab:exponents}). It is worth noting that, while it is straightforward to measure $T_c$ from the behaviour of $\Delta w$, this is not always the case from the curve of the test error as a function of $T$. For instance, in the case of a CNN on CIFAR, the curves of the test error vs $T$ change shape when changing $P$ (Fig. \ref{fig:CNN_feature}-(a)). This change in shape makes it impossible to measure $T_c$ directly from these curves.

In table \ref{tab:exponents} we report the exponents $a$, $b$, $\ec$ and $\ed$ of the observations $T_c\sim P^{-a}$, $t^*\sim T P^b$ and $\Delta w\sim T^\ed P^\ec$. These are extracted from fitting the data in the Figs. \ref{fig:FC_lazy}, \ref{fig:CNN_lazy}, \ref{fig:time_lazy}, \ref{fig:FC_cifar}, \ref{fig:MNAS_mnist}, \ref{fig:simpleCNN_mnist}, \ref{fig:simpleCNN_cifar} for the lazy regime, and Figs. \ref{fig:FC_feature}, \ref{fig:FC_feature_cifar}, \ref{fig:CNN_feature} for the feature regime.
We observe that the relationships $a = \ec/\ed$ and $a=(\ec-\ei)/\delta$ are approximately verified.

\begin{table}[t]
\caption{Exponents $b$, $\ec$, $\ed$, $a$ of the empirical observations \ref{crit},\ref{eq:lazy-weights},\ref{eq:lazy-time}, including the perceptron model with data distribution parameter $\chi$. The error bar on the fit of the exponents is around $\pm 0.2$ (see Appendix \ref{app:error} for further details).
}
\label{tab:exponents}
\begin{center}
\begin{tabular}{llllll}
\multicolumn{1}{c}{\small \bf MODEL, lazy regime}  
&\multicolumn{1}{l}{$b$}
&\multicolumn{1}{l}{$\ec$}
&\multicolumn{1}{l}{$\ed$}
&\multicolumn{1}{l}{$\ec/\ed$}
&\multicolumn{1}{l}{$a$}
\\ \hline \\
FC on CIFAR & 1.4 & 0.5 & 1 & 0.5 & 0.5\\
FC on MNIST & 1.3 & 0.4 & 1 & 0.4 & 0.5\\
MNAS on CIFAR & 1.3 & 0.5 & 0.8 & 0.6 & 0.5\\
MNAS on MNIST & 1.2 & 0.3 & 0.75 & 0.4 & 0.5\\
simpleCNN on CIFAR & 1.5 & 0.6 & 0.9 & 0.67 & 0.6\\
simpleCNN on MNIST & 1.4 & 0.35 & 0.9 & 0.45 & 0.5\\
perceptron $\chi=1.5$  & 1.8 & 0.4 & 1 & 0.4 & \\
perceptron $\chi=4$  & 1.4 & 0.2 & 1 & 0.2 & \\[.5cm]
\multicolumn{1}{c}{\small \bf MODEL, feature regime}  
&\multicolumn{1}{c}{$b$}
&\multicolumn{1}{c}{$\ec$}
&\multicolumn{1}{c}{$\ed$}
&\multicolumn{1}{c}{$\frac{\ec-\ei}{\ed}$}
&\multicolumn{1}{c}{$a$}
\\ \hline \\
FC on CIFAR & 1.4 & 0.6 & 0.5 & 0.9 & 0.9\\
FC on MNIST & 1.4 & 0.45 & 0.5 & 0.7 & 0.7\\
MNAS on CIFAR & 1.3 & 0.5 & 0.6 & 0.5 & 0.5\\
\end{tabular}
\end{center}
\end{table}

\section{Interpretation of the observations} 
\label{sec:toy}

In this section we provide an understanding for Eq. \ref{eq:lazy-weights}, which justifies Eqs. \ref{crit} and \ref{eq:lazy-time}, based on the local alignment of the model decision boundary with the true one. We then test it in the perceptron model, where relevant quantities can be easily measured.

\subsection{Neural networks}
\label{sec:interpretation}
\begin{figure}
\centering
\def\svgwidth{1\columnwidth}
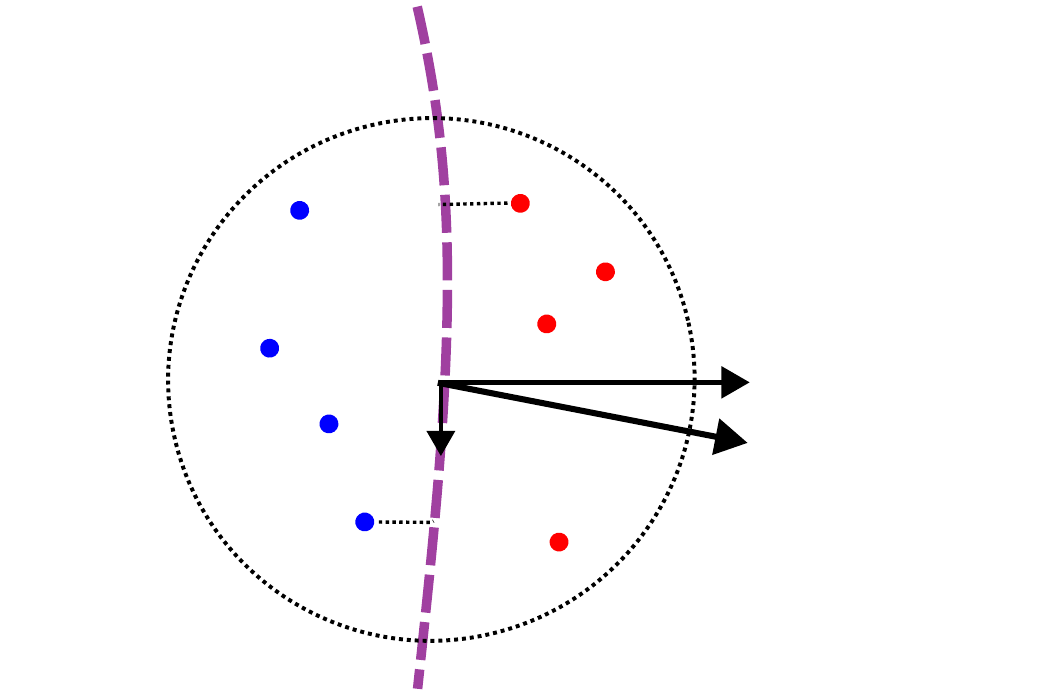
\caption{\textbf{Pictorial representation of a neighbourhood $B_\epsilon$ of the true decision boundary (purple dashed line).} Red (blue) dots are training points with labels $+1$ ($-1$) and the point $\x^{+}$ ($\x^-$) is the closest to the true decision boundary. The decision boundary of the trained model $F(\x)$ corresponds to the $\x$'s such that $F(\x)=0$ (black line). The gradients $\partial_\x F$ on it quantify the local alignment between the model boundary and the true one: $\partial_{\x} F_\parallel $ is the component in the direction of correct alignment, while $\partial_{\x} F_{\perp}$ is orthogonal to it.}
\label{fig:scheme}
\end{figure}

\paragraph{Local alignment of decision boundaries.}
In binary classification, the true decision boundary in data space is the locus of points between $\x$'s with different labels $y(\x)=\pm 1$, while the decision boundary learnt by the model $F(\x)$ corresponds to the $\x$'s such that $F(\x)=0$.
Considering a point $\x^*$ where the two boundaries cross and its neighbourhood $B_{\epsilon}$ of diameter $\epsilon$, the local alignment of the model boundary with the true one is given by
\beq
\frac{|| \partial_\x F_\parallel ||}{||\partial_\x F_{\perp}||}
\eeq
at linear order in $\epsilon$, where $\partial_\x F_\parallel $ is the component of the gradient $\partial_\x F(\x^*)$ in the direction perpendicular to the true decision boundary, while $\partial_\x F_\perp = \partial_\x F(\x^*) - \partial_\x F_\parallel $ is orthogonal to it (see Fig. \ref{fig:scheme}).
The angle between the two boundaries corresponds to $\theta = \text{arccot}\lpa\frac{||\partial_\x F_\parallel ||}{||\partial_\x F_\perp||}\rpa$ and perfect learning requires that $\frac{||\partial_\x F_\parallel ||}{||\partial_\x F_\perp||}\rightarrow\infty$.\\
$\partial_\x F_\parallel $ identifies the direction that is informative for the task, while $\partial_\x F_\perp$ is the component in the non-informative directions, which act as noise.
It is worth noting that in the lazy regime, the gradient components $\partial_\x F_\parallel$ and $\partial_\x F_\perp$ are linear functions of the variation of the weights, as recalled in \footnotemark{}. This fact allows defining informative and uninformative weight components $\w_\parallel$ and $\w_\perp$, respectively, around a data point $\x^*$. The condition we obtain below on the magnitude of $\|\partial_\x F_\parallel\| / \|\partial_\x F_\perp\|$ to fit the data thus corresponds to a bound on  $\| \w_\parallel\|/ \|\w_\perp\|$, as shown in Section \ref{sec:perceptron_problem} using the example of the perceptron.
\footnotetext{The predictor defined in Sec. \ref{sec:definition} $F(\w,\x) = f(\w^t,\x_\mu) - f(\w^0,\x)$, at linear order in the weight variation $\Delta \w$, reads $F(\w, \x) = \nabla_\w f(\w^0,\x) \cdot \Delta \w$. Therefore $\partial_\x F(\w, \x^*) = \bm{\mathcal{T}}\Delta \w$ with the tensor $\mathcal{T}_{ik} = \partial_{x_i}\nabla_{w_k} f(\w^0,\x^*)$. Performing a projection of $\bm{\mathcal{T}}$ onto the informative and uninformative directions in data space, $\bm{\mathcal{T}} = \bm{\mathcal{T}}^\parallel + \bm{\mathcal{T}}^\perp$, we obtain $\partial_\x F_\parallel = \bm{\mathcal{T}}^\parallel\Delta \w \equiv \w_\parallel$ and $\partial_\x F_\perp = \bm{\mathcal{T}}^\perp\Delta \w \equiv \w_\perp$ which corresponds to different components of the weight variation. Therefore Eq. \ref{eq:fit_ineq3} becomes a condition, dependent on $\x^*$, on the components of the weights: $\frac{\|\w_\parallel \|}{\|\w_\perp\|} \geq \frac{1}{\delta_\parallel } \lpa \frac{2 \alpha^{-1}}{\|\w_\perp\|} + c \rpa.$}

\paragraph{Fitting condition.}
When considering the hinge loss in Eq. \ref{eq:hingeLoss} with margin $\alpha^{-1}$ defined in Sec. \ref{sec:definition}, a training point $(\x^\mu, y^\mu)$ is fitted (i.e. it has zero training loss) when
$y^{\mu}\ F(\x^\mu)\geq\alpha^{-1}$.
Having $P$ training points, we call $\x^{\pm}$ the two of them in $B_\epsilon$ with $y(\x^{\pm})=\pm 1$ that have the shortest distances $\delta^{\pm}$ from the true decision boundary. Their fitting conditions $\pm F(\x^{\pm})\geq \alpha^{-1}$ imply
$F(\x^{+}) - F(\x^{-}) \geq 2\alpha^{-1}$.
Assuming $F(\x)$ is differentiable in $B_\epsilon$, the last inequality can be approximated at linear order in $\epsilon$ as
\beq
    \partial_\x F(\x^{*}) \cdot \lpa \x^+ - \x^{-}\rpa \geq 2\alpha^{-1}.
\label{eq:fit_ineq2}
\eeq
Defining $\delta_\parallel$ and $c$ as $\delta_\parallel  = \delta^+ + \delta^-= \frac{\partial_\x F_\parallel}{||\partial_\x F_\parallel ||} \cdot \lpa \x^+ - \x^{-}\rpa$ and $c = -\frac{\partial_\x F_\perp}{||\partial_\x F_{\perp}||} \cdot \lpa \x^+ - \x^{-}\rpa$,
inequality \ref{eq:fit_ineq2} becomes
\beq
    \frac{||\partial_\x F_\parallel ||}{||\partial_\x F_\perp||} \geq \frac{1}{\delta_\parallel } \lpa \frac{2 \alpha^{-1}}{||\partial_\x F_\perp||} + c \rpa.
\label{eq:fit_ineq3}
\eeq

\paragraph{Role of the training set size $P$ and of the SGD temperature $T$.}
Considering Eq. \ref{eq:fit_ineq3}:
\begin{itemize}    
    \item[(1)] we argue that increasing $P$ corresponds to shorter distances $\delta_{\parallel}$, which require a better alignment of the model decision boundary with the true one, that is a larger $\frac{||\partial_\x F_\parallel ||}{||\partial_\x F_\perp||}$.
    \item[(2)] Since increasing $T$ makes the training dynamics more noisy, we propose that a larger $T$ increases the non-informative component $||\partial_\x F_\perp||$. This implies, according to Eq. \ref{eq:fit_ineq3}, a larger informative component $||\partial_\x F_\parallel ||$ to fit the training set.
\end{itemize}

\begin{figure*}
    \centering
    \includegraphics[width=.31\textwidth]{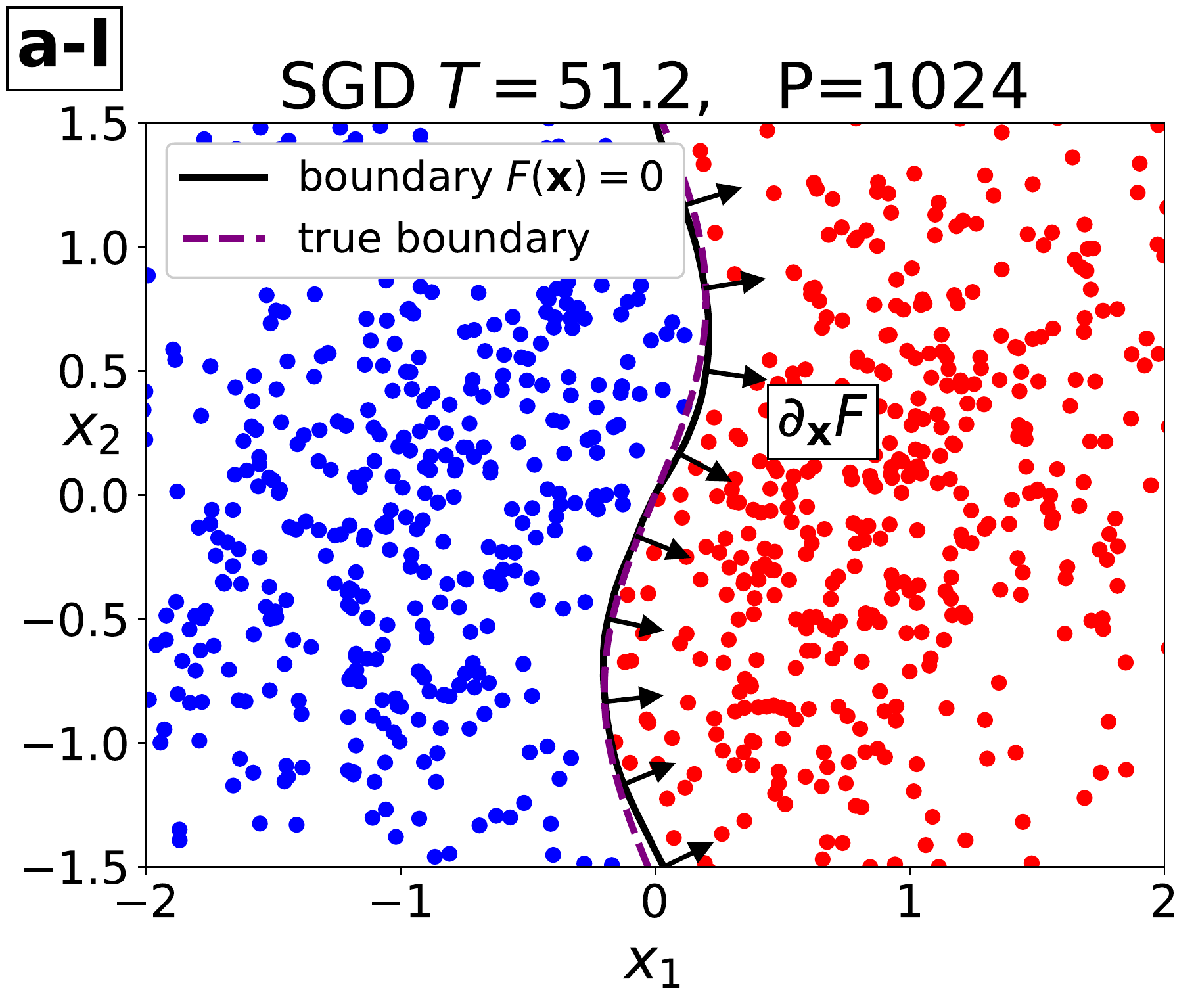}
    \includegraphics[width=.31\textwidth]{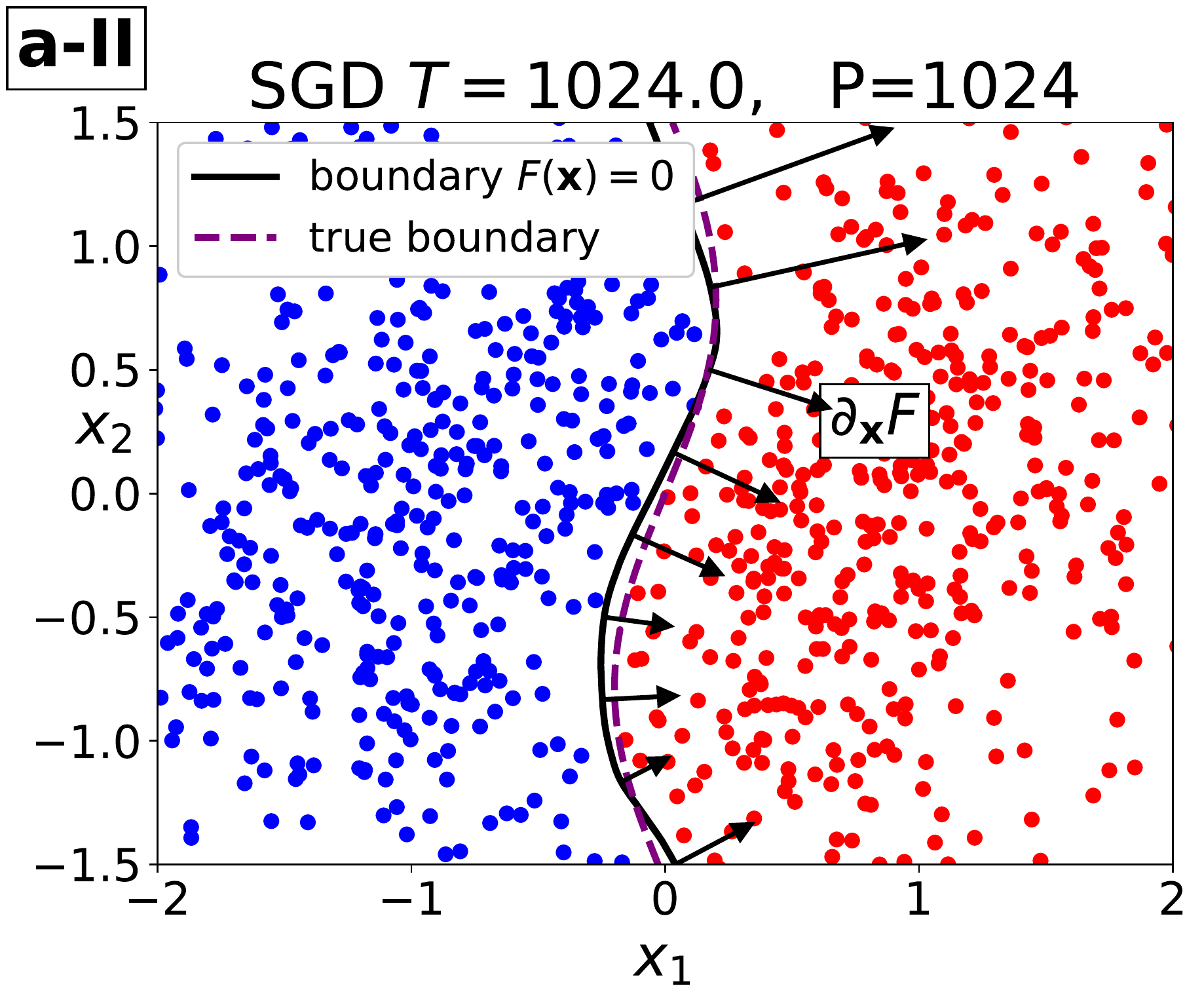}
    \includegraphics[width=.31\textwidth]{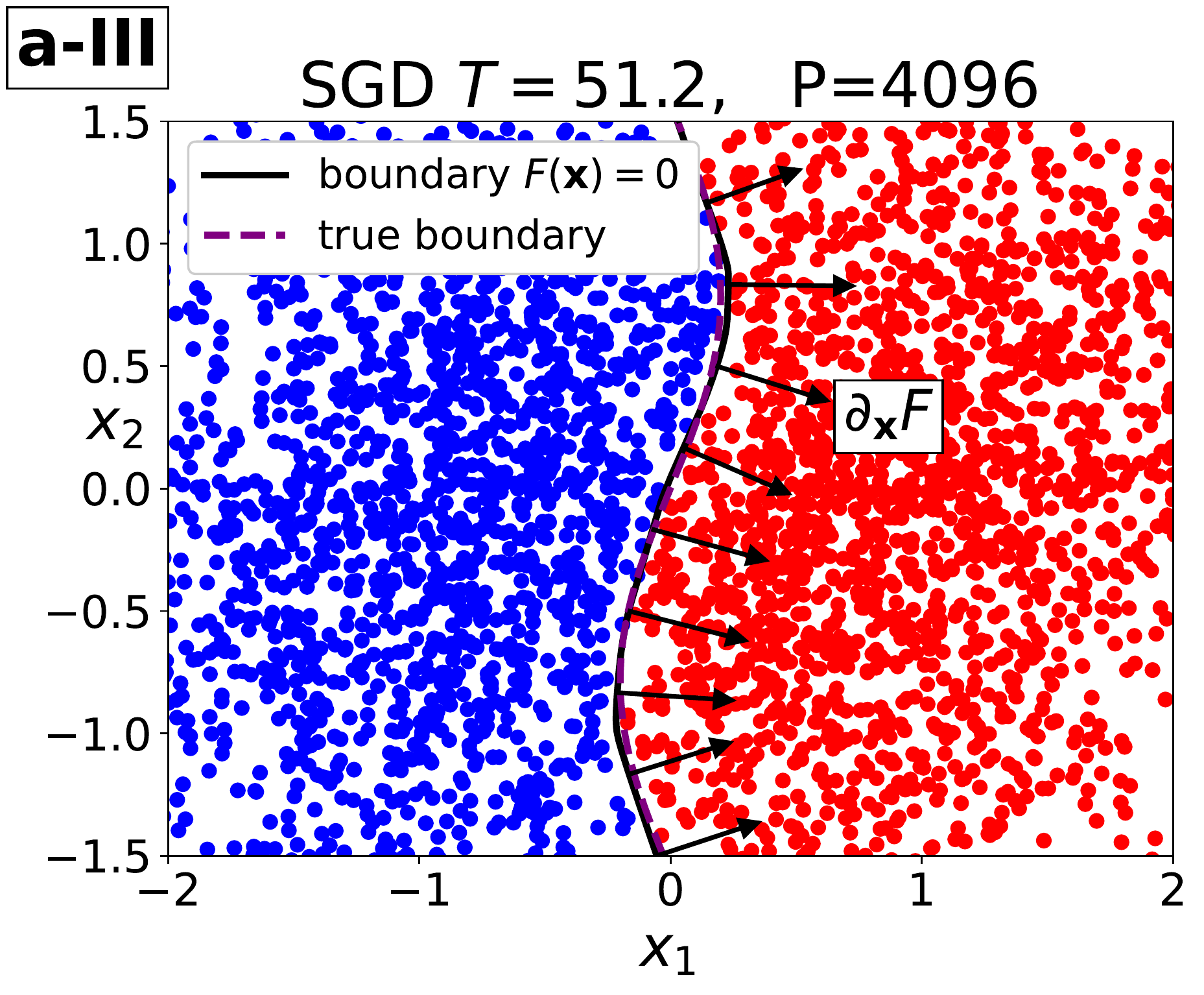}
    \includegraphics[width=.31\textwidth]{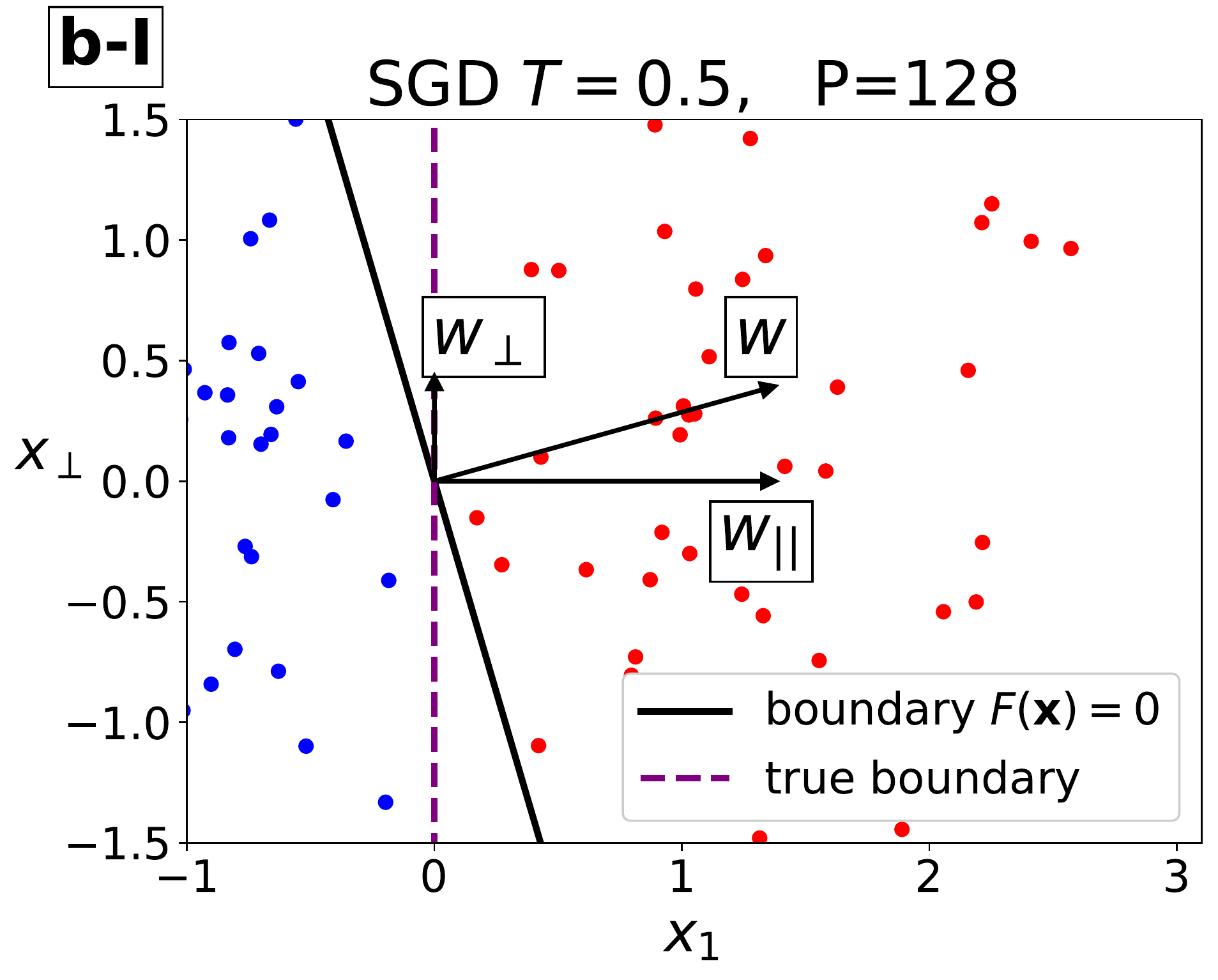}
    \includegraphics[width=.31\textwidth]{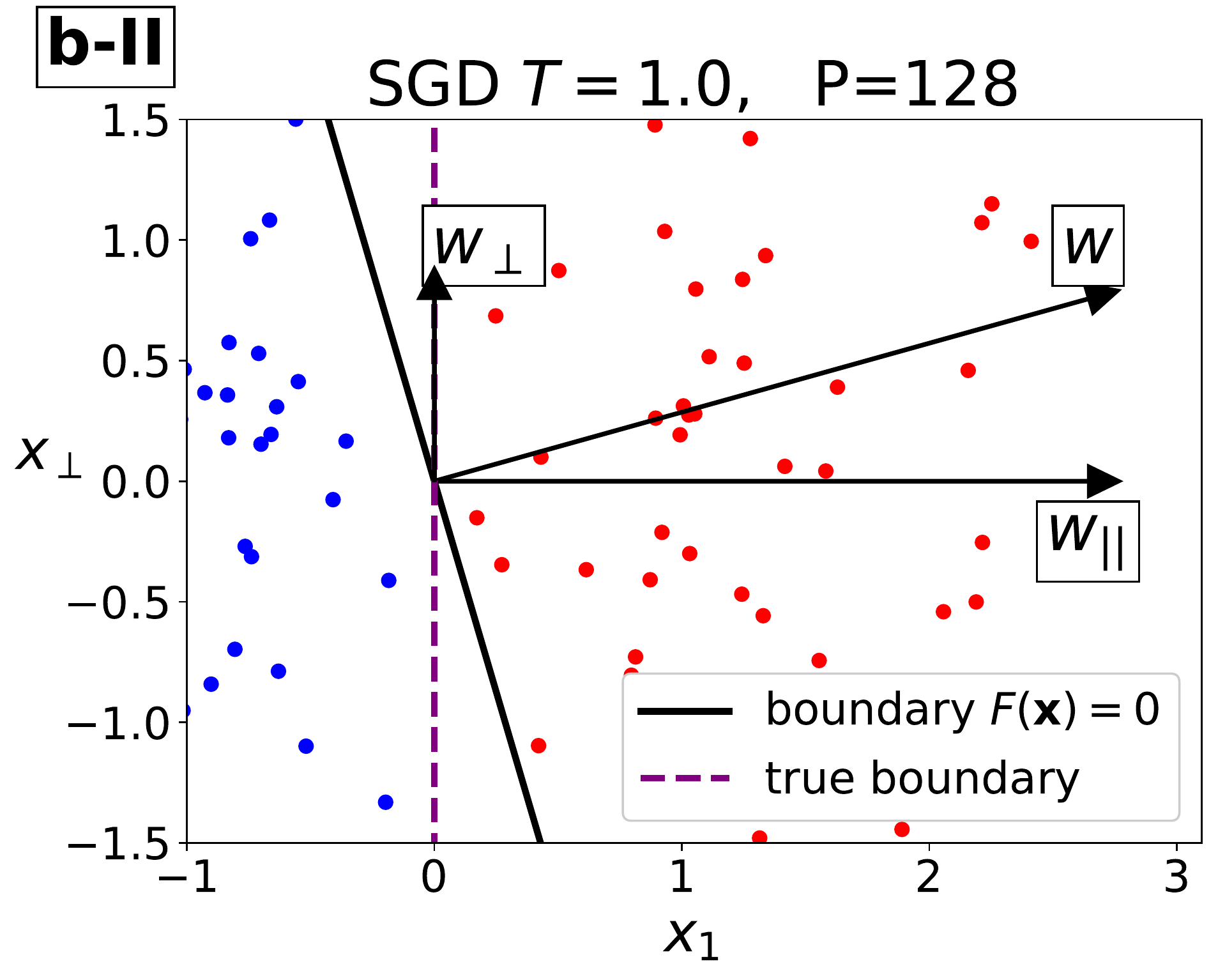}
    \includegraphics[width=.31\textwidth]{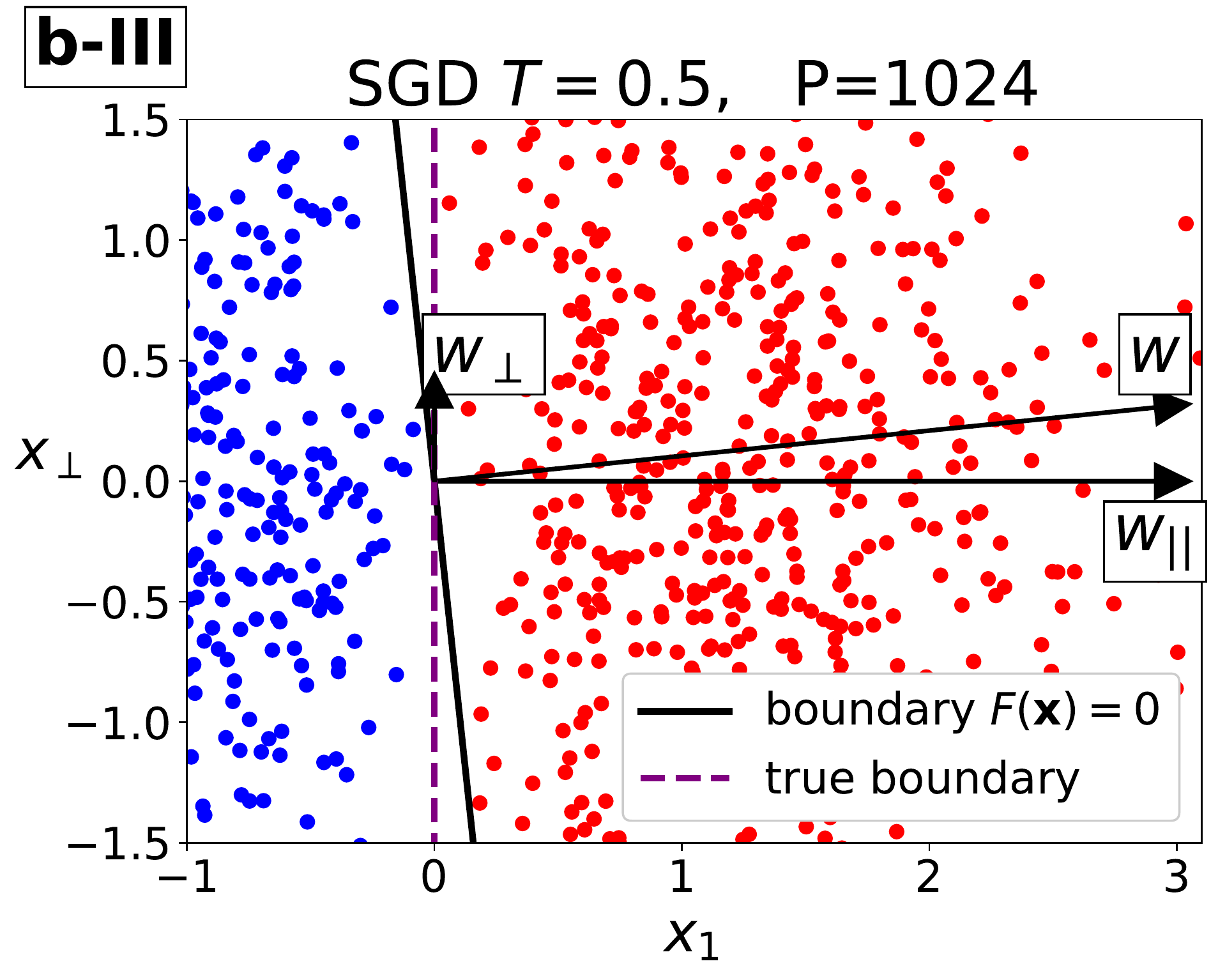}
    \vspace{-.5 cm}
    \caption{\textbf{Decision boundary for binary classification in 2 dimensions: (a) one-hidden-layer FC neural network; (b) perceptron model.} Red (blue) dots are training points with label $+1$ ($-1$) and the purple dashed line is the true decision boundary. The black line is the decision boundary obtained from training the model $F(\x)$ with SGD.
    \textbf{(I)-(II).} Increasing the SGD temperature $T$ gives larger gradients $\partial_\x F$ but not a better alignment between the decision boundaries: it increases the non-informative component ($\w_\perp$ for the perceptron).
    \textbf{(I)-(III).} Increasing the number of training points $P$ gives larger gradients $\partial_\x F$ and a better alignment between the decision boundaries.
    }
    \label{fig:boundary_gradients}
\end{figure*}

According to (1) and (2), both $T$ and $P$ increase the gradients magnitude $||\partial_\x F(\x^*)||$, but only increasing $P$ gives a better boundary alignment, that is a larger $||\partial_\x F_\parallel||/||\partial_\x F_\perp||$. This effect is illustrated in Fig. \ref{fig:boundary_gradients} for two-dimensional data.

\begin{figure}
    \centering
    \includegraphics[width=.8\columnwidth]{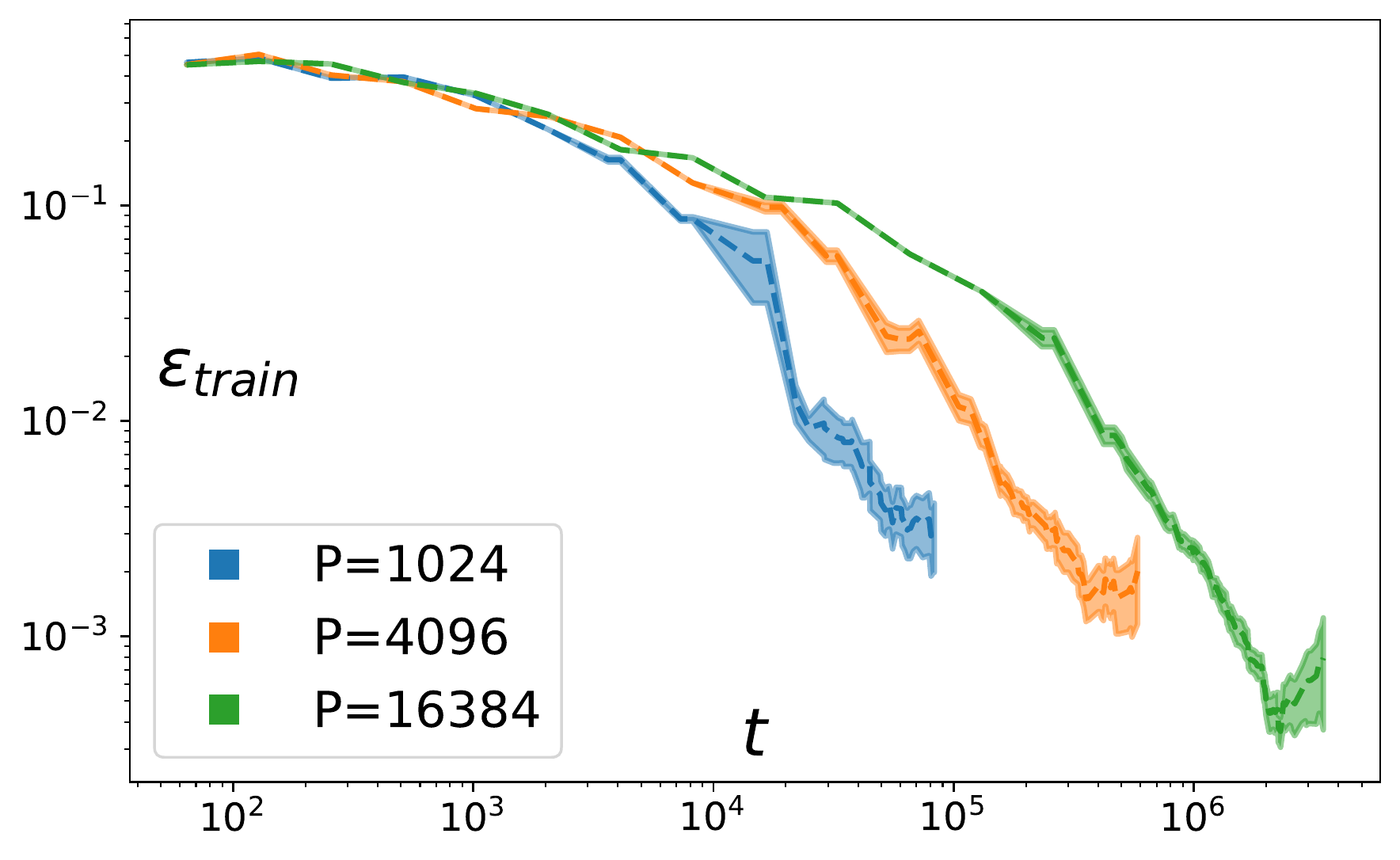}
    \vspace{-.5cm}
    \caption{\textbf{FC on MNIST: training error in time, fixed $T$, changing $P$.} Increasing the training set size $P$ delays the point when the training error goes to zero, while the first part of the dynamics stays unchanged.}
    \label{fig:dynamics_TP}
\end{figure}

Overall, both increasing $P$ and $T$ require larger gradient magnitudes $||\partial_\x F(\x^*)||$ to fit the training set, which corresponds to a larger relative variation of the weights, in accordance with the observation of Eq. \ref{eq:lazy-weights}. This larger growth of the weights requires a longer training time, in accordance with the observation of Eq. \ref{eq:lazy-time}.
In this view, a key effect of increasing $P$ is to diminish the distance between data of different labels, which are the last points to be fitted. We thus expect that changing $P$ affects the dynamics only late in training, as we demonstrate in Fig. \ref{fig:dynamics_TP}. Therefore, the hardest data to fit affect both the growth of the weights and  the training time.

\subsection{Perceptron model}
\label{sec:perceptron_problem}

We consider a linearly-separable classification task with high-dimensional data $\x \in \R^d$, $d\gg 1$, with labels $y(\x)=\pm 1$ given by the signs of the first components: 
\beq
y(\x) = \text{sign}(x_{1}).
\eeq
The true decision boundary in this problem is the hyper-plane $x_1=0$.
We study this problem with a linear classifier, called perceptron: 
\beq
    F(\w,\x) = \frac{1}{\sqrt{d}} \w \cdot \x
\eeq
initialized with $\w^0=0$.\\
Although the perceptron is always in the lazy regime\footnote{Because it is linear with respect to the weights $\w$.} and does not have a characteristic temperature of SGD controlling performance, it is of interest because the interpretation discussed in Sec. \ref{sec:interpretation} can be tested. In fact, the gradient $\partial_\x F(\x^*)$ corresponds to the perceptron's weights $\w/\sqrt{d}$, with the informative and non-informative components respectively $||\partial_\x F_\parallel|| = w_1/\sqrt{d}$ and $||\partial_\x F_\perp|| = ||\w_\perp||/\sqrt{d}$. The alignment of the perceptron decision boundary with the true one is given by the ratio 
\beq
    w_1/||\w_\perp||.
\eeq

The fitting condition on the data point $(\x^\mu, y^\mu)$ requires that the weights $\w = [w_1; \w_\perp]$ satisfy
\beq    
    w_1 |x^{\mu}_1| + y^{\mu} \w_\perp \cdot \x^{\mu}_\perp \geq \frac{\sqrt{d}}{\alpha}
    \label{eq:sat}
\eeq
which, by defining the random quantities $c_\mu = -y^{\mu} \frac{\w_\perp}{||\w_\perp||} \cdot \x^{\mu}_\perp$, can be recast as 
\beq    
    \frac{w_1}{||\w_\perp||} \geq \frac{1}{|x^{\mu}_1|} \lpa\frac{\sqrt{d}}{\alpha||\w_\perp||} + c_{\mu}\rpa.
    \label{eq:sat1}
\eeq
This relationship is a special case of Eq. \ref{eq:fit_ineq3}.
In fact, increasing $P$ gives smaller values of $|x^\mu_1|$ which require larger $\frac{w_1}{||\w_\perp||}$ to fit the training set, while increasing $T$ corresponds to increasing $||\w_\perp||$. A qualitative confirmation of this effect is reported in Fig. \ref{fig:boundary_gradients}-(b).\\
In the following, we consider the regime of large $T$ and large $\alpha$, corresponding to $\frac{\sqrt{d}}{\alpha||\w_\perp||}\ll |c_\mu|$, for which condition \ref{eq:sat1} becomes 
\beq
\frac{w_1}{||\w_\perp||}\geq \frac{c_{\mu}}{|x^{\mu}_1|} \lpa 1+ o(1)\rpa.
\label{eq:sat2}
\eeq

\begin{figure}
    \centering
    \includegraphics[width=1\columnwidth]{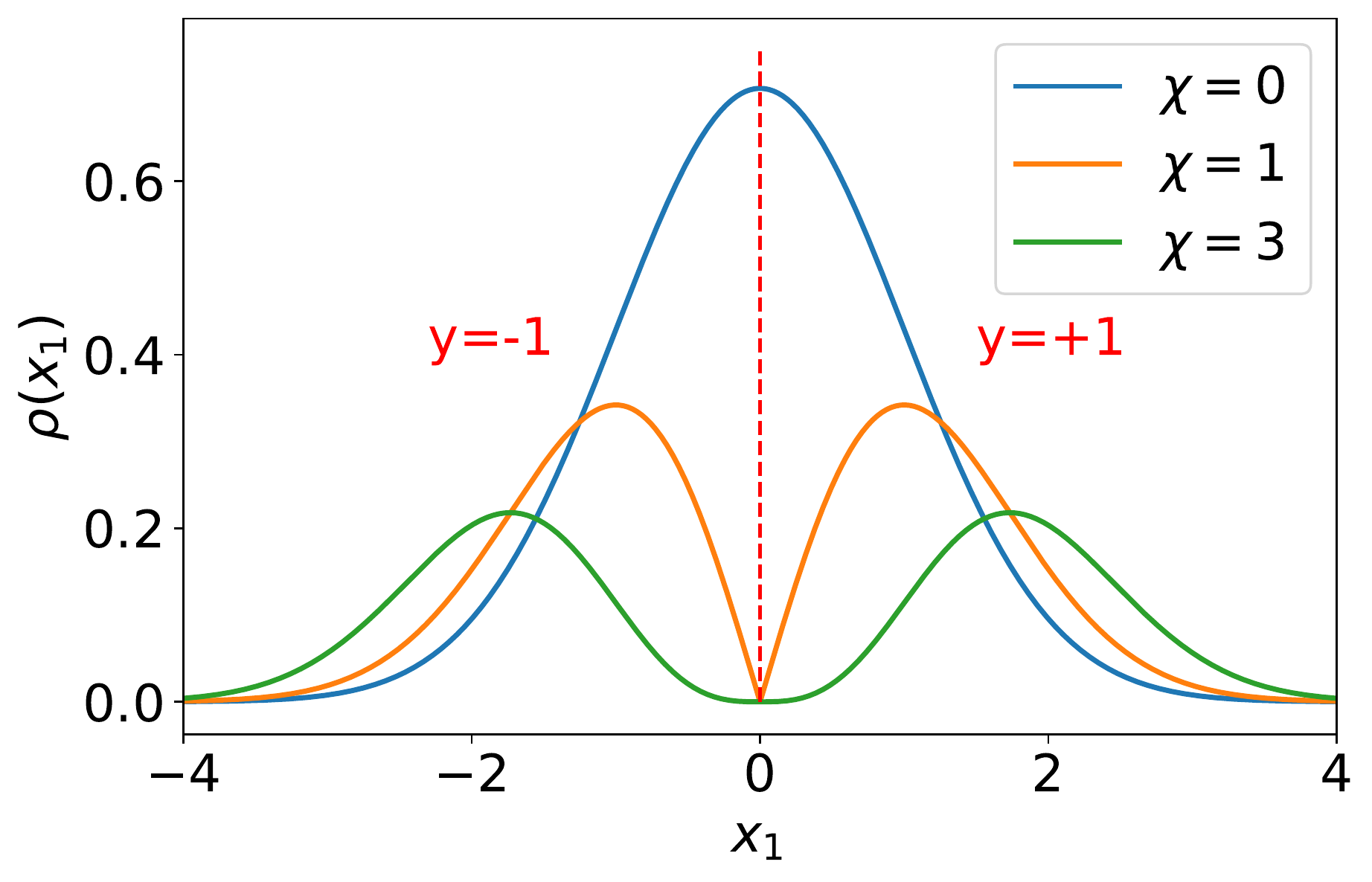}
    \vspace{-.8cm}
    \caption{\textbf{Perceptron model, data distribution on the $x_1$ component.} The sign of $x_1$ determines the class $y=\text{sign}(x_1)$. For $\chi=0$ the distribution is Gaussian.}
    \label{fig:perceptron_data}
\end{figure}

\paragraph{Data distribution and setting.}
To control the density of data near the decision boundary $x_1=0$, we consider a distribution on the first component $x_1$ parametrized by $\chi \geq 0$ (Fig. \ref{fig:perceptron_data}):
\beq
\rho(x_1) = |x_1|^\chi e^{-x_1^2/2} / Z,
\label{eq:rho_x1}
\eeq
with $Z=2^{\frac{1+\chi}{2}}\Gamma(\frac{1+\chi}{2})$ the normalization constant. The other $d-1$ components $\x_\perp = [x_i]_{i=2,...,d}$ are distributed as standard multivariate Gaussian numbers, i.e. $\x_\perp \sim \mathcal{N}({\bm 0}, \I_{d-1})$.
$\chi=0$ corresponds to the Gaussian case. This data distribution has been first considered in \citet{tomasini2022failure}.
The learning setting is defined identically to the one of neural networks in Sec. \ref{sec:definition}.
We consider the case $1 \ll d \ll P$,
where $d$ is the dimension of the data and the perceptron weights and $P$ is the number of training points. We consider this being a realistic limit when considering the effective dimension $d_{\text{eff}}$ of real datasets ($d_{\text{eff}}\approx 15$ for MNIST and $d_{\text{eff}}\approx 35$ for CIFAR-10 \citep{spigler2020asymptotic}) with respect to the number of training samples $P>10^3$.

\begin{figure*}
    \centering
    \includegraphics[width=.31\textwidth]{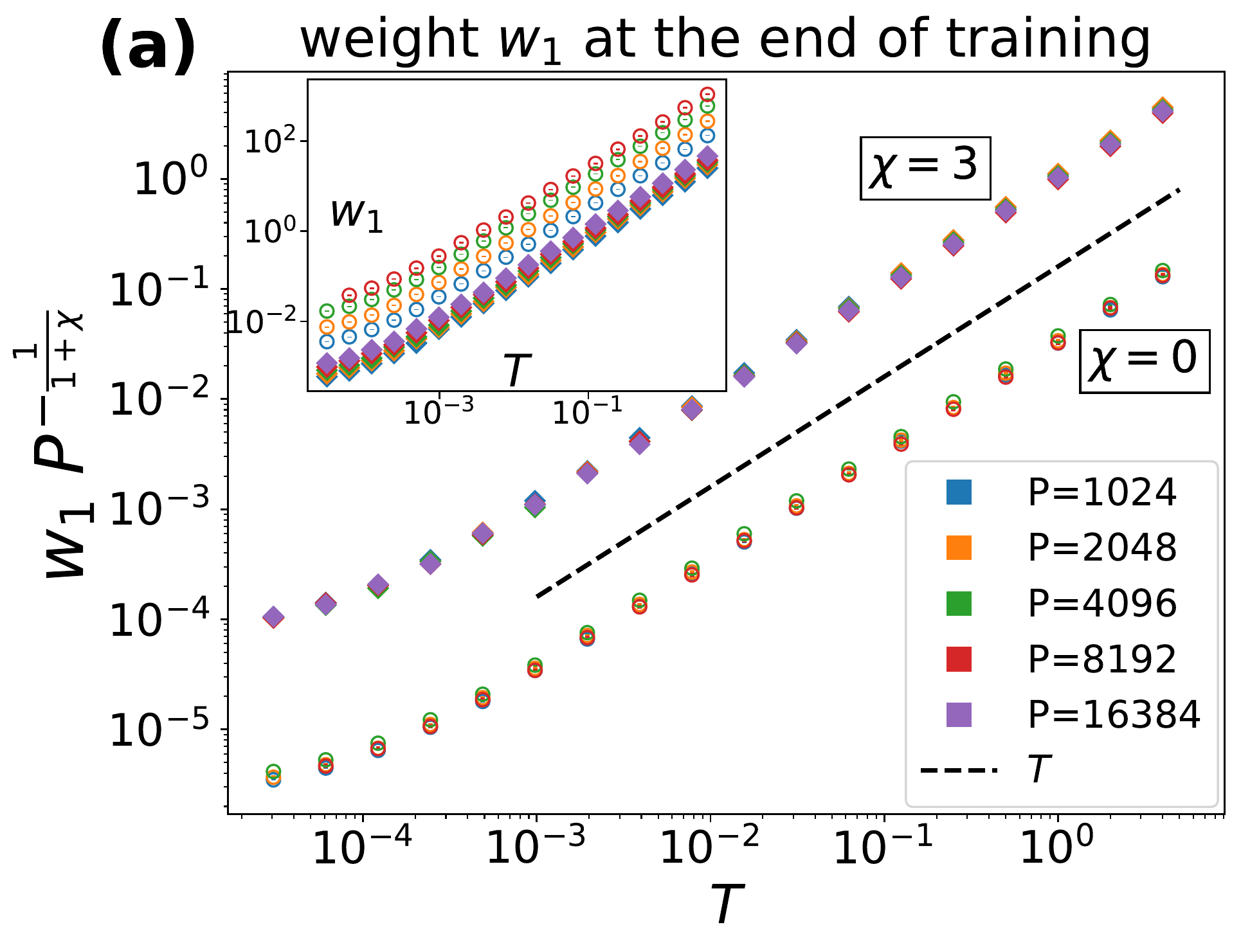}
    \includegraphics[width=.31\textwidth]{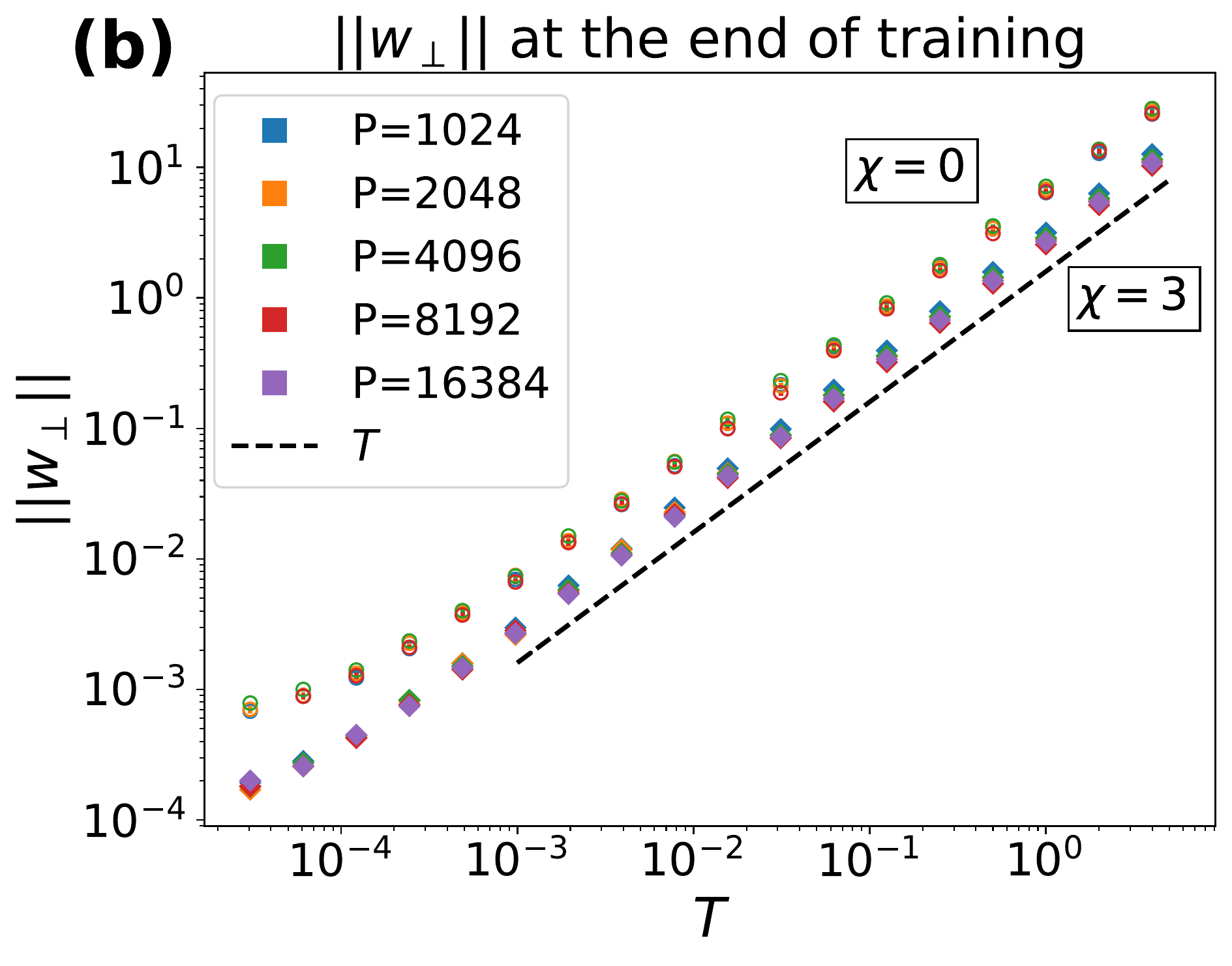}
    \includegraphics[width=.31\textwidth]{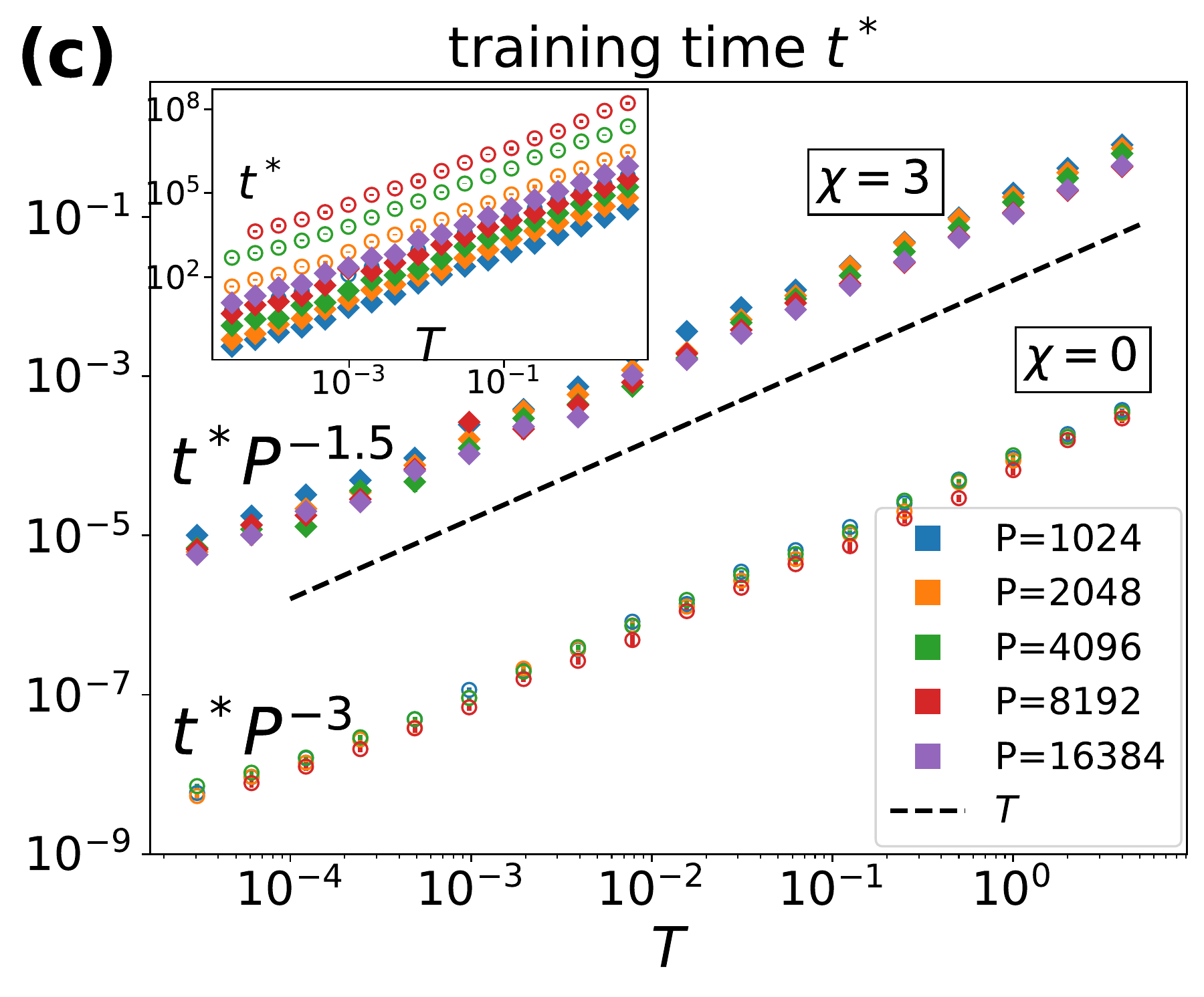}
    \vspace{-.5cm}
    \caption{
    \textbf{Perceptron model, $d=128$, $B=2$, varying $T$ and $P$.}
    \textbf{(a)} \textit{Inset:} Total variation of the weight $w_1$ at the end of training with respect to SGD noise $T$ and training set size $P$ (colors), for different data distributions $\chi=0$ (empty circles) and $\chi=3$ (full diamonds). \textit{Main:} Plotting $w_1 P^{-\frac{1}{1+\chi}}$ gives a curve proportional to $T$ for each value of $\chi$, revealing the asymptotic behaviour $w_1\sim T P^{\ec}$ (Eq. \ref{eq:lazy-weights} for neural networks) with a data-dependent exponent $\ec=\frac{1}{1+\chi}$ in accordance with prediction \ref{eq:gamma_chi}.
    \textbf{(b)} Total variation of $||w_\perp||$ for the same setting of panel (a). $||w_\perp||$ is proportional to $T$ independently of $P$, as stated in Eq. \ref{eq:wp_T}.
    \textbf{(c)} \textit{Inset:} Total training time $t^*$ for the same setting as panel (a): $t^*$ increases with both $T$ and $P$. \textit{Main:} Plotting $t^* P^{-b}$, with $b$ depending on $\chi$, gives approximately one curve proportional to $T$ for each value of $\chi$, corresponding to the asymptotic behaviour $t^*\sim T P^{b}$ as found for neural networks (Eq. \ref{eq:lazy-time}).
    }
    \label{fig:perceptron_scheme}
\end{figure*}

\textbf{Empirical observations.} A key result is that the perceptron displays asymptotic behaviours in the change of weights and training time similar to those of neural networks.
For the considered perceptron initialized with $\w^0=0$, the weight variation $\Delta w$ corresponds to $||\w||$. Since $w_1/||\w_\perp||\gg 1$ for large $P$, we have $\Delta w = ||\w||\simeq w_1$.
Eqs. \ref{eq:lazy-weights} and \ref{eq:lazy-time} are verified with exponents reported in Table \ref{tab:exponents}, as shown in Fig. \ref{fig:perceptron_scheme}-(a,c). These data are produced with $d=128$, therefore in a high-dimensional setting.\\
In addition, we observe that $||\w_\perp||$ at the end of training is proportional to $T$ and independent of $P$ (Fig. \ref{fig:perceptron_scheme}-(b)):
\beq
    ||\w_\perp||\sim T.
\label{eq:wp_T}
\eeq
This observation is a positive test about the effect of $T$ on $||\partial_\x F_\perp||$ proposed in Sec. \ref{sec:interpretation}.

\paragraph{Non-universality of the exponents.} Remarkably, the exponents $\ec$ and $b$ of $P$ for the perceptron depend on the parameter $\chi$ of the data distribution. This finding can be rationalized by considering condition \ref{eq:sat2} at the end of training. In fact, satisfying \ref{eq:sat2} for every training point requires $\frac{w_1}{||\w_\perp||}\geq \underset{\mu}{\text{max}} \frac{c_{\mu}}{|x^{\mu}_1|}$. In Appendix \ref{app:max}, classical extreme value theory is used to show that, for large $P$, the typical value of $\underset{\mu}{\text{max}} \frac{c_{\mu}}{|x^{\mu}_1|}$ behaves asymptotically as $\langle \underset{\mu}{\text{max}} \frac{c_{\mu}}{|x^{\mu}_1|} \rangle = C P^{\frac{1}{1+\chi}} + o\lpa P^{\frac{1}{1+\chi}}\rpa$ for some constant $C$. Therefore we obtain a prediction for the exponent $\ec$:
\beq
    \ec = \frac{1}{1+\chi},
\label{eq:gamma_chi}
\eeq
in excellent agreement with data (Fig \ref{fig:perceptron_scheme}-(a)). This further confirms that the asymptotic behaviour with respect to $P$ is controlled by the statistics of the points close to the decision boundary. Thus the exponents are non-universal, since they depend directly on the data distribution.\\
An estimate of the parameter $\chi$ for some images datasets is reported in \citet{tomasini2022failure} through the study of kernel ridge regression. For binary CIFAR10, $\chi_{CIFAR}=1.5$ is reported, that according to \ref{eq:gamma_chi} corresponds to $\ec = 0.4$, a value compatible with those observed in neural networks (Table \ref{tab:exponents}).

\section{Conclusions}
\label{sec:discussion}
In this work we have explored the effect of SGD noise in different training regimes of neural networks using the hinge loss, which is analogous to the widely used cross-entropy loss and performing early-stopping. Since the hinge loss goes to zero at the end of training, the minima found by the algorithm are always flat: a static view explaining the benefit of SGD in terms of the flatness of minima cannot be applied. Instead, we propose a dynamical view where SGD noise increases the weights of the model in directions that are detrimental for learning, which in turn induces an increase in the useful directions to fit the training set. 
Fitting is the hardest for data close to the decision boundary, whose statistics depends both  on the size of the training set and the distribution of data close to the decision boundary. This view naturally explained  our observations that the total weight variation, and  the training time,  depend on both the SGD noise and the size of the training set. It also rationalizes the puzzling observation that the characteristic SGD temperature for which weight changes become significant and the test error is affected by the noise depends on the training set size. Exponents characterizing this relationship are non-universal. We expect them to depend on the data distribution near the decision boundary, as we demonstrated for the perceptron.\\

Our work thus clarifies a key effect of SGD, and explains the range of temperatures where SGD noise matters. 
However, understanding the sign of the effect of this noise on performance (beneficial or detrimental), and how it relates to the data structure and the network architecture, appears to be a particularly vexing question. 
For example, for the lazy regime of CNNs, we observe a non-monotonic behaviour of the test error, which initially grows and then decays as the SGD noise is increased.
What determines this behavior is an open question that requires further investigation.

\section*{Acknowledgments}
We thank Francesco Cagnetta, Alessandro Favero, Bastien Olivier Marie Göransson, Leonardo Petrini and Umberto Maria Tomasini for helpful discussions. This work was supported by a grant from the Simons Foundation (\# 454953 Matthieu Wyart).

\newpage
\bibliography{bibliography}
\bibliographystyle{icml2023}

%%%%%%%%%%%%%%%%%%%%%%%%%%%%%%%%%%%%%%%%%%%%%%%%%%%%%%%%%%%%%%%%%%%%%%%%%%%%%%%
%%%%%%%%%%%%%%%%%%%%%%%%%%%%%%%%%%%%%%%%%%%%%%%%%%%%%%%%%%%%%%%%%%%%%%%%%%%%%%%
% APPENDIX
%%%%%%%%%%%%%%%%%%%%%%%%%%%%%%%%%%%%%%%%%%%%%%%%%%%%%%%%%%%%%%%%%%%%%%%%%%%%%%%
%%%%%%%%%%%%%%%%%%%%%%%%%%%%%%%%%%%%%%%%%%%%%%%%%%%%%%%%%%%%%%%%%%%%%%%%%%%%%%%
\newpage
\appendix
\onecolumn

\section{Other related works}
\label{app:other_works}
As reviewed in the introduction, various works have studied empirically the role of SGD noise on performance. Our work goes beyond these studies by systematically studying the role of initialization scale and size of the training set for a large range of noise magnitude.
Some recent studies have analysed the relationship between the implicit bias of SGD and the initialization scale in simple regression models \citep{haochen2021,flammarion2021}, showing that SGD bias the model towards the feature-learning regime. Our work tests this hypothesis for image classification, showing that the effect is not captured by a simple reduction of the initialization scale, but confirming that SGD stochasticity can bring the model outside the kernel regime.
Several works have showed that larger SGD stochasticity leads to flatter minima of the loss landscape and it has been argued that this leads to improved performances \citep{hochreiter1997flat, keskar2016, zhang2018energy, smith2018bayesian, wu2018sgd}. Our results show that in some regimes performances can behave non-monotonically with respect to increasing SGD stochasticity, in contradiction with simple arguments based on the flatness of the landscape. Therefore our observations call for a theory of generalization that goes beyond the flatness view and explains at least the sign of change in performances.\\
The importance of the stopping criterion when evaluating the performances of SGD has already been emphasized \citep{hoffer2017, shallue2018, smith2020}. In this work we remove the ambiguity in the choice of the computational budget and show the effect of the size of the training set on the time needed to reach convergence. Moreover, we show how the size of the training set affects the noise scale at which we observe a change in performances. To the best of our knowledge, this relationship constitutes a novelty in the literature.\\
Previous works have showed that the noise scale of SGD is controlled by the ratio between the learning rate and the batch size when the batch is smaller than some cross-over value \citep{jastrzkebski2017, shallue2018, smith2020}. On the theoretical side, a description of SGD based on a continuous-time stochastic differential equation (SDE) driven by Gaussian noise was derived \citep{li2017stochastic, li2019stochastic}. In our work, we consider SGD in the ``small batch regime'' where we can describe its noise magnitude by the ratio between learning rate and batch size. 
% Moreover, we consider the SDE description in the feature regime as a starting point for a scaling argument between the noise and the initialization scales.

\section{Scaling argument for the $\alpha$ dependence of the characteristic temperatures in the feature regime}
\label{app:scaling_feature}

The covariance of the mini-batch gradients when $B\ll P$ is given by $\cov\lpa \w\rpa/B$ \citep{chaudhari2018}, with
\beq
\begin{aligned}
    \cov\lpa \w\rpa =
    \frac{1}{P}\sum_{\mu=1}^P \theta_\mu \ 
    \nabla_{\w} f(\w,\x_\mu) \otimes \nabla_{\w} f(\w,\x_\mu)
    - \nabla_\w L(\w)\otimes \nabla_\w L(\w),
\end{aligned}
\label{eq:general_cov}
\eeq
where $\theta_{\mu}=\theta\lpa\alpha^{-1}-y_{\mu} F(\w,\x_{\mu})\rpa$. The stochastic differential equation (SDE) matching the first two moments of the SGD update \ref{eq:SGD} corresponds to \citep{smith2020, zhang2019}:
\beq
d\w^t = - dt \nabla L(\w^t) + \sqrt{T} \sqrt{\cov\lpa \w^t\rpa} d\bm{W}^t
\label{eq:general_SDE}
\eeq
where $\bm{W}^t$ is Brownian motion (Ito's convention) and $T=\dt/B$.

{\it Heuristic argument for observation that} $T_{max}\sim T_{opt}\sim \alpha^k$:
Considering the SDE description \ref{eq:general_SDE} of SGD, the corresponding flux $\bm{J}$ for the weights distribution $\rho(\w,t)$ can be written as \citep{chaudhari2018}
\beq
\label{flux}
    \bm{J}\lpa\w,t\rpa = \rho(\w,t) \nabla L(\w) + \frac{1}{2} T \nabla\cdot\lpa\cov(\w) \rho(\w,t)\rpa
\eeq
where the divergence operator $\nabla\cdot$ is applied column-wise to the matrix $\cov \rho$. We notice that the probability flux receives a contribution from both the loss gradient and the covariance divergence. To understand the effect of SGD in the feature regime, we need to compare the scaling of the two terms $\rho(\w,t) \nabla L(\w)$ and $\nabla\cdot\lpa\cov(\w) \rho(\w,t)\rpa$ in the limit of $\alpha\ll 1$.\\
In this limit and with the network initialization considered in Sec. \ref{sec:definition}, the variation of the weights in every layer has the same scale $w$ with respect to $\alpha$. Therefore, for a network of depth $D+1$ with ReLU activation functions, the predictor variation $\Delta f$ is related to $w$ by $\Delta f \sim w^{D+1}$. To bring the hinge loss to zero, the predictor has to be of the same order of the margin $\alpha^{-1}$, which corresponds to the scaling $w^{D+1}\sim \alpha^{-1}$ or, equivalently, 
\beq
w\sim \alpha^{-1/(D+1)}.
\label{eq:w_alpha}
\eeq
The scaling of $\nabla L$ and $\nabla\cdot\cov$ with respect to $w$ is easily obtained by inspecting the definitions of $L$ (Eq. \ref{eq:hingeLoss}) and $\cov$ (Eq. \ref{eq:general_cov}). For the hinge loss,  we have
\beq
\nabla L \sim \nabla f \sim w^D
\label{eq:gradL}
\eeq
and 
\beq
\nabla\cdot\cov \sim \nabla (\nabla f)^2\sim w^{2D-1}.
\label{eq:divS}
\eeq
Therefore, the two terms $\nabla L$ and $T \nabla\cdot\cov$ become comparable when $w^D \sim T w^{2D-1}$, that is for a characteristic temperature $T\sim w^{-D+1}$. By using \ref{eq:w_alpha}, this corresponds to
\beq
    T \sim \alpha^{(D-1)/(D+1)}.
    \label{eq:T_feature}
\eeq
For much larger temperatures,  the noise term $T \nabla\cdot\cov$ is much larger than the signal $\nabla L$, and we expect the dynamics not to converge.  For much smaller temperatures, noise is negligible. These arguments support that $T_{max}\sim T_{opt}\sim \alpha^k$ with $k=(D-1)/(D+1)$, as confirmed in Fig. \ref{fig:test_phase}. 

In the lazy regime, the network weights are always of $O(1)$ with respect to $\alpha$, therefore the two terms \ref{eq:gradL} and \ref{eq:divS} don't scale with $\alpha$ for $\alpha\gg1$. Consequently, the characteristic temperatures in this regime are independent of $\alpha$.

\section{Additional plots in the lazy regime}
\label{app:plots_lazy}

\begin{figure}[H]
    \centering
    \includegraphics[width=.48\textwidth]{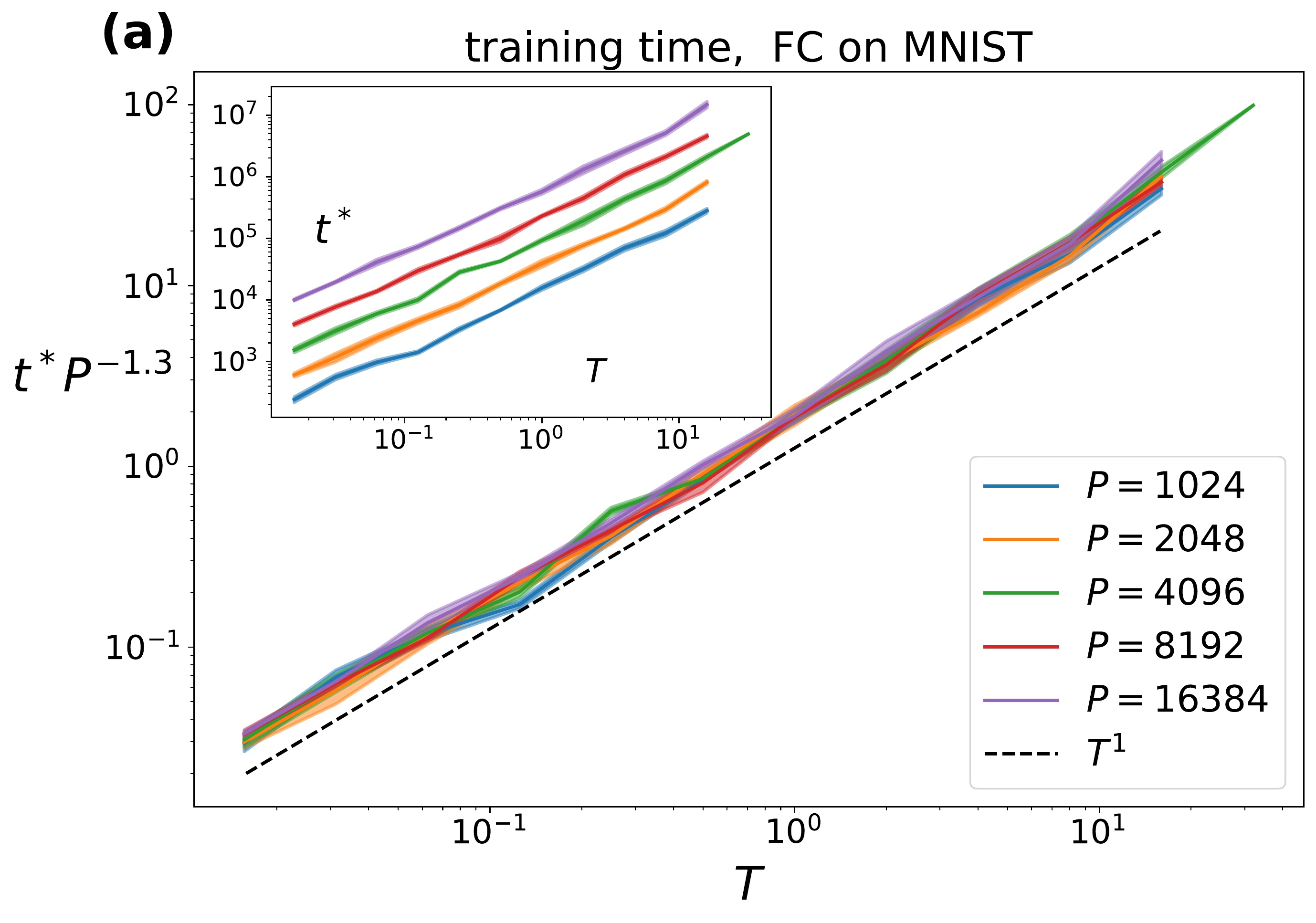}
    \includegraphics[width=.48\textwidth]{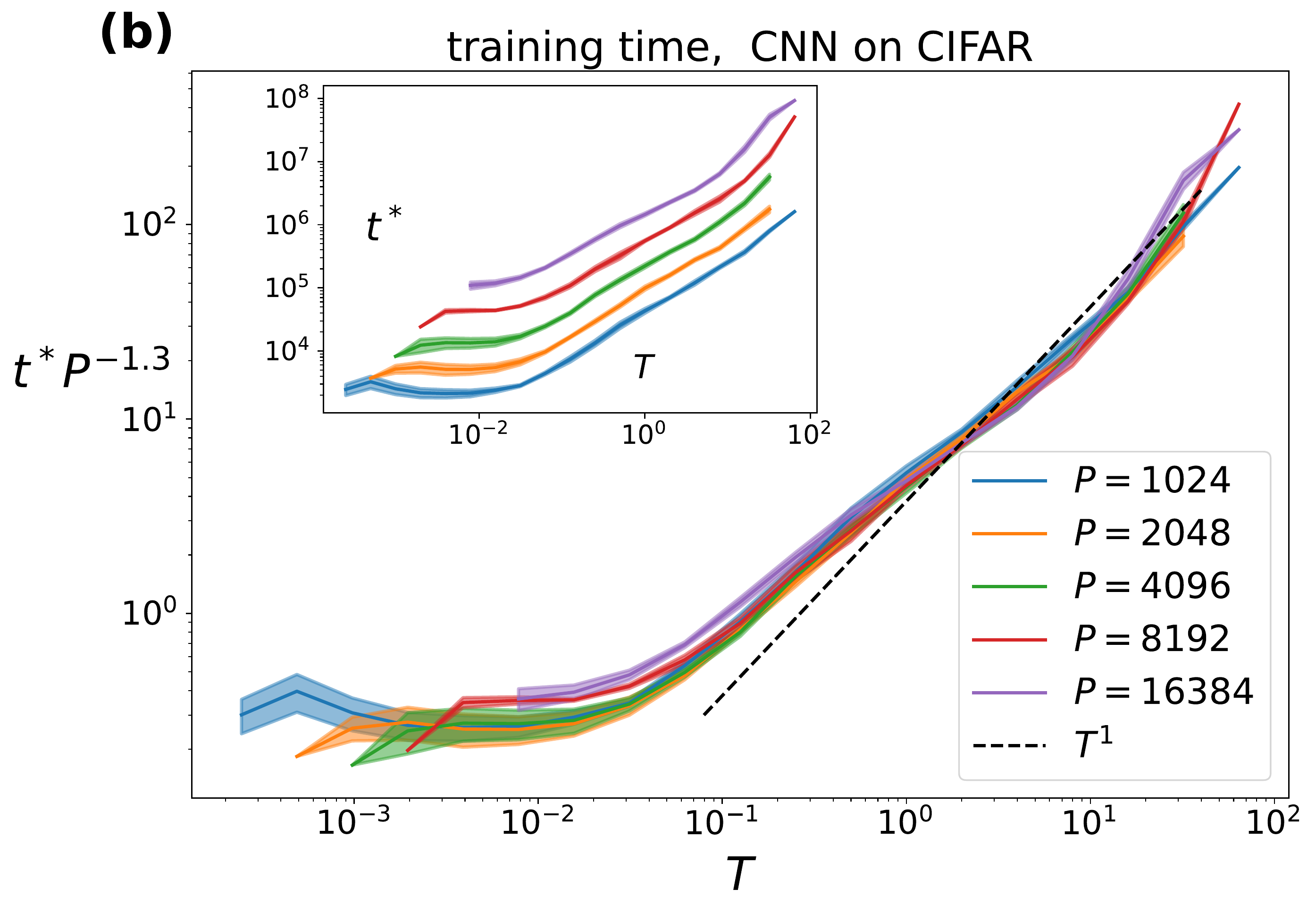}
    \caption{\textbf{Training time, lazy regime, $\alpha=32768$, $B=16$, varying $P$ and $T$: (a) FC on MNIST, (b) CNN (MNAS) on CIFAR.} \textit{Inset:} $t^*$ increases with both $T$ and $P$. \textit{Main:} Plotting $t^* P^{-b}$, with $b$ a fitting exponent ($b\approx 1.3$), yields a curve increasing approximately linearly in $T$, suggesting a dependence $t^*\sim T P^{b}$.}
    \label{fig:time_lazy}
\end{figure}

\begin{figure}[H]
    \centering
    \includegraphics[width=\textwidth]{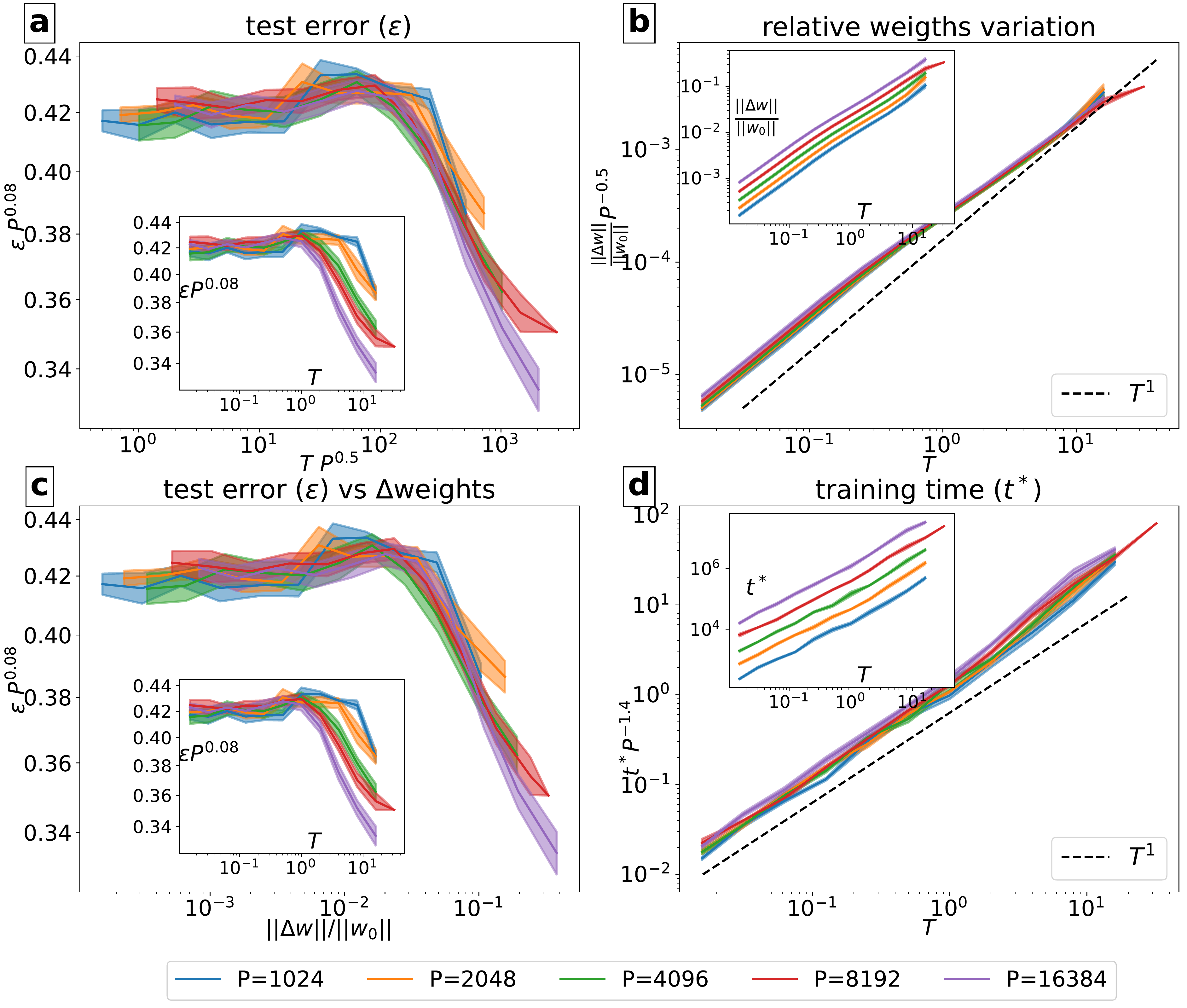}
    \caption{\textbf{FC on CIFAR, $\alpha=32768$, $B=16$, varying $P$ and $T$.}
    \textbf{(a): test error $\epsilon$.} \textit{Inset:} $\epsilon$ starts improving at a cross-over temperature $T_{c}$ depending on $P$. The y-axis is rescaled by $P^\beta$, with $\beta$ some fitting exponent, to align $\epsilon$ at $T_c$. \textit{Main:} Rescaling the x-axis by $P^{0.5}$ aligns horizontally the points where $\epsilon$ starts improving, suggesting a dependence $T_c \sim P^{-0.5}$.
    \textbf{(b): total weight variation at the end of training normalized with respect to their initialization ($\Delta w$).} \textit{Inset:} $\Delta w$ increases with both $T$ and $P$. \textit{Main:} Plotting $\Delta w P^{-\ec}$ yields a curve increasing approximately as $T^\ed$, suggesting $\Delta w\sim T^\ed P^{\ec}$, with $\ec$ and $\ed$ some fitting exponents.
    \textbf{(c): test error vs weight variation.} The point where the test error starts improving shows a better alignment when plotted as a function of the weight variation (\textit{main plots}) rather than temperature alone (\textit{insets}).
    \textbf{(d): training time $t^*$.} \textit{Inset:} $t^*$ increases with both $T$ and $P$. \textit{Main:} Plotting $t^* P^{-b}$, with $b$ a fitting exponent, yields a curve increasing approximately linearly in $T$, suggesting a dependence $t^*\sim T P^{b}$.
    }
    \label{fig:FC_cifar}
\end{figure}
\begin{figure}[H]
    \centering
    \includegraphics[width=\textwidth]{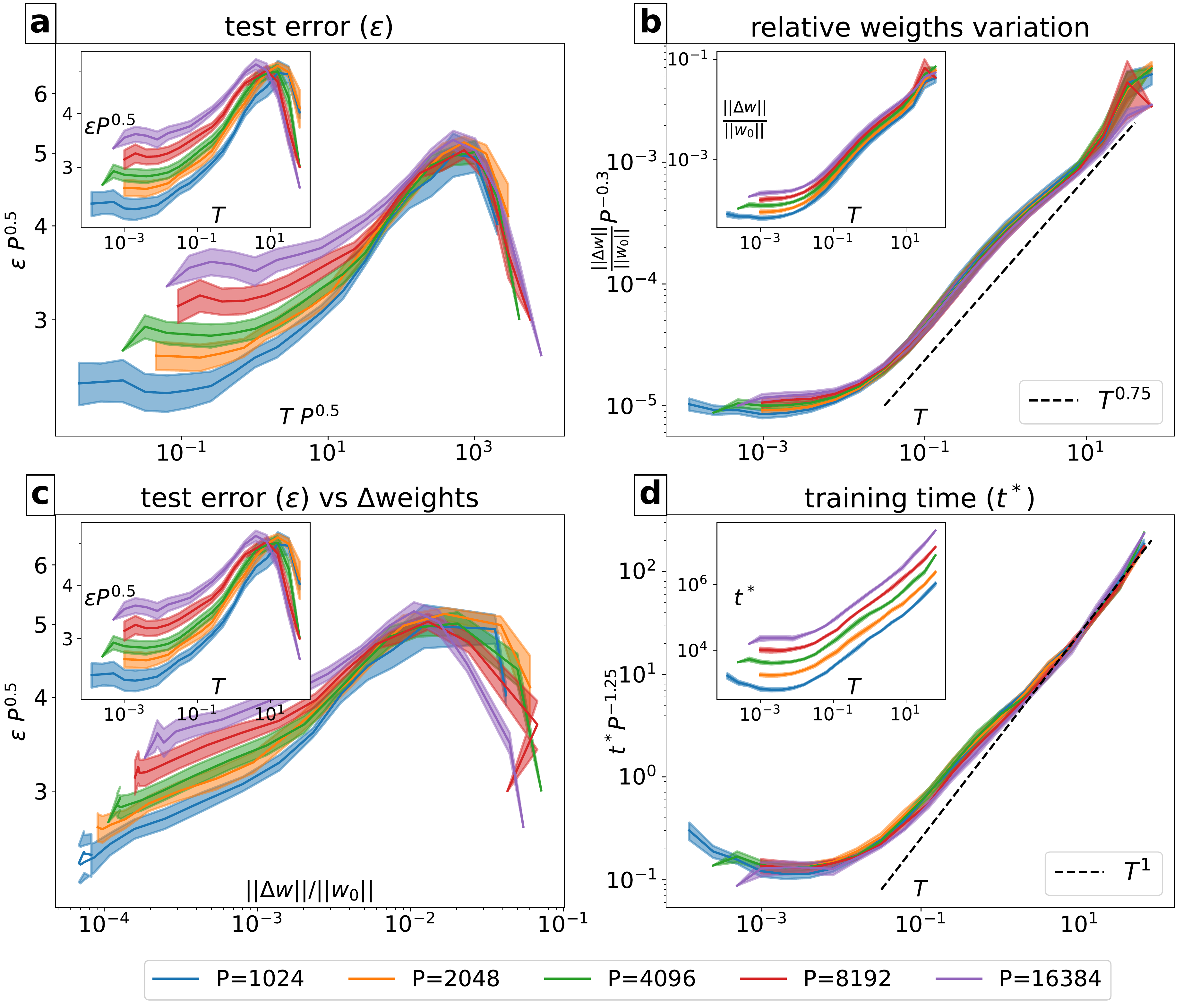}
    \caption{\textbf{CNN (MNAS) on MNIST, $\alpha=32768$, $B=16$, varying $P$ and $T$: (a) test error, (b) relative weight variation, (c) test error vs relative weight variation, (d) training time.} Same quantities as Fig. \ref{fig:FC_cifar}, see its caption.}
    \label{fig:MNAS_mnist}
\end{figure}
\begin{figure}[H]
    \centering
    \includegraphics[width=\textwidth]{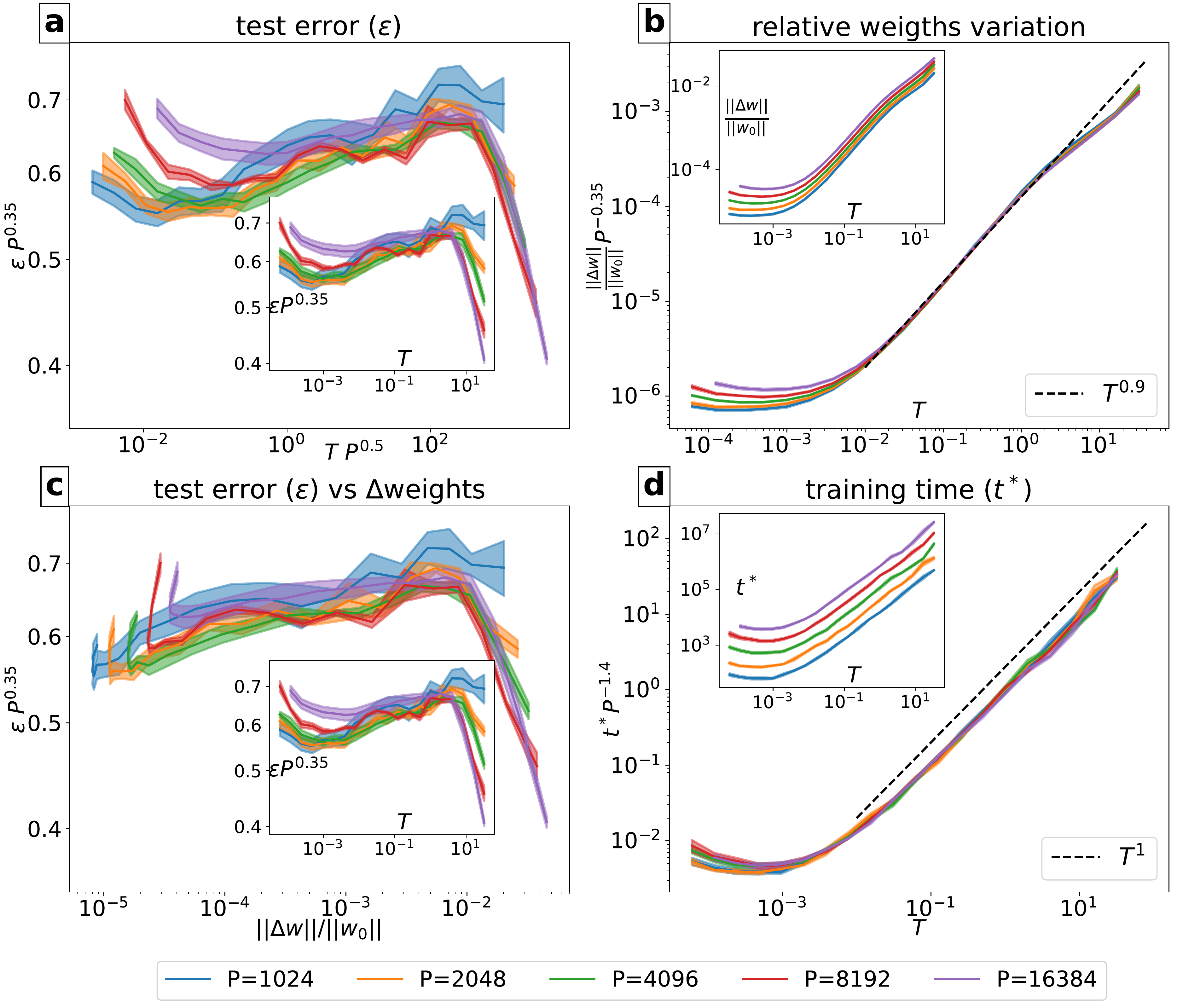}
    \caption{\textbf{simpleCNN on MNIST, $\alpha=32768$, $B=16$, varying $P$ and $T$: (a) test error, (b) relative weight variation, (c) test error vs relative weight variation, (d) training time.} Same quantities as Fig. \ref{fig:FC_cifar}, see its caption.}
    \label{fig:simpleCNN_mnist}
\end{figure}
\begin{figure}[H]
    \centering
    \includegraphics[width=\textwidth]{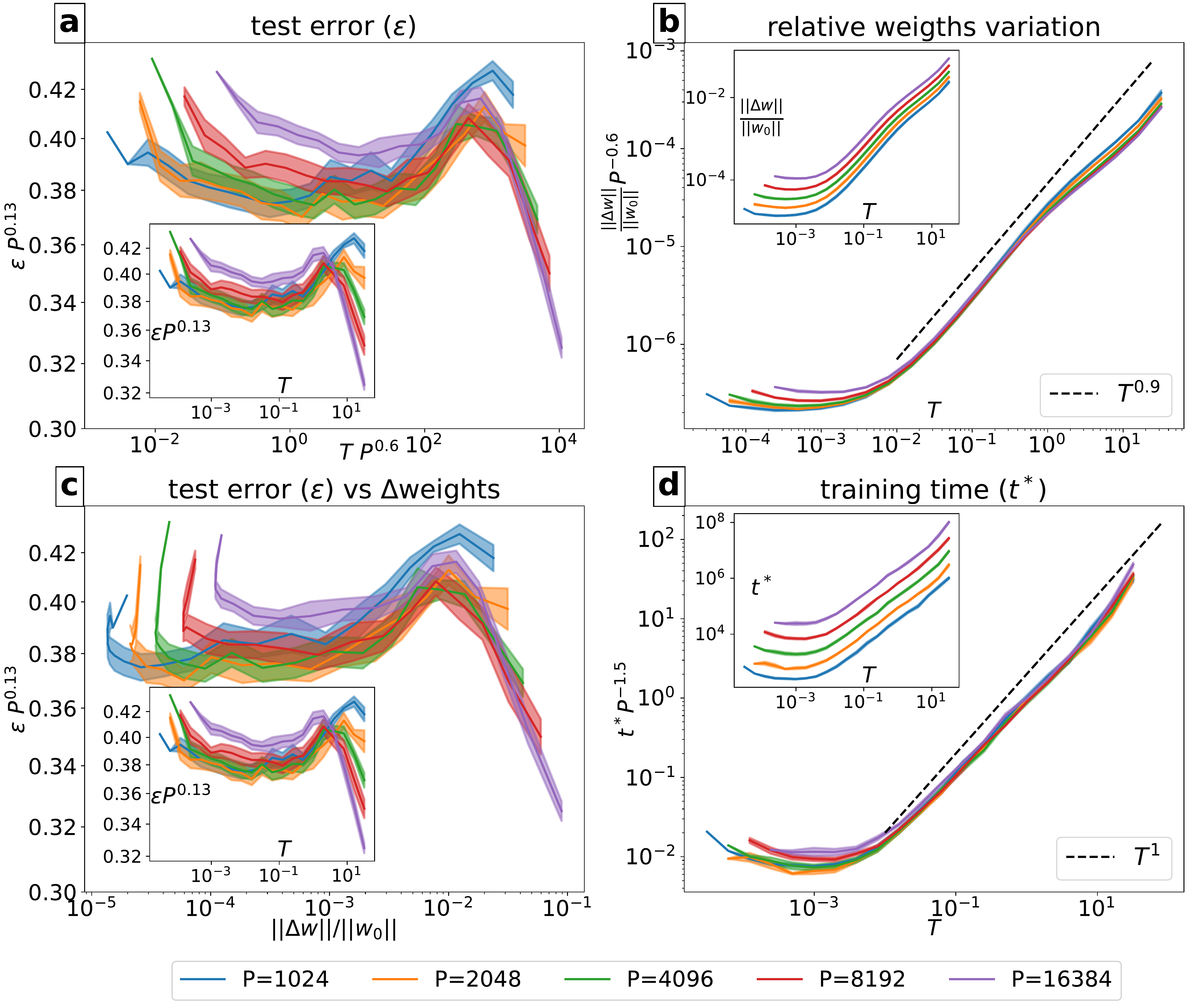}
    \caption{\textbf{simpleCNN on CIFAR, $\alpha=32768$, $B=16$, varying $P$ and $T$: (a) test error, (b) relative weight variation, (c) test error vs relative weight variation, (d) training time.} Same quantities as Fig. \ref{fig:FC_cifar}, see its caption.}
    \label{fig:simpleCNN_cifar}
\end{figure}

\section{Impact of the training set size $P$ in the feature-learning regime}
\label{app:feature_regime}
\begin{figure}[H]
    \centering
    \includegraphics[width=\textwidth]{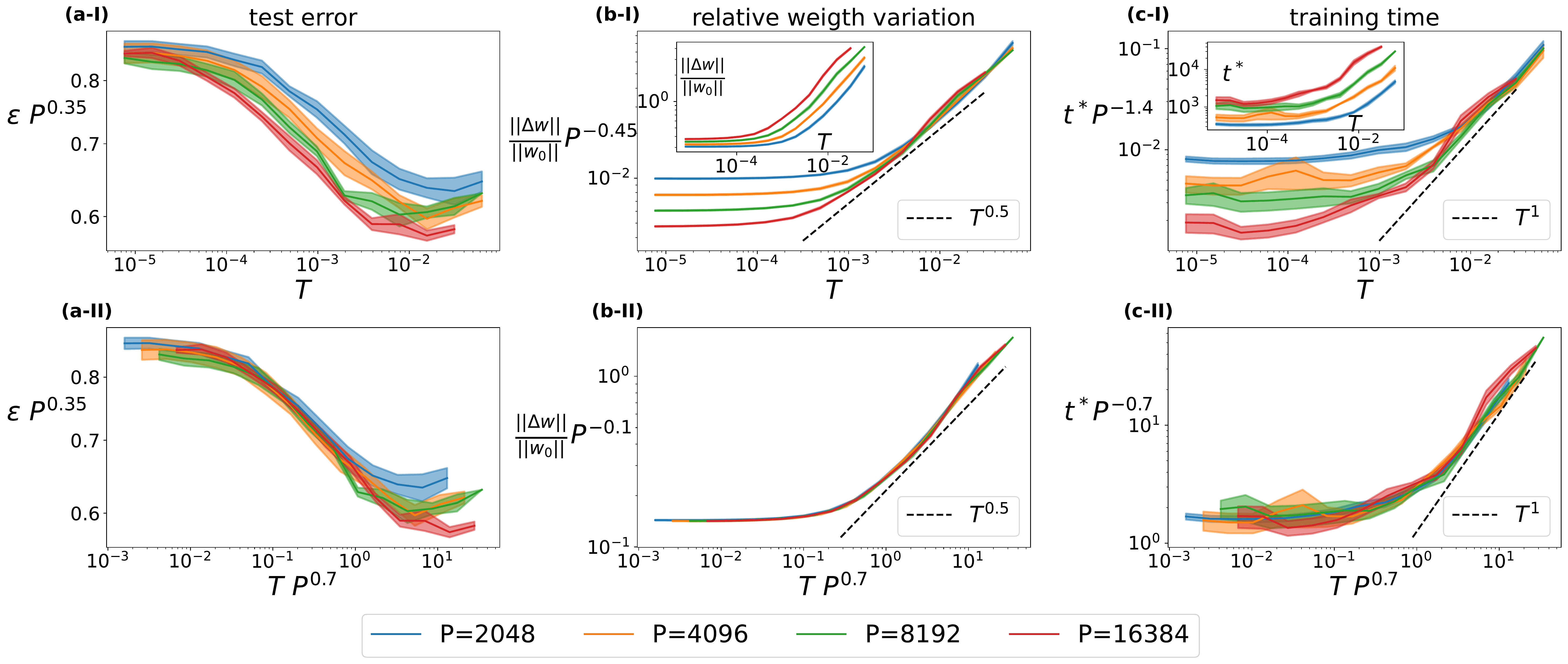}
    \caption{\textbf{FC on MNIST, feature regime, $\alpha=2^{-10}$, $B=16$, $T=\eta/B$.} 
    \textbf{(a-I, a-II): test error ($\epsilon$) vs temperature ($T$).} \textit{(a-I):} $\epsilon$ starts improving at a cross-over temperature $T_{c}$ depending on $P$. The y-axis is rescaled by $P^\beta$, with $\beta$ some fitting exponent, to align $\epsilon$ at the lowest $T$. \textit{(a-II):} Rescaling the x-axis by $P^{0.7}$ aligns horizontally the points where $\epsilon$ starts improving, suggesting a dependence $T_c \sim P^{-0.7}$.
    \textbf{(b-I, b-II): total weight variation at the end of training normalized with respect to their initialization ($\Delta w=||\Delta \w||/||\w_0||$) vs $T$.} 
    \textit{(b-I, inset):} $\Delta w$ increases with both $T$ and $P$. 
    \textit{(b-I, main):} Plotting $\Delta w P^{-\ec}$ yields a curve, for large $T$, increasing approximately as $T^\ed$, suggesting $\Delta w\sim T^\ed P^{\ec}$, with $\ec\approx 0.45$ and $\ed\approx 0.5$.
    \textit{(b-II):} Rescaling the x-axis by $P^{0.7}$ aligns horizontally the points where $T$ starts having an effect on the weights, corresponding to $T_c \sim P^{-0.7}$. For $T\ll T_c$, the weight variation scale as $\Delta w \sim P^{\ei}$, with $\ei\approx 0.1$.
    \textbf{(c-I, c-II): training time ($t^*$) vs temperature ($T$).} \textit{(c-I, inset):} $t^*$ increases with both $T$ and $P$. \textit{(c-I, main):} Plotting $t^* P^{-b}$ ($b\approx 1.4$) yields a curve, for large $T$, increasing approximately linearly in $T$, suggesting a dependence $t^*\sim T P^{b}$.
    \textit{(c-II):} Rescaling the x-axis by $P^{0.7}$ aligns horizontally the points where $T$ starts having an effect on the training time, corresponding to $T_c \sim P^{-0.7}$. For $T\ll T_c$, $t^*$ scales as $t^* \sim P^{0.7}$.
    }
    \label{fig:FC_feature}
\end{figure}

\begin{figure}[H]
    \centering
    \includegraphics[width=\textwidth]{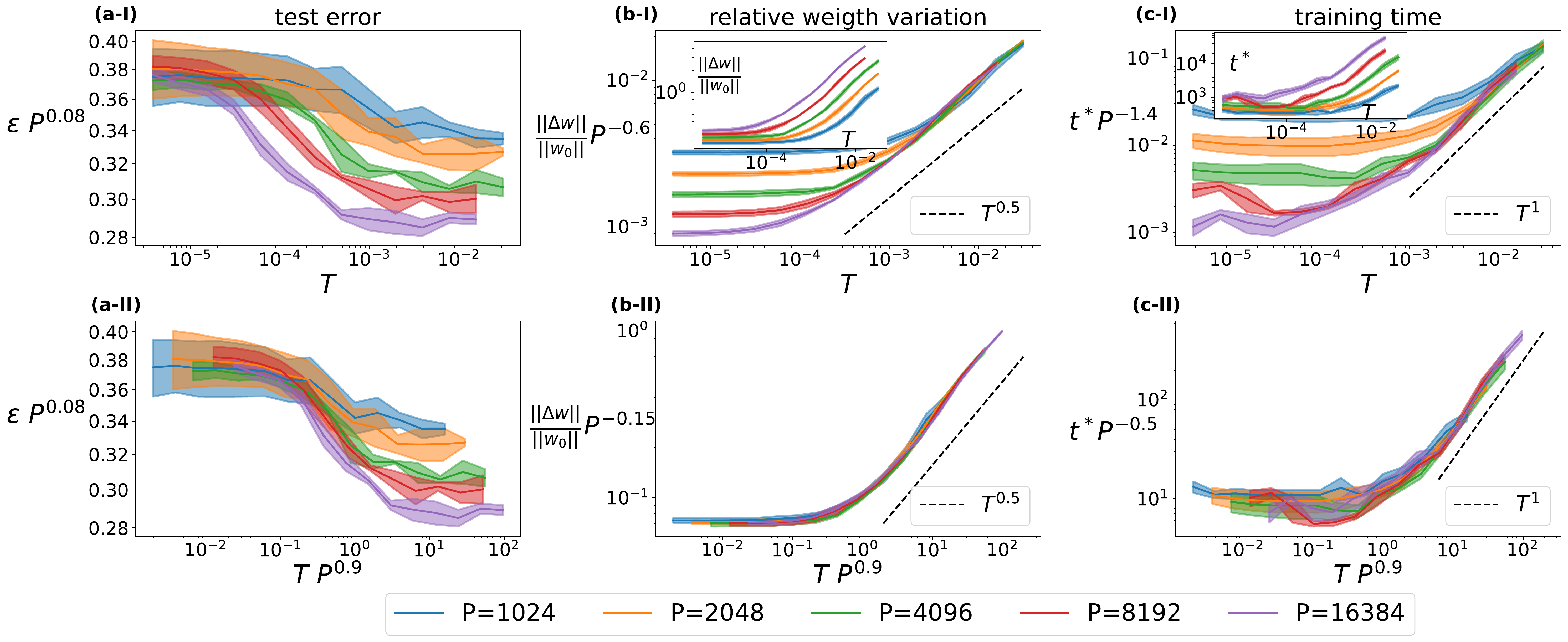}
    \caption{\textbf{FC on CIFAR, feature regime, $\alpha=2^{-10}$, $B=16$, $T=\eta/B$.}
    Same quantities as Fig. \ref{fig:FC_feature}, see its caption.
    }
    \label{fig:FC_feature_cifar}
\end{figure}

\begin{figure}[H]
    \centering
    \includegraphics[width=\textwidth]{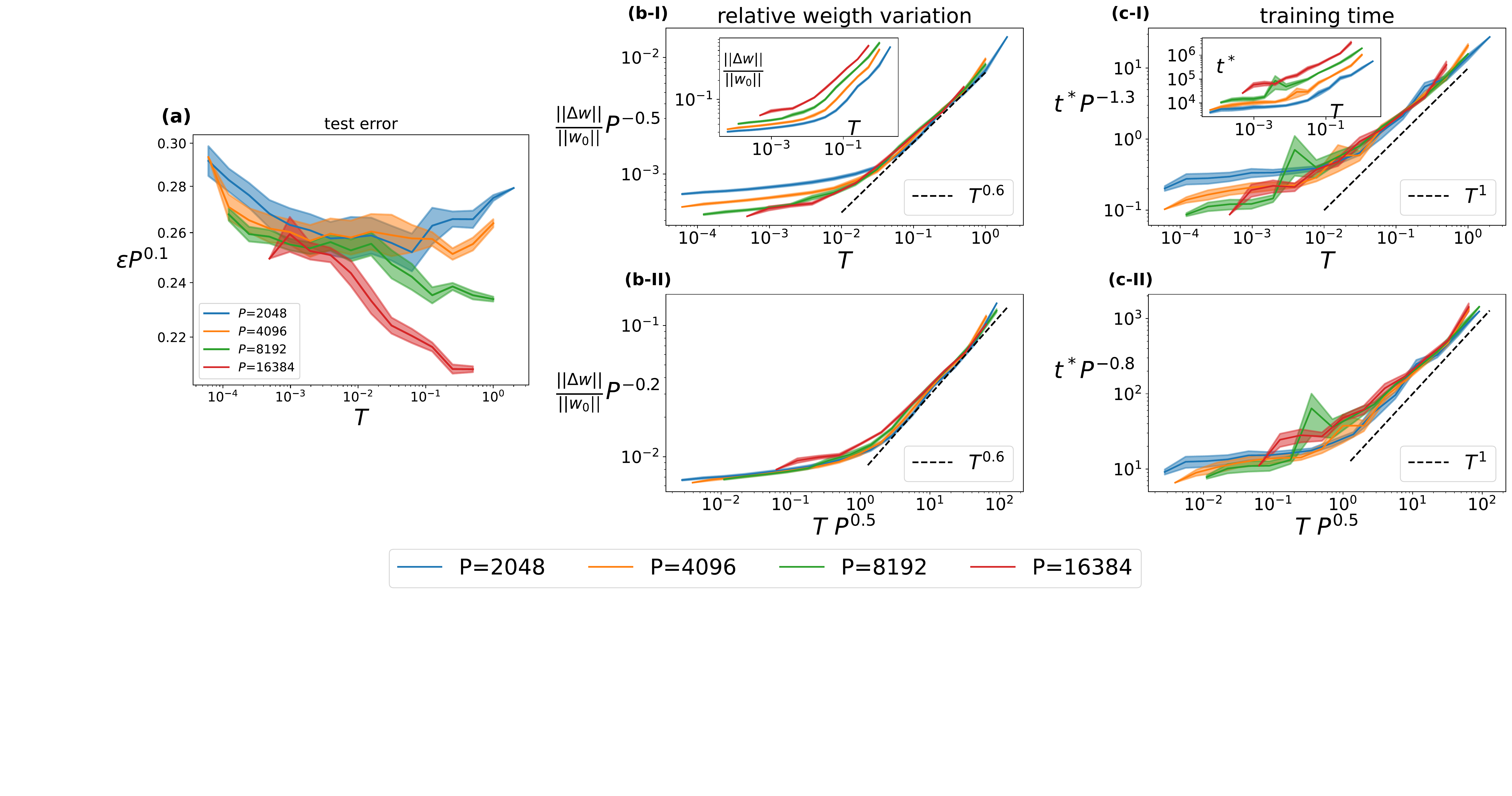}
    \caption{\textbf{CNN (MNAS) on CIFAR, feature regime, $\alpha=1$, $B=16$, $T=\eta/B$.} 
    \textbf{(a): test error ($\epsilon$) vs temperature ($T$).} $\epsilon$ improves more significantly with $T$ when increasing $P$. In this case the curves have different shapes and cannot be matched by rescaling the x-axis. The y-axis is rescaled by $P^\beta$, with $\beta$ some fitting exponent, to make the curves easier to compare.
    \textbf{(b-I, b-II): total weight variation at the end of training normalized with respect to their initialization ($\Delta w = ||\Delta \w||/||\w_0||$) vs $T$.} 
    \textit{(b-I, inset):} $\Delta w$ increases with both $T$ and $P$. 
    \textit{(b-I, main):} Plotting $\Delta w P^{-\ec}$ yields a curve, for large $T$, increasing approximately as $T^\ed$, suggesting $\Delta w\sim T^\ed P^{\ec}$, with $\ec\approx 0.5$ and $\ed\approx 0.6$.
    \textit{(b-II):} Rescaling the x-axis by $P^{0.5}$ aligns horizontally the points where $T$ starts having an effect on the weights, corresponding to $T_c \sim P^{-0.5}$. For $T\ll T_c$, the weight variation scales as $\Delta w \sim P^{\ei}$, with $\ei\approx 0.2$.
    \textbf{(c-I, c-II): training time ($t^*$) vs temperature ($T$).} \textit{(c-I, inset):} $t^*$ increases with both $T$ and $P$. \textit{(c-I, main):} Plotting $t^* P^{-b}$ ($b\approx 1.3$) yields a curve, for large $T$, increasing approximately linearly in $T$, suggesting a dependence $t^*\sim T P^{b}$.
    \textit{(c-II):} Rescaling the x-axis by $P^{0.5}$ aligns horizontally the points where $T$ starts having an effect on the training time, corresponding to $T_c \sim P^{-0.5}$. For $T\ll T_c$, $t^*$ scales approximately as $t^* \sim P^{0.8}$.
    }
    \label{fig:CNN_feature}
\end{figure}

\paragraph{Empirical observations.} 
In the feature-learning regime, we observe the same scaling behaviors as in the lazy regime, which are discussed in Section \ref{sec:role_P}. In particular, 
the relative change of weights ($\Delta w = \frac{||\w^{t^*}-\w^{0}||}{||\w^{0}||}$), the training time ($t^*$) and the characteristic temperature ($T_c$) where  $\Delta w$ and $t^*$ start being affected by SGD noise exhibit asymptotic behavior as follows:
\beq
    T_{c} \sim P^{-a},
    \qquad
    \Delta w \sim T^\ed P^\ec,
    \qquad
    t^*\sim T P^b.
\label{eq:feature_scaling}
\eeq
From the data, we measure $T_c$ as the temperature at which $\Delta w$ starts increasing with $T$. In some cases, this also corresponds to the temperature scale where the test error starts improving (e.g. for the FC architecture in Figs. \ref{fig:FC_feature}-(a-I, a-II) and \ref{fig:FC_feature_cifar}-(a-I, a-II)). In some other cases, instead, the curve of the test error vs $T$ can take different shapes when varying $P$, and therefore extracting a $T_c$ from it is not possible (e.g. for the CNN architecture in Fig. \ref{fig:CNN_feature}-(a)).\\
The values of the exponents $a$, $b$, $\ec$, $\ed$ in Eq. \ref{eq:feature_scaling} are slightly different from those measured in the lazy regime. For instance, for the fully connected neural network on MNIST in feature learning, we observe $a\approx 0.7$, $\ed\approx 0.5$, $\ec\approx 0.45$, and $b\approx 1.4$, while in lazy learning, we observe $a\approx 0.5$, $\ed\approx 1.0$, $\ec\approx 0.4$, and $b\approx 1.3$. Table \ref{tab:exponents} provides a comparison of the exponents.
%The important difference with the lazy regime is that the scaling relationships for $\Delta w$ and $t^*$ apply only for $T>T_c$, while in the lazy regime they hold also for lower $T$ (i.e. for any $T\gg\alpha^{-1}$).\\

\paragraph{Interpretation.}
The same scaling behaviors of \eqref{eq:feature_scaling} are observed in both the feature and lazy regimes. We argue that this similarity comes from the fact that the two training regimes are similar at late times- their main difference corresponds to early times in the dynamics.
In the feature regime, the weights need to grow considerably to make the output $\mathcal{O}(1)$\footnote{In our setting, this corresponds to $\alpha F(\w,\x)\sim \mathcal{O}(1)$.}
and fit the data. At the beginning of the training dynamics, before fitting any data, the weights grow exponentially in time- an initial phase of training that we refer to as an `inflation period' \citep{geiger2020disentangling,paccolat2021geometric}.
Afterwards, the network starts fitting the data, and the dynamics is similar to that of the lazy regime, with the exception that the neural tangent kernel has evolved during inflation. In this second part of the dynamics, we expect the arguments presented in Section \ref{sec:interpretation} to apply, as supported by our empirical observations.\\

\paragraph{Characteristic temperature.} In feature learning, the characteristic temperature $T_c$ corresponds to the cross-over point between the `inflation dominated' and the `noise dominated' dynamics.\\
Specifically, for $T\ll T_c$, weight variation is mainly concentrated in the initial part of the dynamics, as observed in studies on gradient flow \citep{geiger2020disentangling} that corresponds to the limit $T\rightarrow 0$. In this case, the total weight variation is independent of $T$ and appears to be a function of $P$ as 
\beq
\Delta w_{INFL} \sim P^\ei,
\eeq
where $\ei$ is a fitting exponent. For instance, in a fully connected network on the MNIST dataset, $\ei\approx 0.1$ (see Figure \ref{fig:FC_feature}-(b-II)).
Conversely, for $T\gg T_c$, most of the weight variation occurs in the later part of the dynamics, when SGD noise becomes relevant. Thus, $\Delta w_{NOISE} \sim T^\ed P^\ec$ (\eqref{eq:feature_scaling}).\\
Being the cross-over between these two regimes, the characteristic temperature $T_c$ is determined by the condition $\Delta w_{INFL}\sim \Delta w_{NOISE}$, which corresponds to $P^\ei \sim T_c^\ed P^\ec$. Therefore,
\beq
    T_c \sim P^{-\frac{\ec-\ei}{\ed}}
\eeq
which yields the relationship $T_c\sim P^{-a}$ with an exponent $a$ satisfying
\beq
    a = \frac{\ec-\ei}{\ed},
\eeq
in accordance with the experiments (see Table \ref{tab:exponents}).
It should be noted that this relation differs somewhat from the one observed in the lazy regime, where $a = \ec / \ed$ and the characteristic temperature is determined by comparing weight variation to their initialization.

\section{Comparison between hinge loss and cross-entropy loss}
\label{app:hinge_cross}
\begin{figure}
    \centering
    \includegraphics[width=.7\textwidth]{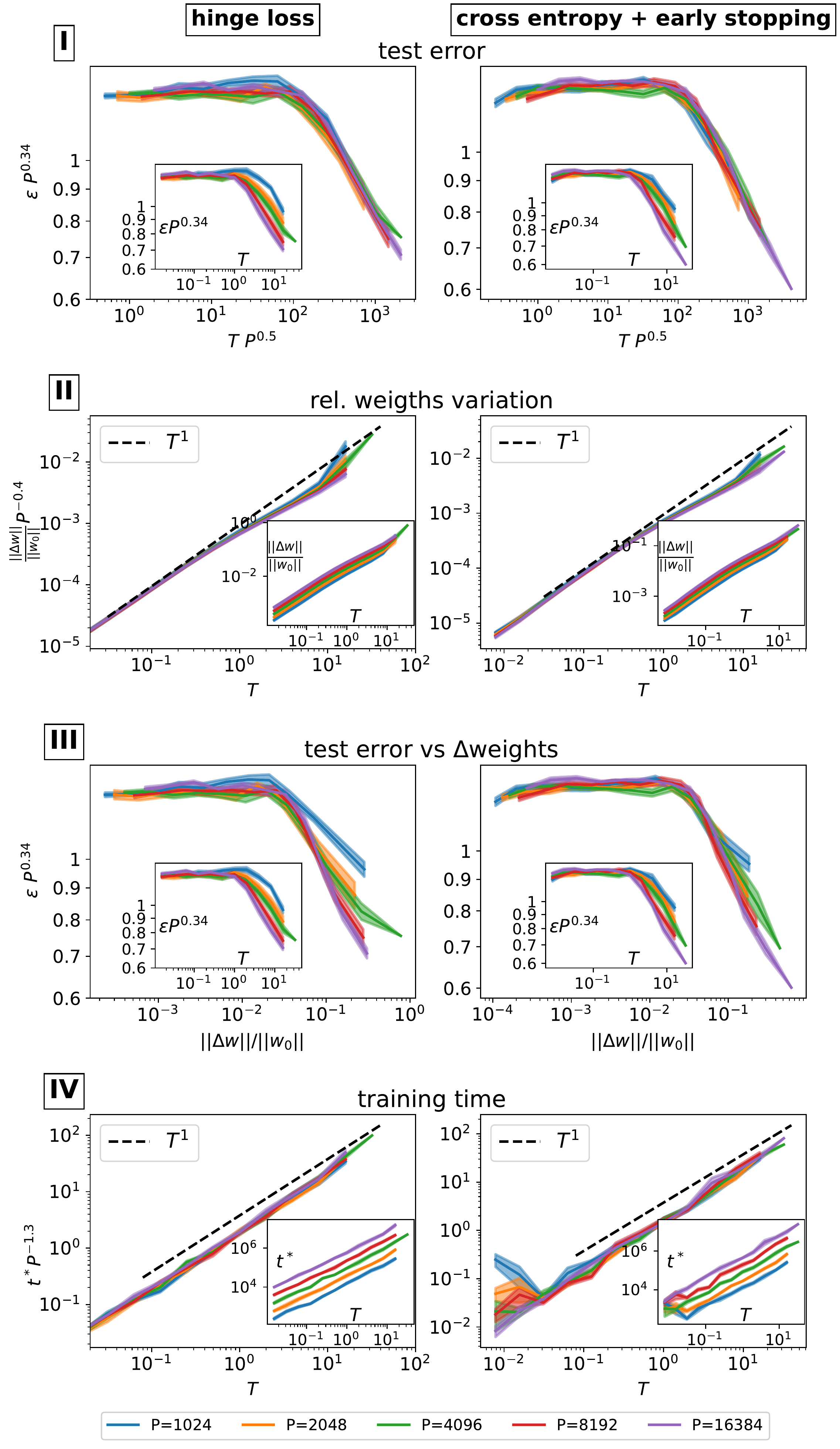}
    \caption{\textbf{Comparison of results obtained with the hinge loss trained until $0$ loss and the cross-entropy loss performing early-stopping. FC on MNIST, lazy regime, $\alpha=32768$, $B=16$, $T=\eta/B$.} On the left column, there are the data obtained with the hinge loss, already presented in Fig. \ref{fig:FC_lazy} and Fig. \ref{fig:time_lazy}-(a) (see their captions). On the right column, the same quantities obtained by training with the cross-entropy loss show identical dependence on $T$ and $P$, with power-laws having the same exponents.}
    \label{fig:FC_cross}
\end{figure}

This section shows that the setting of our work, using the hinge loss and training until it reaches zero value, is very similar to training with the cross-entropy loss and performing early stopping 
\footnote{In this case, we define the early stopping procedure as follows: \textit{(i)} we store the model weights and the validation error at various checkpoints (e.g. every epoch) during the training dynamics; \textit{(ii)} the training dynamics is considered terminated when the training error is zero and the test error is not improving between consecutive checkpoints; \textit{(iii)} we take as final weights of the network those which gave the lowest validation error during the training dynamics. They correspond to some checkpoint before reaching zero training error, since some over-fitting is observed in the last part of the dynamics.}.\\
Figure \ref{fig:FC_cross} shows that the two training procedures give identical power-law dependencies on $T$ and $P$ for all the quantities we analyse, meaning that the exponents of the power-laws are the same. Therefore, our results are relevant for training networks in practical classification tasks.

\section{Distribution of the maximum of $c_{\mu}/|x_1^\mu|$}
\label{app:max}
In this section we compute the distribution of the random variable $M_P = \underset{\mu}{\text{max}} \frac{c_{\mu}}{|x^{\mu}_1|}$, $\mu=1,...,P$.\\
Considering that the problem is rotationally invariant in the $(d-1)$-subspace $\x_\perp$, since $y^{\mu} = \text{sign}(x_1^{\mu})$ and $\x_\perp$ is normally distributed, we make the following assumptions for $c_{\mu} = -\frac{\w_\perp}{||\w_\perp||} \cdot \x^{\mu}_\perp y^{\mu}$ and $|x_1^\mu|$:
\begin{itemize}
    \item $c_{\mu}$ are independent and identically distributed (i.i.d.) random variables, whose probability distribution $\rho_{c_\mu}$ is Gaussian with zero mean and variance $\sigma^2$;
    \item $c_\mu$ and $|x_1^\mu|$ are independent.
\end{itemize}

Calling $z_\mu = |x_1^\mu|^{-1}$, from \ref{eq:rho_x1} the probability distribution of $z_\mu$ is given by
\beq
    \rho_{z_\mu}(z) = z^{-\chi-2} e^{1/(2 z^2)}/\Tilde{Z}
\eeq
with $\Tilde{Z}$ the normalization constant.\\
Since $q_\mu = c_\mu z_\mu = c_\mu |x_1^\mu|^{-1}$ is the product of two independent random variables, its probability distribution is given by the basic formula $\rho_{q_\mu}(q) = \int_{0}^{\infty} \rho_{z_\mu}(z) \rho_{c_\mu}(q/z) z^{-1} dz$, which in this case reads:
\beq
\begin{aligned}
    \rho_{q_\mu}(q) &= \int_{0}^{\infty} \rho_{z_\mu}(z) \rho_{c_\mu}(q/z) z^{-1} dz =\\
    &= \frac{1}{\Tilde{Z}\sqrt{2\pi}\sigma} \int_{0}^{\infty} z^{-\chi-3} e^{-\frac{1}{2z^2}} e^{-\frac{q^2}{2\sigma^2 z^2}} dz = \\
    &= \lpa 1+\frac{q^2}{\sigma^2}\rpa^{-\frac{1}{2}\lpa\chi+2\rpa} \frac{1}{\Tilde{Z}\sqrt{2\pi}\sigma} \int_{0}^{\infty} {z'}^{-\chi-3} e^{-\frac{1}{2z'^2}} dz' = \\
    & = K \lpa 1+\frac{q^2}{\sigma^2}\rpa^{-\frac{1}{2}\lpa\chi+2\rpa}
\end{aligned}
\eeq
with the normalization constant $K = \frac{2 \Gamma\lpa\frac{\chi+2}{2}\rpa}{\sqrt{\pi}\sigma \Gamma\lpa\frac{\chi+1}{2}\rpa}$. 
% The probability $\mathcal{P}\lpa q_\mu>q\rpa$ that $q_\mu>q$ is given by
% \beq
%     \mathcal{P}\lpa q_\mu>q\rpa = \frac{K \sigma^{\chi+2}}{\chi+1} q^{-\chi-1} + o\lpa q^{-\chi-1} \rpa, \text{ for } q\rightarrow\infty.
% \eeq

Therefore, since in the limit $q\rightarrow\infty$ the distribution $\rho_{q_\mu}(q)$ behaves as a power law $\rho_{q_\mu}(q)\sim K \lpa\frac{q}{\sigma}\rpa^{-(\chi+2)}$, the distribution of the maximum $M_P = \underset{\mu}{\text{max}}\ q_\mu$, with $\mu=1,...,P$, in the limit of large $P$, converges to the Fréchet distribution \citep{gnedenko1943,leadbetterExtremes}:
\beq
    \mathcal{P}\lpa a_P M_P < t\rpa \underset{P\rightarrow\infty}{\rightarrow} \exp\lpa -t^{-\chi-1}\rpa, \quad \text{for }t>0
\eeq
with 
\beq
    a_P = \lpa\frac{K\sigma^{\chi+2}}{\chi+1} P\rpa^{-\frac{1}{\chi+1}}.
\eeq
Thus, we obtain that the typical value of the maximum $\langle M_P\rangle\propto a_P^{-1}$ behaves asymptotically for $P\rightarrow\infty$ as
\beq
    \langle M_P \rangle = C P^{\frac{1}{\chi+1}} + o\lpa P^\frac{1}{\chi+1}\rpa
\eeq
with a constant $C$.\\
This asymptotic behaviour can also be found simply by imposing the condition $\int_{\langle M_P \rangle}^\infty \rho_{q_\mu}(q) \sim P^{-1}$ and expanding $\rho_{q_\mu}(q)$ for large $q$.

\section{Additional plots}
\label{app:details}

\subsection{Learning rate and batch size}
\label{app:eta_B}
\begin{figure}[H]
    \centering
    \includegraphics[width=.31\textwidth]{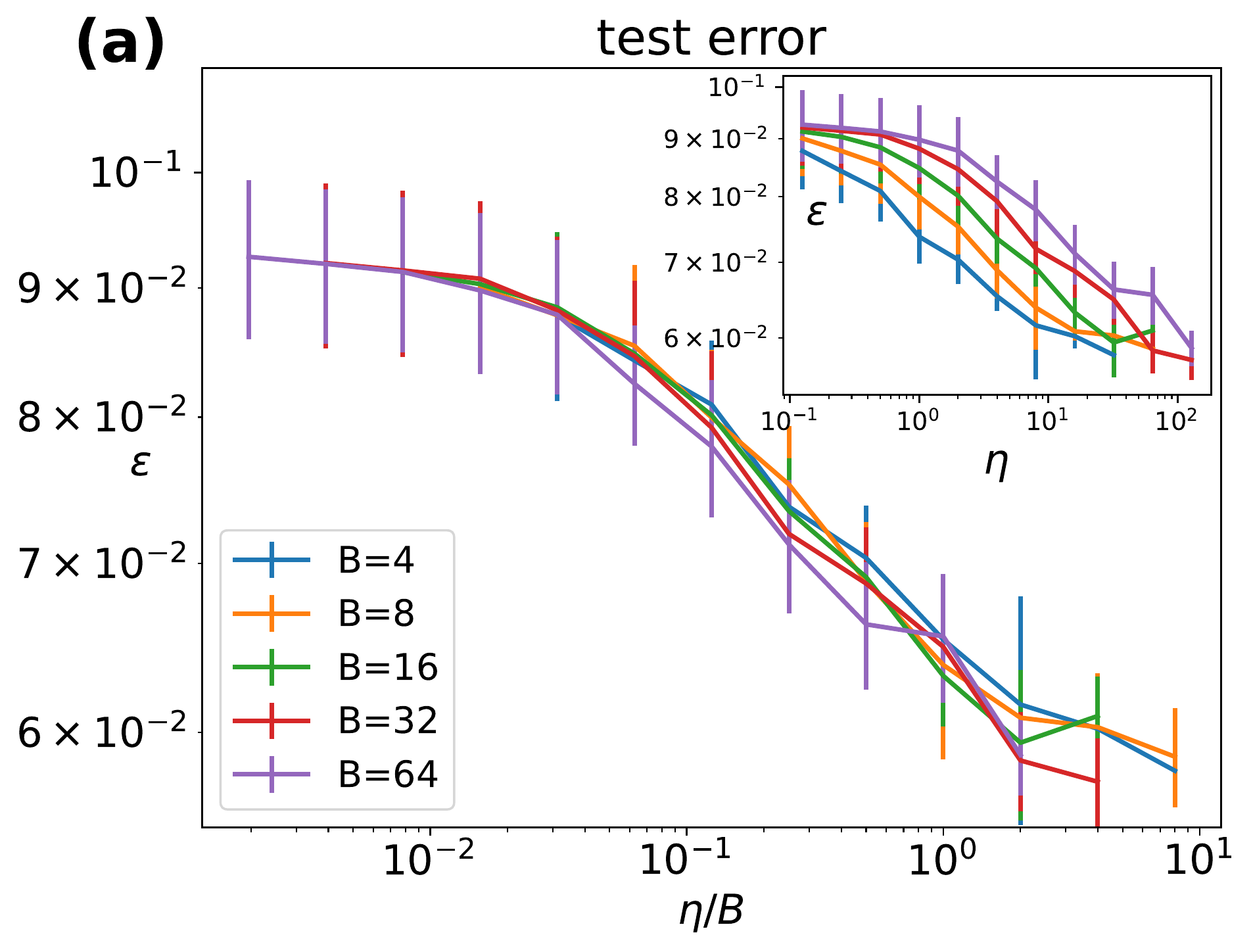}
    \includegraphics[width=.31\textwidth]{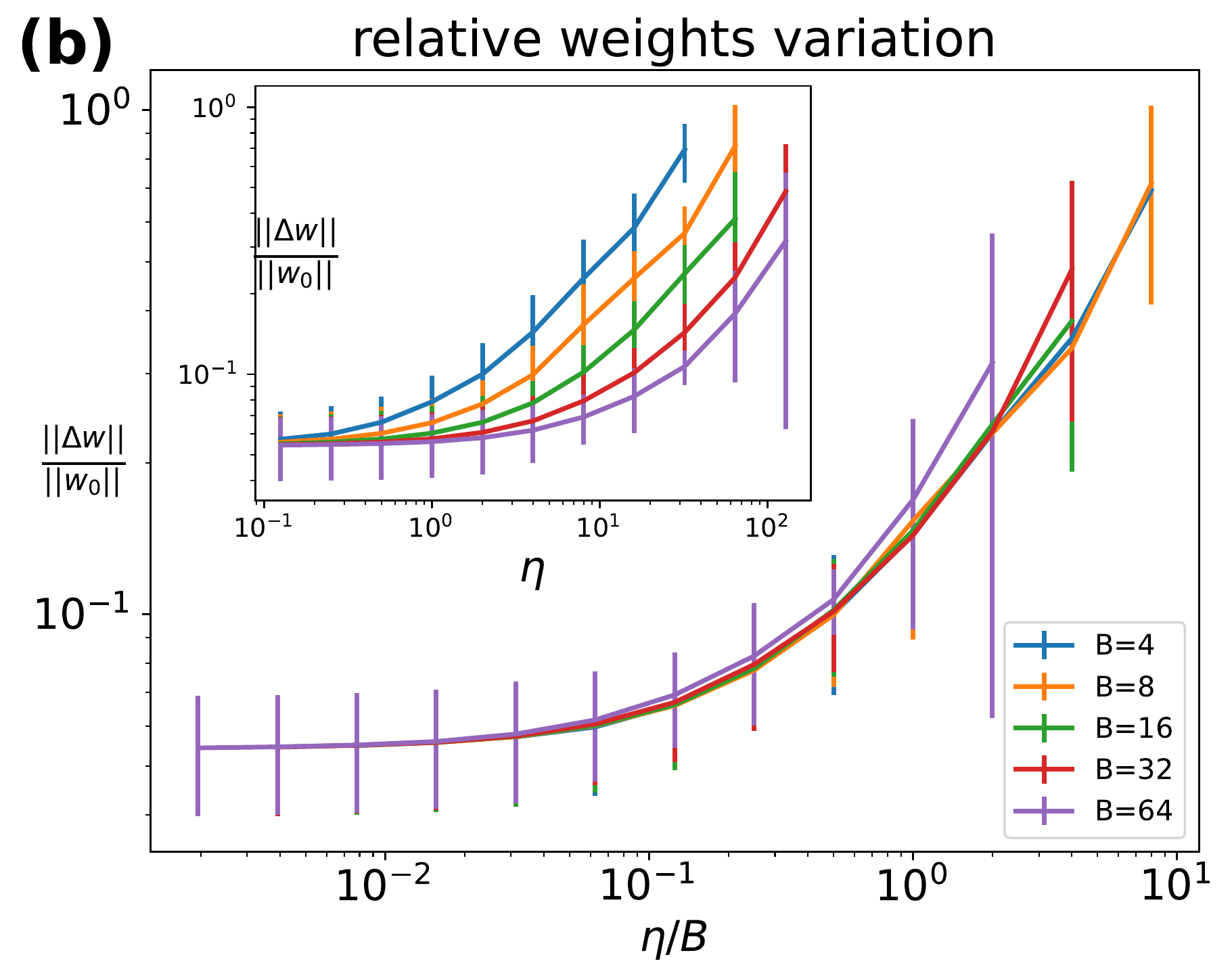}
    \includegraphics[width=.31\textwidth]{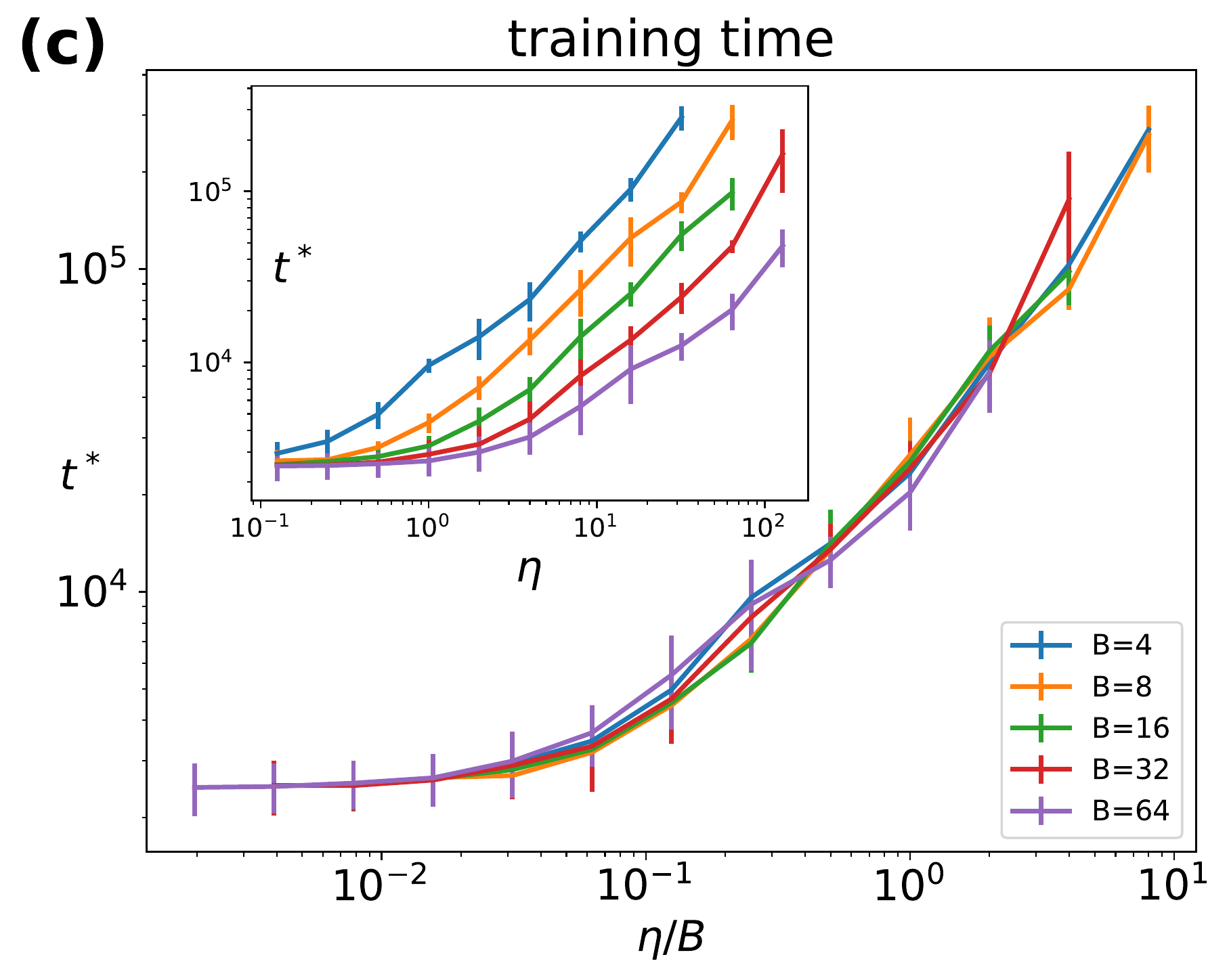}
    \caption{\textbf{FC on MNIST, $\alpha=1$, $P=1024$: (a) test error, (b) relative weight variation, (c) training time with respect to learning rate $\eta$ and batch size $B$.} The represented quantities depend on the ratio $\eta/B$.}
    \label{fig:FC-eta_B}
\end{figure}

\begin{figure}[H]
    \centering
    \includegraphics[width=.31\textwidth]{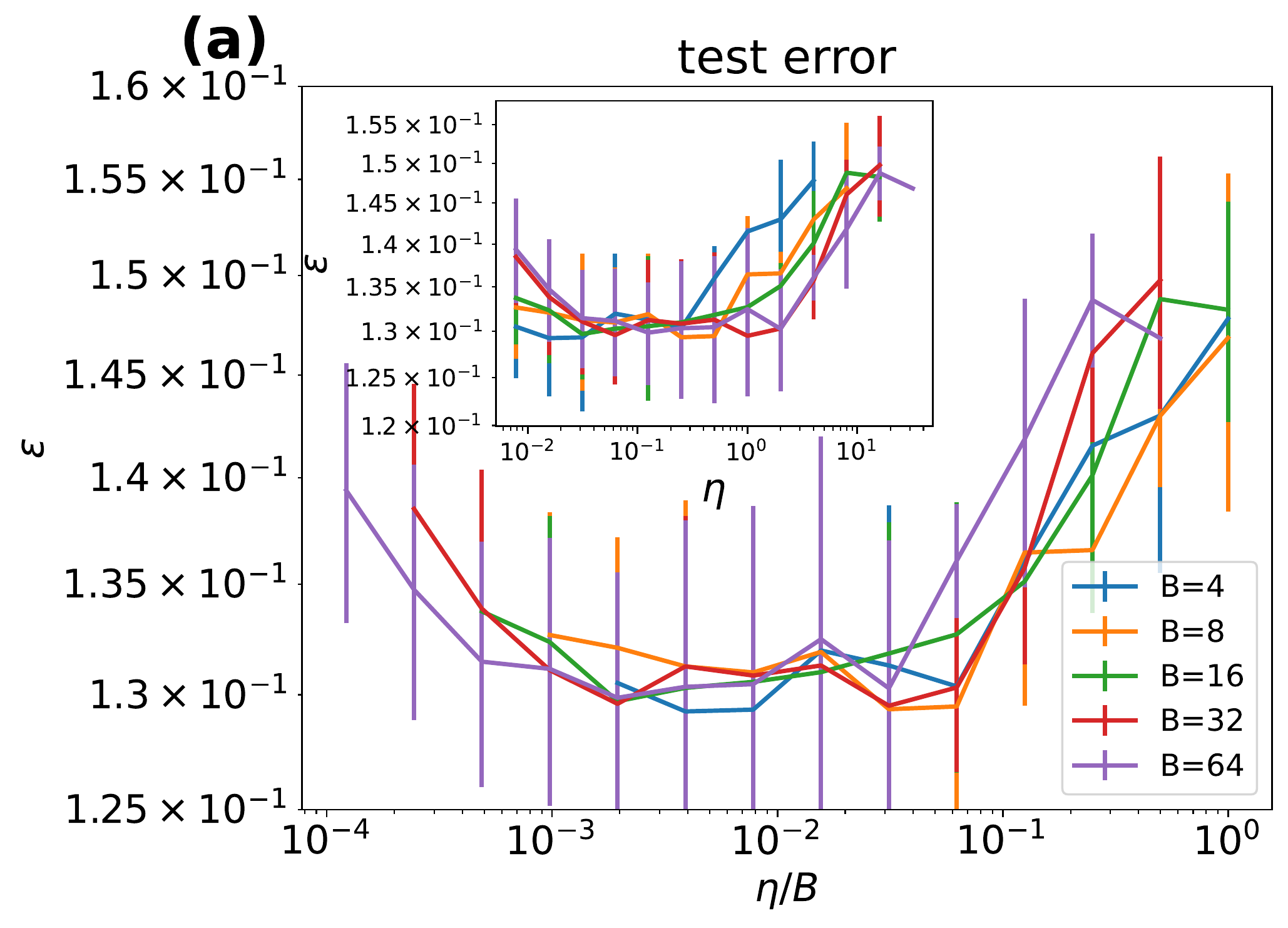}
    \includegraphics[width=.31\textwidth]{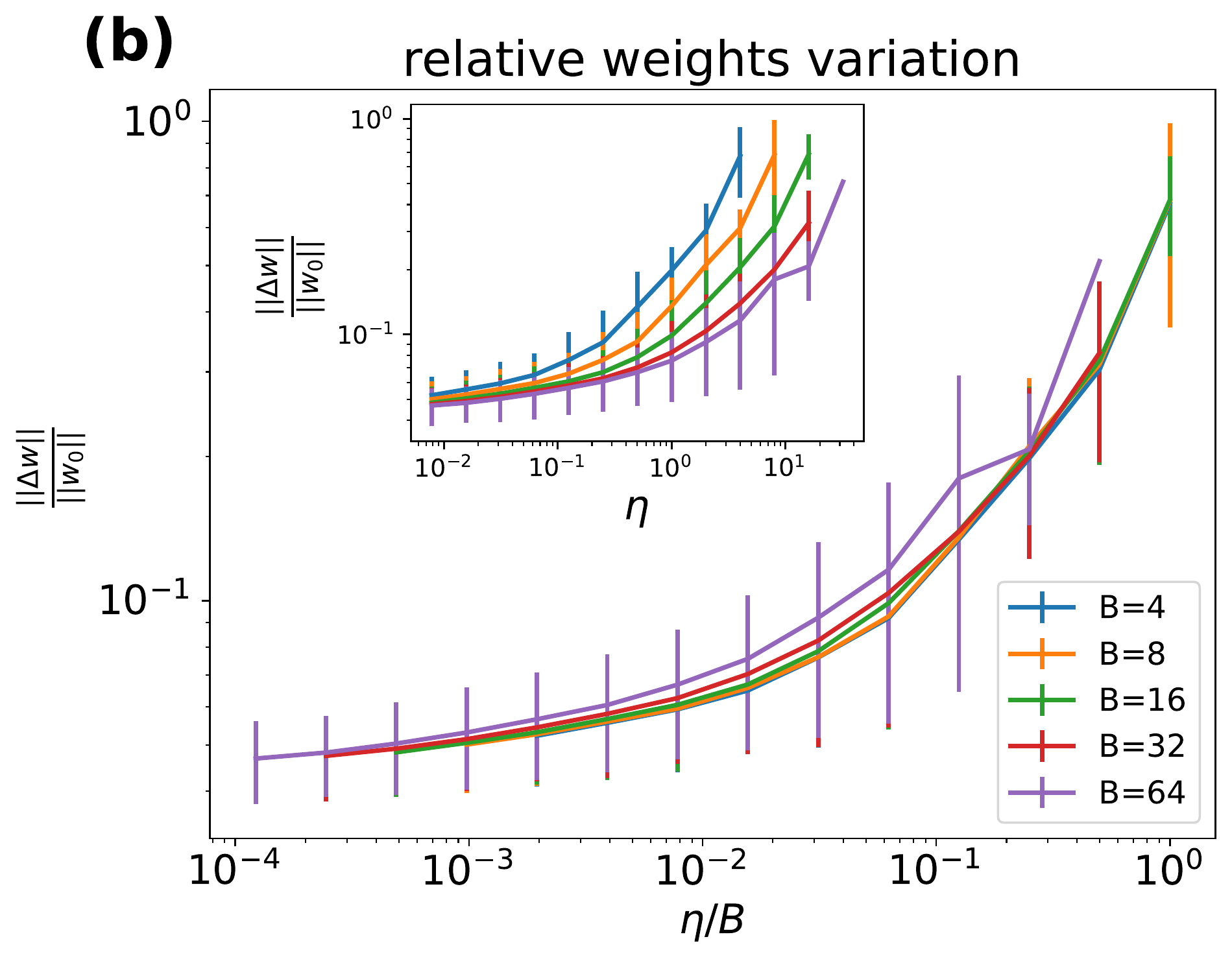}
    \includegraphics[width=.31\textwidth]{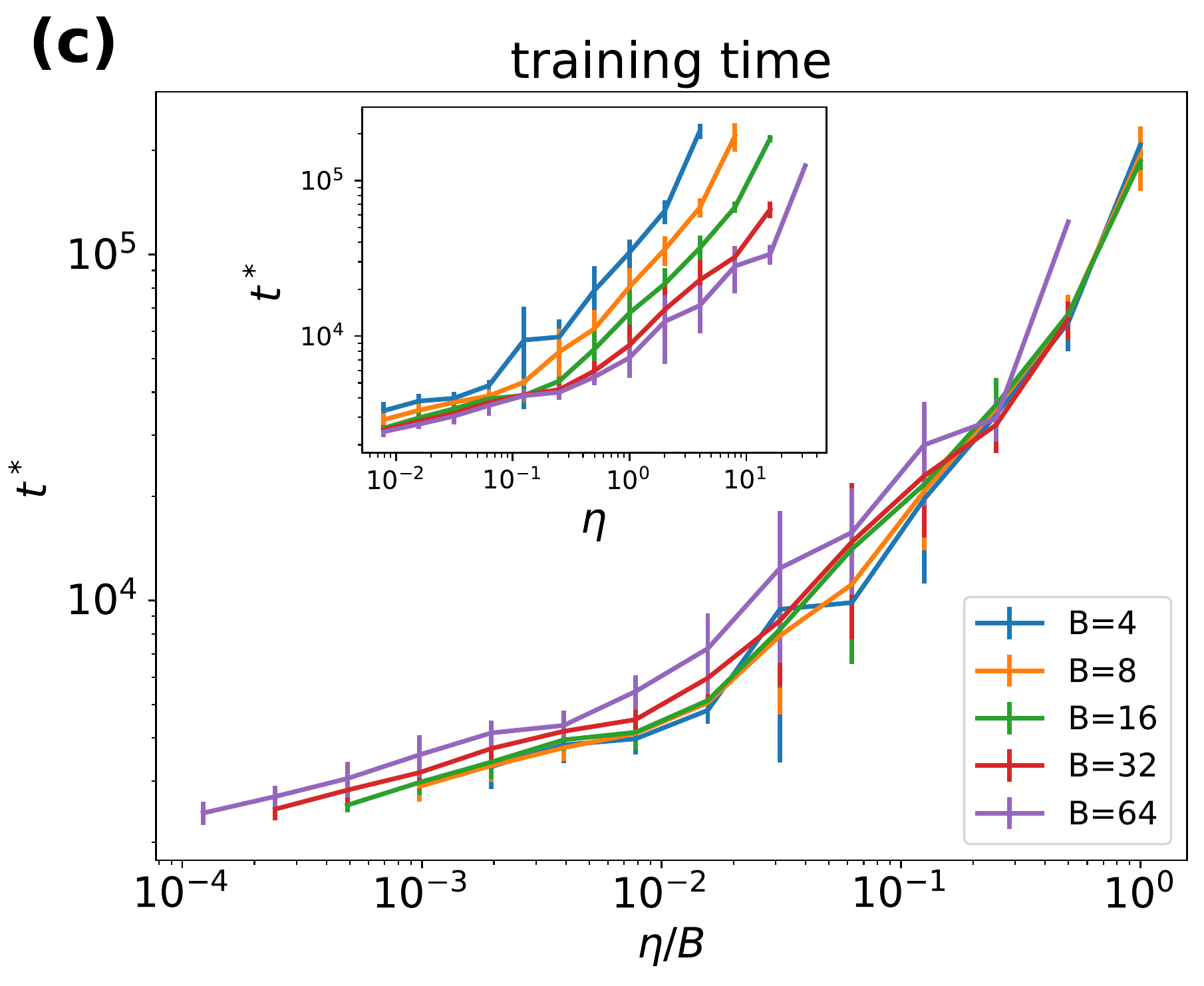}
    \caption{\textbf{CNN (MNAS) on CIFAR, $\alpha=1$, $P=1024$: (a) test error, (b) relative weight variation, (c) training time with respect to learning rate $\eta$ and batch size $B$.} The represented quantities depend on the ratio $\eta/B$.}
    \label{fig:CNN-eta_B}
\end{figure}

\subsection{Error estimation}
\label{app:error}
\begin{figure}[H]
    \centering
    \includegraphics[width=\textwidth]{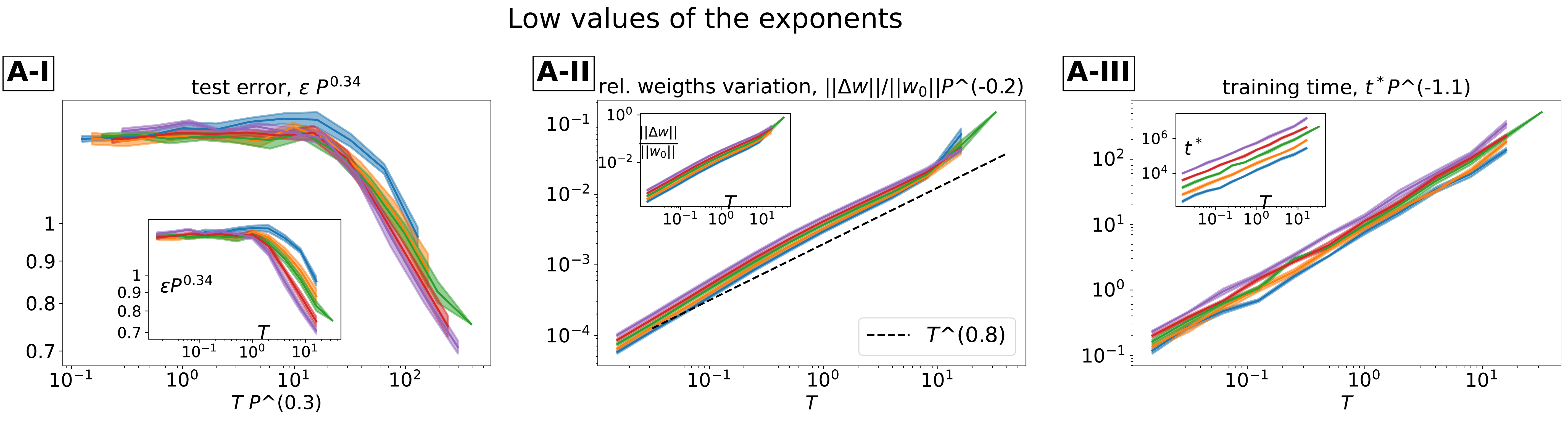}\vspace{.25cm}
    \includegraphics[width=\textwidth]{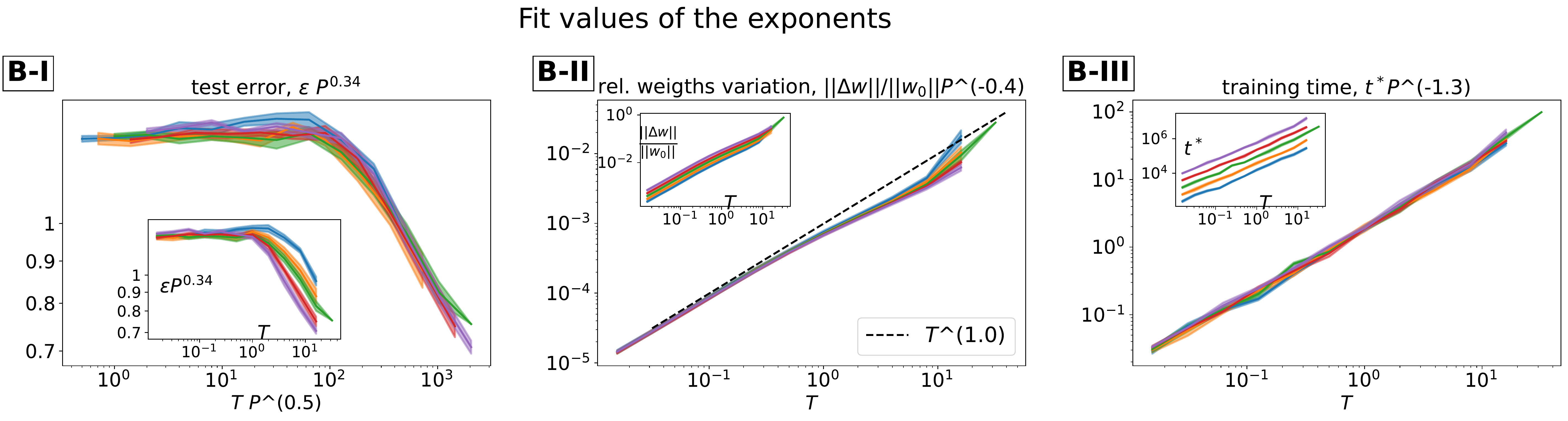}\vspace{.25cm}
    \includegraphics[width=\textwidth]{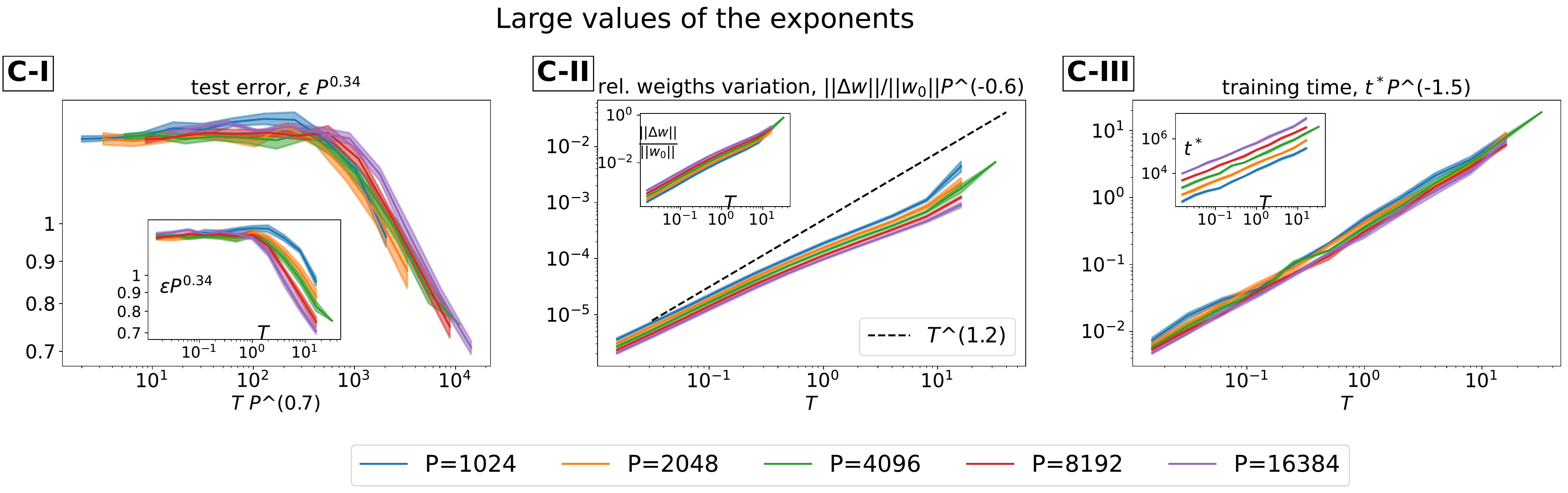}
    \caption{\textbf{Error estimation on the exponents, FC on CIFAR, lazy regime, $\alpha=32768$, $B=16$.}
    \textit{First column (A-I, B-I, C-I): test error $\epsilon$ vs $T$.} The insets show that $\epsilon$ starts improving at $T_c$ depending on $P$. The main panels show that the best curves collapse is obtained plotting $\epsilon$ vs $T P^{0.5}$ (B-I) rather than $T P^{0.3}$ (A-I) or $T P^{0.7}$ (C-I), indicating $T_c\sim P^{-a}$ with $a=0.5\pm 0.2$.
    \textit{Second column (A-II, B-II, C-II): relative weight variation $\Delta w$ vs $T$.} The insets show that $\Delta w$ increases with both $T$ and $P$. The main panels show that the best curve collapse is obtained plotting $\Delta w P^{-0.4}$ (B-II) rather than $\Delta w P^{-0.2}$ (A-II) or $\Delta w P^{-0.6}$ (C-II). Similarly, the slope of the curve is best matched by $T^1$ (B-II) rather than $T^{0.8}$ (A-II) or $T^{1.2}$ (C-II). This indicates that  $\Delta w\sim P^{\ec}T^\ed$ with $\ec=0.4\pm 0.2$ and $\ed = 1 \pm 0.2$.
    \textit{Third column (A-III, B-III, C-III): training time $t^*$ vs $T$.} The insets show that $t^*$ increases with both $T$ and $P$. The main panels show that the best curve collapse is obtained plotting $t^* P^{-1.3}$ (B-III) rather than $t^* P^{-1.1}$ (A-III) or $t^* P^{-1.5}$ (C-III), indicating $t^*\sim TP^{b}$ with $b=1.3\pm 0.2$.
    }
    \label{fig:error}
\end{figure}

This section describes the method used to estimate errors on the exponents presented in Table \ref{tab:exponents}. We choose the exponents such that the rescaled curves overlap (i.e. the curves `collapses'). We estimate the error bars on the exponents based on the quality of this collapse, which we determine to be approximately $\pm 0.2$. To illustrate this process, we consider the data for one example, the fully-connected architecture on MNIST in the lazy regime (Fig. \ref{fig:error}).

In the first column of Fig. \ref{fig:error} (A-I, B-I, C-I), we observe that the test error $\epsilon$ starts decreasing at a characteristic temperature $T_c$, which depends on $P$ as $T_c \sim P^{-a}$. Therefore, plotting $\epsilon$ versus $T P^a$ should align the curves. We find that $a=0.5$ produces the best collapse, while $a=0.3$ (A-I) and $a=0.7$ (C-I) respectively underestimate and overestimate the value of $a$. Hence, we estimate $a$ to be $0.5 \pm 0.2$. The same procedure is used to estimate the errors on the exponents $\ec$, $\ed$ of $\Delta w\sim T^\ed P^\ec$ (Fig. \ref{fig:error} (A-II, B-II, C-II)) and the exponent $b$ of $t^*\sim T P^b$ (Fig. \ref{fig:error} (A-III, B-III, C-III)).

\subsection{Perceptron dynamics}
\begin{figure*}
    \centering
    \includegraphics[width=0.48\textwidth]{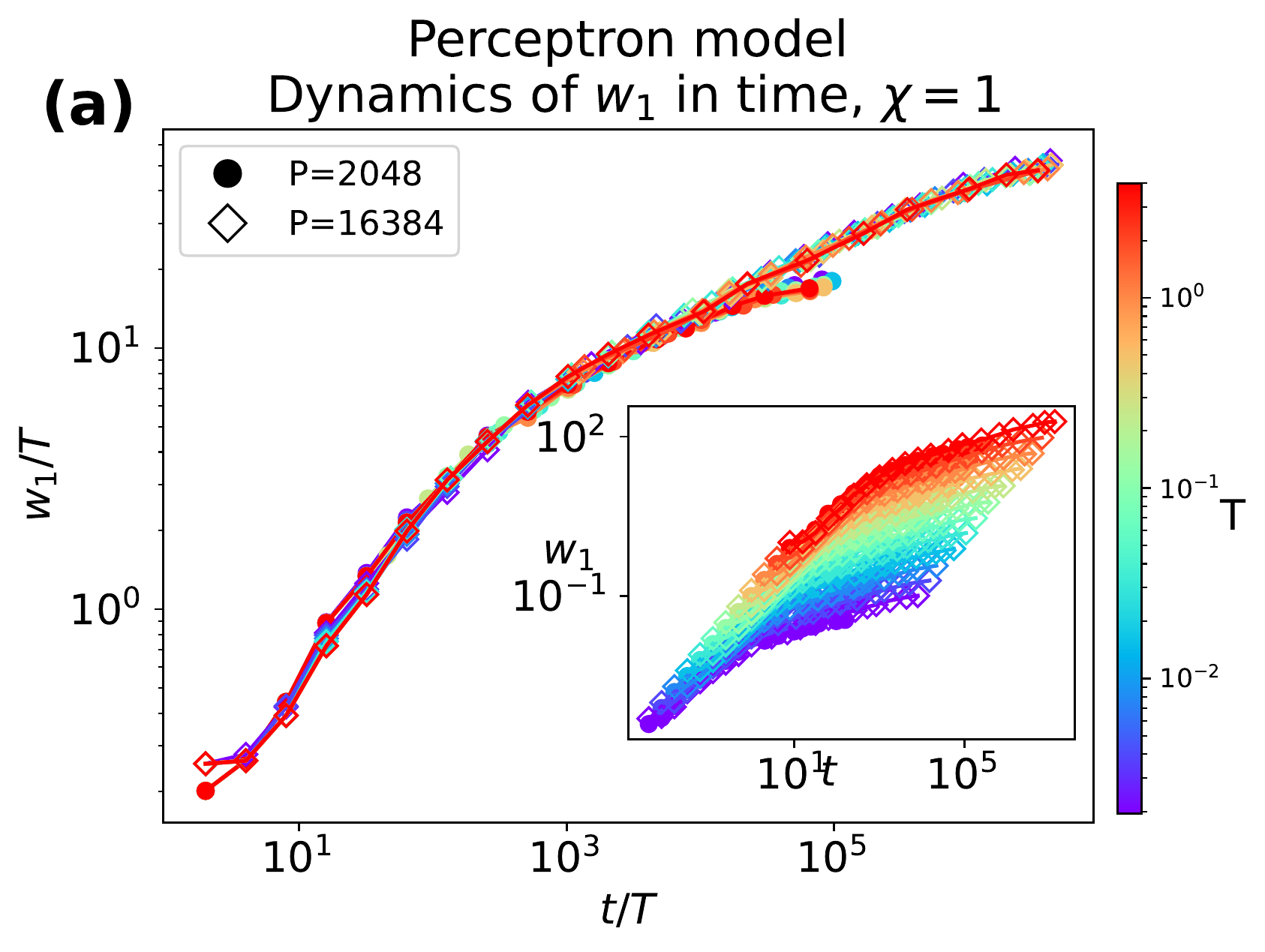}
    \includegraphics[width=0.48\textwidth]{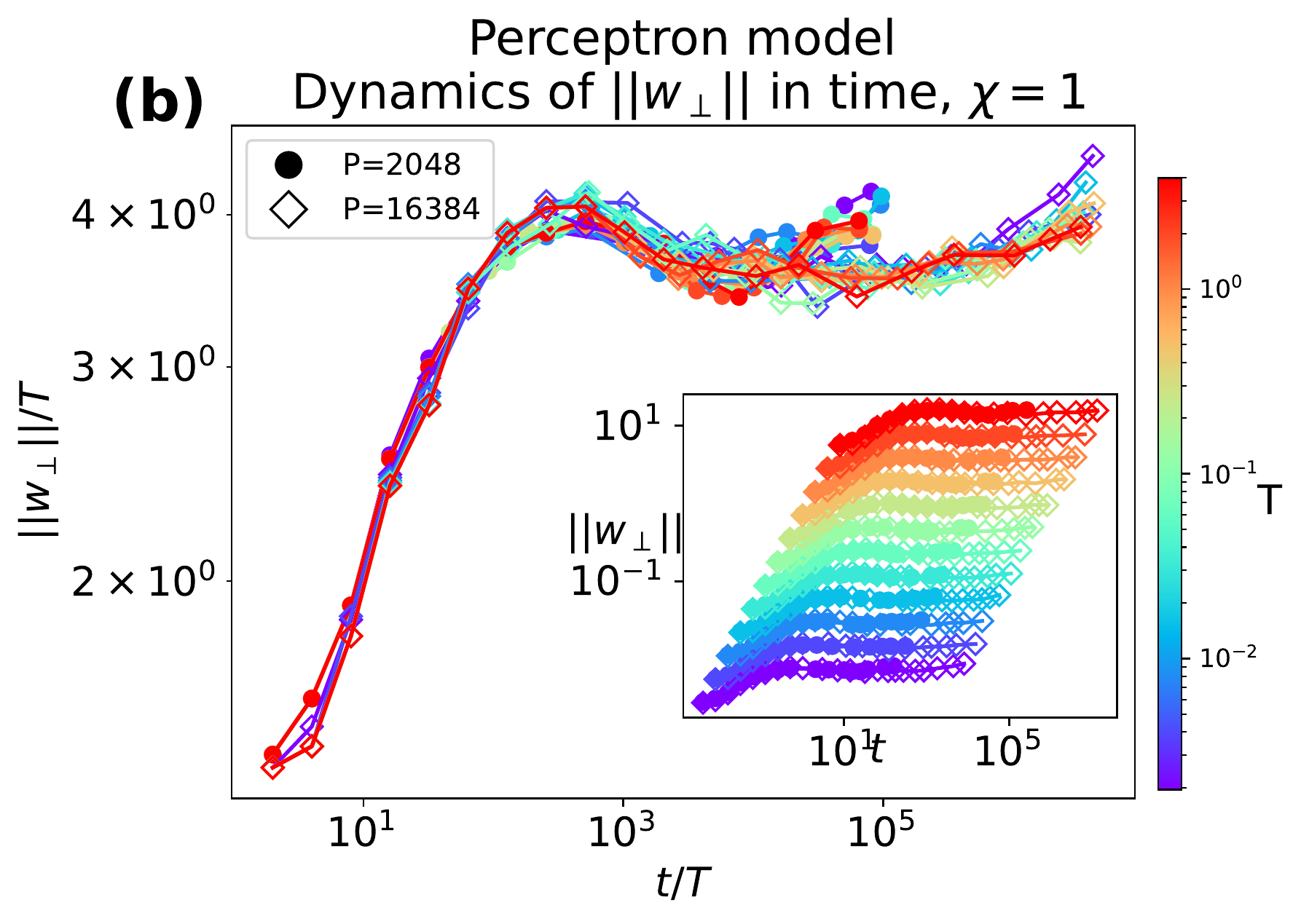}
    \caption{\textbf{Training dynamics in the perceptron model.}
    Data are obtained with data distribution $\chi=1$, dimension $d=128$, batch size $B=2$, varying learning rate $\eta$ ($T=\eta/B$).
    \textbf{(a)} Evolution of the weight $w_1$ with respect to time $t$ (=number of steps times learning rate) for different SGD temperatures $T$ (colors) and training set sizes $P$ (symbols). A larger $T$ corresponds to a larger variation of $w_1$ in a longer time, while the training set size $P$ determines only the end-point of the dynamics.
    \textbf{(b)} Evolution of the norm of the orthogonal weights $||\w_\perp||$ in time, for the same setting as panel (a). For larger $T$, $||\w_\perp||$ reaches higher plateau values while $P$ determines only the end-point of the dynamics. 
    }
    \label{fig:dynamics}
\end{figure*}

\end{document}

%% file: math_commands.tex
\usepackage{amsmath,amsfonts,bm}

\def\eqref#1{equation~\ref{#1}}
\def\1{\bm{1}}

\DeclareMathAlphabet{\mathsfit}{\encodingdefault}{\sfdefault}{m}{sl}
\SetMathAlphabet{\mathsfit}{bold}{\encodingdefault}{\sfdefault}{bx}{n}

\newcommand{\R}{\mathbb{R}}

%% file: figures/scheme_boundary-expo.pdf_tex
%% Creator: Inkscape 1.2 (dc2aeda, 2022-05-15), www.inkscape.org
%% Accompanies image file 'scheme_boundary-expo.pdf' (pdf, eps, ps)
\begingroup%
  \makeatletter%
  \providecommand\color[2][]{%
    \errmessage{(Inkscape) Color is used for the text in Inkscape, but the package 'color.sty' is not loaded}%
    \renewcommand\color[2][]{}%
  }%
  \providecommand\transparent[1]{%
    \errmessage{(Inkscape) Transparency is used (non-zero) for the text in Inkscape, but the package 'transparent.sty' is not loaded}%
    \renewcommand\transparent[1]{}%
  }%
  \providecommand\rotatebox[2]{#2}%
  \newcommand*\fsize{\dimexpr\f@size pt\relax}%
  \newcommand*\lineheight[1]{\fontsize{\fsize}{#1\fsize}\selectfont}%
  \ifx\svgwidth\undefined%
    \setlength{\unitlength}{301.76043653bp}%
    \ifx\svgscale\undefined%
      \relax%
    \else%
      \setlength{\unitlength}{\unitlength * \real{\svgscale}}%
    \fi%
  \else%
    \setlength{\unitlength}{\svgwidth}%
  \fi%
  \global\let\svgwidth\undefined%
  \global\let\svgscale\undefined%
  \makeatother%
  \begin{picture}(1,0.6670247)%
    \lineheight{1}%
    \setlength\tabcolsep{0pt}%
    \put(0,0){\includegraphics[width=\unitlength,page=1]{scheme_boundary-expo.pdf}}%
    \put(0.71172004,0.23557677){\color[rgb]{0,0,0}\makebox(0,0)[lt]{\lineheight{1.04999995}\smash{\begin{tabular}[t]{l}$\partial_{\x} F$\end{tabular}}}}%
    \put(0.67021777,0.3237496){\color[rgb]{0,0,0}\makebox(0,0)[lt]{\lineheight{1.04999995}\smash{\begin{tabular}[t]{l}$\partial_{\x} F_{\parallel}$\end{tabular}}}}%
    \put(0.4273805,0.21485878){\color[rgb]{0,0,0}\makebox(0,0)[lt]{\lineheight{1.04999995}\smash{\begin{tabular}[t]{l}$\partial_{\x} F_{\perp}$\end{tabular}}}}%
    \put(0.63545507,0.48814155){\color[rgb]{0,0,0}\makebox(0,0)[lt]{\lineheight{1.25}\smash{\begin{tabular}[t]{l}$B_{\epsilon}$\end{tabular}}}}%
    \put(0,0){\includegraphics[width=\unitlength,page=2]{scheme_boundary-expo.pdf}}%
    \put(0.15757749,0.57350261){\color[rgb]{0.50196078,0,0.50196078}\makebox(0,0)[lt]{\lineheight{1.25}\smash{\begin{tabular}[t]{l}true boundary\end{tabular}}}}%
    \put(0.47276855,0.58362133){\color[rgb]{0,0,0}\makebox(0,0)[lt]{\lineheight{1.25}\smash{\begin{tabular}[t]{l}boundary $F(\x)=0$\end{tabular}}}}%
    \put(0,0){\includegraphics[width=\unitlength,page=3]{scheme_boundary-expo.pdf}}%
    \put(0.28087372,0.16145765){\color[rgb]{0,0,0}\makebox(0,0)[lt]{\lineheight{1.25}\smash{\begin{tabular}[t]{l}$\x^{-}$\end{tabular}}}}%
    \put(0.50769677,0.46392965){\color[rgb]{0,0,0}\makebox(0,0)[lt]{\lineheight{1.25}\smash{\begin{tabular}[t]{l}$\x^{+}$\end{tabular}}}}%
    \put(0.36622645,0.29946382){\color[rgb]{0,0,0}\makebox(0,0)[lt]{\lineheight{1.25}\smash{\begin{tabular}[t]{l}$\x^{*}$\end{tabular}}}}%
    \put(0,0){\includegraphics[width=\unitlength,page=4]{scheme_boundary-expo.pdf}}%
    \put(0.35739919,0.42121197){\color[rgb]{0,0,0}\makebox(0,0)[lt]{\lineheight{1.04999995}\smash{\begin{tabular}[t]{l}$\delta_{\parallel}$\end{tabular}}}}%
    \put(0,0){\includegraphics[width=\unitlength,page=5]{scheme_boundary-expo.pdf}}%
    \put(0.07204182,0.30744766){\color[rgb]{0,0,0}\makebox(0,0)[lt]{\lineheight{1.04999995}\smash{\begin{tabular}[t]{l}$\delta_{\perp}$\end{tabular}}}}%
    \put(0,0){\includegraphics[width=\unitlength,page=6]{scheme_boundary-expo.pdf}}%
  \end{picture}%
\endgroup%

%% file: main.bbl
\begin{thebibliography}{34}
\providecommand{\natexlab}[1]{#1}
\providecommand{\url}[1]{\texttt{#1}}
\expandafter\ifx\csname urlstyle\endcsname\relax
  \providecommand{\doi}[1]{doi: #1}\else
  \providecommand{\doi}{doi: \begingroup \urlstyle{rm}\Url}\fi

\bibitem[Blanc et~al.(2020)Blanc, Gupta, Valiant, and
  Valiant]{blanc2020implicit}
Blanc, G., Gupta, N., Valiant, G., and Valiant, P.
\newblock Implicit regularization for deep neural networks driven by an
  ornstein-uhlenbeck like process.
\newblock In \emph{Conference on learning theory}, pp.\  483--513. PMLR, 2020.

\bibitem[Chaudhari \& Soatto(2018)Chaudhari and Soatto]{chaudhari2018}
Chaudhari, P. and Soatto, S.
\newblock Stochastic gradient descent performs variational inference, converges
  to limit cycles for deep networks.
\newblock In \emph{2018 Information Theory and Applications Workshop (ITA)},
  pp.\  1--10. IEEE, 2018.

\bibitem[Chaudhari et~al.(2019)Chaudhari, Choromanska, Soatto, LeCun, Baldassi,
  Borgs, Chayes, Sagun, and Zecchina]{entropysgd2019}
Chaudhari, P., Choromanska, A., Soatto, S., LeCun, Y., Baldassi, C., Borgs, C.,
  Chayes, J., Sagun, L., and Zecchina, R.
\newblock Entropy-sgd: Biasing gradient descent into wide valleys.
\newblock \emph{Journal of Statistical Mechanics: Theory and Experiment},
  2019\penalty0 (12):\penalty0 124018, 2019.

\bibitem[Chizat et~al.(2019)Chizat, Oyallon, and Bach]{chizat2019lazy}
Chizat, L., Oyallon, E., and Bach, F.
\newblock On lazy training in differentiable programming.
\newblock \emph{Advances in Neural Information Processing Systems}, 32, 2019.

\bibitem[Dinh et~al.(2017)Dinh, Pascanu, Bengio, and Bengio]{dinh2017}
Dinh, L., Pascanu, R., Bengio, S., and Bengio, Y.
\newblock Sharp minima can generalize for deep nets.
\newblock In \emph{International Conference on Machine Learning}, pp.\
  1019--1028. PMLR, 2017.

\bibitem[Geiger et~al.(2020)Geiger, Spigler, Jacot, and
  Wyart]{geiger2020disentangling}
Geiger, M., Spigler, S., Jacot, A., and Wyart, M.
\newblock Disentangling feature and lazy training in deep neural networks.
\newblock \emph{Journal of Statistical Mechanics: Theory and Experiment},
  2020\penalty0 (11):\penalty0 113301, 2020.

\bibitem[Gnedenko(1943)]{gnedenko1943}
Gnedenko, B.
\newblock Sur la distribution limite du terme maximum d'une serie aleatoire.
\newblock \emph{Annals of mathematics}, pp.\  423--453, 1943.

\bibitem[HaoChen et~al.(2021)HaoChen, Wei, Lee, and Ma]{haochen2021}
HaoChen, J.~Z., Wei, C., Lee, J., and Ma, T.
\newblock Shape matters: Understanding the implicit bias of the noise
  covariance.
\newblock In \emph{Conference on Learning Theory}, pp.\  2315--2357. PMLR,
  2021.

\bibitem[Heskes \& Kappen(1993)Heskes and Kappen]{heskes1993}
Heskes, T.~M. and Kappen, B.
\newblock On-line learning processes in artificial neural networks.
\newblock In \emph{North-Holland Mathematical Library}, volume~51, pp.\
  199--233. Elsevier, 1993.

\bibitem[Hochreiter \& Schmidhuber(1997)Hochreiter and
  Schmidhuber]{hochreiter1997flat}
Hochreiter, S. and Schmidhuber, J.
\newblock Flat minima.
\newblock \emph{Neural computation}, 9\penalty0 (1):\penalty0 1--42, 1997.

\bibitem[Hoffer et~al.(2017)Hoffer, Hubara, and Soudry]{hoffer2017}
Hoffer, E., Hubara, I., and Soudry, D.
\newblock Train longer, generalize better: closing the generalization gap in
  large batch training of neural networks.
\newblock \emph{Advances in neural information processing systems}, 30, 2017.

\bibitem[Jacot et~al.(2018)Jacot, Gabriel, and Hongler]{jacot2018}
Jacot, A., Gabriel, F., and Hongler, C.
\newblock Neural tangent kernel: Convergence and generalization in neural
  networks.
\newblock \emph{Advances in neural information processing systems}, 31, 2018.

\bibitem[Jastrzebski et~al.(2017)Jastrzebski, Kenton, Arpit, Ballas, Fischer,
  Bengio, and Storkey]{jastrzkebski2017}
Jastrzebski, S., Kenton, Z., Arpit, D., Ballas, N., Fischer, A., Bengio, Y.,
  and Storkey, A.
\newblock Three factors influencing minima in sgd.
\newblock \emph{arXiv preprint arXiv:1711.04623}, 2017.

\bibitem[Keskar et~al.(2016)Keskar, Mudigere, Nocedal, Smelyanskiy, and
  Tang]{keskar2016}
Keskar, N.~S., Mudigere, D., Nocedal, J., Smelyanskiy, M., and Tang, P. T.~P.
\newblock On large-batch training for deep learning: Generalization gap and
  sharp minima.
\newblock \emph{arXiv preprint arXiv:1609.04836}, 2016.

\bibitem[Leadbetter et~al.(2012)Leadbetter, Lindgren, and
  Rootz{\'e}n]{leadbetterExtremes}
Leadbetter, M.~R., Lindgren, G., and Rootz{\'e}n, H.
\newblock \emph{Extremes and related properties of random sequences and
  processes}.
\newblock Springer Science \& Business Media, 2012.

\bibitem[LeCun et~al.(2012)LeCun, Bottou, Orr, and M{\"u}ller]{lecun2012}
LeCun, Y.~A., Bottou, L., Orr, G.~B., and M{\"u}ller, K.-R.
\newblock Efficient backprop.
\newblock In \emph{Neural networks: Tricks of the trade}, pp.\  9--48.
  Springer, 2012.

\bibitem[Li et~al.(2017)Li, Tai, and Weinan]{li2017stochastic}
Li, Q., Tai, C., and Weinan, E.
\newblock Stochastic modified equations and adaptive stochastic gradient
  algorithms.
\newblock In \emph{International Conference on Machine Learning}, pp.\
  2101--2110. PMLR, 2017.

\bibitem[Li et~al.(2019)Li, Tai, and Weinan]{li2019stochastic}
Li, Q., Tai, C., and Weinan, E.
\newblock Stochastic modified equations and dynamics of stochastic gradient
  algorithms i: Mathematical foundations.
\newblock \emph{The Journal of Machine Learning Research}, 20\penalty0
  (1):\penalty0 1474--1520, 2019.

\bibitem[Mei et~al.(2018)Mei, Montanari, and Nguyen]{mei2018}
Mei, S., Montanari, A., and Nguyen, P.-M.
\newblock A mean field view of the landscape of two-layer neural networks.
\newblock \emph{Proceedings of the National Academy of Sciences}, 115\penalty0
  (33):\penalty0 E7665--E7671, 2018.

\bibitem[Paccolat et~al.(2021)Paccolat, Petrini, Geiger, Tyloo, and
  Wyart]{paccolat2021geometric}
Paccolat, J., Petrini, L., Geiger, M., Tyloo, K., and Wyart, M.
\newblock Geometric compression of invariant manifolds in neural networks.
\newblock \emph{Journal of Statistical Mechanics: Theory and Experiment},
  2021\penalty0 (4):\penalty0 044001, 2021.

\bibitem[Pesme et~al.(2021)Pesme, Pillaud-Vivien, and
  Flammarion]{flammarion2021}
Pesme, S., Pillaud-Vivien, L., and Flammarion, N.
\newblock Implicit bias of sgd for diagonal linear networks: a provable benefit
  of stochasticity.
\newblock \emph{Advances in Neural Information Processing Systems},
  34:\penalty0 29218--29230, 2021.

\bibitem[Rotskoff \& Vanden-Eijnden(2018)Rotskoff and
  Vanden-Eijnden]{rotskoff2018}
Rotskoff, G.~M. and Vanden-Eijnden, E.
\newblock Neural networks as interacting particle systems: Asymptotic convexity
  of the loss landscape and universal scaling of the approximation error.
\newblock \emph{stat}, 1050:\penalty0 22, 2018.

\bibitem[Shallue et~al.(2018)Shallue, Lee, Antognini, Sohl-Dickstein, Frostig,
  and Dahl]{shallue2018}
Shallue, C.~J., Lee, J., Antognini, J., Sohl-Dickstein, J., Frostig, R., and
  Dahl, G.~E.
\newblock Measuring the effects of data parallelism on neural network training.
\newblock \emph{arXiv preprint arXiv:1811.03600}, 2018.

\bibitem[Sirignano \& Spiliopoulos(2020)Sirignano and
  Spiliopoulos]{sirignano2020}
Sirignano, J. and Spiliopoulos, K.
\newblock Mean field analysis of neural networks: A law of large numbers.
\newblock \emph{SIAM Journal on Applied Mathematics}, 80\penalty0 (2):\penalty0
  725--752, 2020.

\bibitem[Smith et~al.(2020)Smith, Elsen, and De]{smith2020}
Smith, S., Elsen, E., and De, S.
\newblock On the generalization benefit of noise in stochastic gradient
  descent.
\newblock In \emph{International Conference on Machine Learning}, pp.\
  9058--9067. PMLR, 2020.

\bibitem[Smith \& Le(2018)Smith and Le]{smith2018bayesian}
Smith, S.~L. and Le, Q.~V.
\newblock A bayesian perspective on generalization and stochastic gradient
  descent.
\newblock In \emph{International Conference on Learning Representations}, 2018.

\bibitem[Spigler et~al.(2020)Spigler, Geiger, and Wyart]{spigler2020asymptotic}
Spigler, S., Geiger, M., and Wyart, M.
\newblock Asymptotic learning curves of kernel methods: empirical data versus
  teacher--student paradigm.
\newblock \emph{Journal of Statistical Mechanics: Theory and Experiment},
  2020\penalty0 (12):\penalty0 124001, 2020.

\bibitem[Tan et~al.(2019)Tan, Chen, Pang, Vasudevan, Sandler, Howard, and
  Le]{mnasnet}
Tan, M., Chen, B., Pang, R., Vasudevan, V., Sandler, M., Howard, A., and Le,
  Q.~V.
\newblock Mnasnet: Platform-aware neural architecture search for mobile.
\newblock In \emph{Proceedings of the IEEE/CVF Conference on Computer Vision
  and Pattern Recognition}, pp.\  2820--2828, 2019.

\bibitem[Tomasini et~al.(2022)Tomasini, Sclocchi, and
  Wyart]{tomasini2022failure}
Tomasini, U.~M., Sclocchi, A., and Wyart, M.
\newblock Failure and success of the spectral bias prediction for laplace
  kernel ridge regression: the case of low-dimensional data.
\newblock In \emph{International Conference on Machine Learning}, pp.\
  21548--21583. PMLR, 2022.

\bibitem[Wu et~al.(2018)Wu, Ma, et~al.]{wu2018sgd}
Wu, L., Ma, C., et~al.
\newblock How sgd selects the global minima in over-parameterized learning: A
  dynamical stability perspective.
\newblock \emph{Advances in Neural Information Processing Systems}, 31, 2018.

\bibitem[Yang \& Hu(2021)Yang and Hu]{yang2021tensor}
Yang, G. and Hu, E.~J.
\newblock Tensor programs iv: Feature learning in infinite-width neural
  networks.
\newblock In \emph{International Conference on Machine Learning}, pp.\
  11727--11737. PMLR, 2021.

\bibitem[Zhang et~al.(2021)Zhang, Bengio, Hardt, Recht, and
  Vinyals]{zhang2021understanding}
Zhang, C., Bengio, S., Hardt, M., Recht, B., and Vinyals, O.
\newblock Understanding deep learning (still) requires rethinking
  generalization.
\newblock \emph{Communications of the ACM}, 64\penalty0 (3):\penalty0 107--115,
  2021.

\bibitem[Zhang et~al.(2019)Zhang, Li, Nado, Martens, Sachdeva, Dahl, Shallue,
  and Grosse]{zhang2019}
Zhang, G., Li, L., Nado, Z., Martens, J., Sachdeva, S., Dahl, G., Shallue, C.,
  and Grosse, R.~B.
\newblock Which algorithmic choices matter at which batch sizes? insights from
  a noisy quadratic model.
\newblock \emph{Advances in neural information processing systems}, 32, 2019.

\bibitem[Zhang et~al.(2018)Zhang, Saxe, Advani, and Lee]{zhang2018energy}
Zhang, Y., Saxe, A.~M., Advani, M.~S., and Lee, A.~A.
\newblock Energy--entropy competition and the effectiveness of stochastic
  gradient descent in machine learning.
\newblock \emph{Molecular Physics}, 116\penalty0 (21-22):\penalty0 3214--3223,
  2018.

\end{thebibliography}
